\documentclass[runningheads]{llncs}
\usepackage[a4paper,left=3cm,right=3cm]{geometry}

\usepackage[T1]{fontenc}
\usepackage{graphicx}
\usepackage{amsfonts}
\usepackage{amsmath}
\usepackage{algorithm}
\usepackage{algpseudocode}
\usepackage{censor,caption}
\usepackage{multirow}

\usepackage{graphicx,verbatim}
\usepackage{color}

\usepackage{booktabs}
\usepackage{array}
\usepackage{longtable}
\usepackage{tabularx}
\usepackage{threeparttable}
\usepackage{caption}
\usepackage{xurl}
\usepackage{amssymb}
\usepackage{xcolor}
\usepackage[colorlinks=true,linkcolor=blue,citecolor=blue,urlcolor=blue]{hyperref}

\begin{document}
\title{Medical foundation models converge less under label supervision}

\author{
Soroosh Tayebi Arasteh\inst{1,2}$^{\ast}$ \and
Sebastian Ziegelmayer\inst{3} \and
Mahshad Lotfinia\inst{4} \and
Lisa Adams\inst{3} \and
Sven Nebelung\inst{1,2} \and
Jakob Nikolas Kather\inst{5,6,7} \and
Daniel Truhn\inst{1,2}
}
\institute{
Lab for AI in Medicine, RWTH Aachen University, Aachen, Germany \and
Department of Diagnostic and Interventional Radiology, University Hospital RWTH Aachen, Aachen, Germany \and
Department of Diagnostic and Interventional Radiology, TUM University Clinic, School of Medicine and Health, Klinikum rechts der Isar, Technical University of Munich, Munich, Germany \and
Pattern Recognition Lab, Friedrich-Alexander-Universit\"at Erlangen-N\"urnberg, Erlangen, Germany \and
Else Kroener Fresenius Center for Digital Health, Technical University Dresden, Dresden, Germany. \and
Department of Medicine I, University Hospital Dresden, Dresden, Germany. \and
National Center for Tumor Diseases (NCT), University Hospital Heidelberg, Heidelberg, Germany.
}

\maketitle
{\footnotesize
\noindent$^{\ast}$Correspondence to: Soroosh Tayebi Arasteh (\email{soroosh.arasteh@rwth-aachen.de})
}

\begin{abstract}
Diagnostic classifiers and imaging biomarkers are fitted on the embeddings of medical foundation models. These models are replaced as new versions appear. This practice assumes that different models represent images alike. We tested this assumption with more than 750,000 images from 14 datasets in 5 imaging modalities, 18 public models, and 101 models trained on chest radiographs and histopathology that differ in pretraining objective, label type, size, random seed, initialization, training patients, or training recipe. Agreement was measured as the mutual $k$-nearest-neighbor overlap, the share of an image's nearest neighbors common to two models. Two public models shared on average 0.091 of the 10 nearest neighbors of a chest radiograph. A randomly initialized network shared 0.036 with them. In the four other modalities, they shared at most 0.300. On chest radiographs, agreement depended more on the source dataset and the radiographic view than on the findings. It was not significantly associated with accuracy. A linear mapping between the embeddings of two models, fitted on 4,096 unlabeled radiographs, transferred classifiers for 14 chest findings with 0.987 of their original area under the receiver operating characteristic curve. Convergence depended on the pretraining objective. Label-supervised models converged least at both model sizes in chest radiography and histopathology. On chest radiographs, two label-supervised models that differed only in their random seed shared 0.062 to 0.100 of their nearest neighbors. Two models trained with self-distillation, masked image modeling, or contrastive pretraining shared 0.411 to 0.802. Developers can therefore choose convergence when they choose the pretraining objective. Before replacing a model, users could test on unlabeled images whether its successor returns the same nearest neighbors.
\end{abstract}

\section*{Introduction}

Clinical artificial intelligence increasingly starts from a medical foundation model, a network pretrained once on large image collections and then used as a frozen feature extractor. Public models exist for chest radiographs~\cite{PerezGarcia2025RadDino,Cohen2022TorchXRayVision}, histopathology~\cite{Chen2024UNI,Vorontsov2024Virchow,Zimmermann2024Virchow2,Xu2024GigaPath,Lu2024CONCH,Filiot2024Phikon}, retinal photographs~\cite{Zhou2023RETFound}, and biomedical figures~\cite{Zhang2023BiomedCLIP}, beside the vision encoders of medical vision-language models~\cite{Li2023LLaVAMed,Sellergren2025MedGemma} and general-purpose models trained on natural images~\cite{Oquab2024DINOv2,Simenoni2025DINOv3,Radford2021CLIP,Tschannen2025SigLIP2}. General-purpose models transfer to diagnostic tasks without medical pretraining~\cite{Arasteh2024NonMedicalSsl,Lotfinia2025FederatedSsl}. Diagnostic classifiers and imaging biomarkers~\cite{Aerts2025TumorImagingBench} are fitted on their embeddings. Self-supervised image features are also used to search slide archives for similar cases~\cite{Chen2022SISH}. Deep-learning retrieval of similar cases improved the diagnostic accuracy of radiologists for interstitial lung disease on chest computed tomography~\cite{Choe2022CBIR}.

These models are replaced frequently. The 18 public models in this study appeared between 2020 and 2026. When the model under a classifier is replaced, the classifier has to be refitted with new labels and validated again unless it can be transferred to the new model. Under the guidance of the US Food and Drug Administration on predetermined change control plans, a planned replacement of the model under a cleared device can be specified in advance, together with the protocol that validates it~\cite{FDA2024PCCP}. Transferring a classifier, and comparing, combining, or exchanging models from different groups~\cite{Moor2023Generalist}, assumes that different models represent the same image alike.

The Platonic Representation Hypothesis proposes that representations converge toward one shared model of the world as models grow and train on more data~\cite{Huh2024Platonic}. In medicine, clinical supervision through diagnostic labels or radiology reports is expected to move models toward one clinically meaningful representation~\cite{Tiu2022CheXzero,Zhang2022ConVIRT}. Neither expectation has been tested in a way that distinguishes its possible causes. Convergence is inferred from similarity statistics, from canonical correlation~\cite{Raghu2017SVCCA,Morcos2018PWCCA} and centered kernel alignment (CKA)~\cite{Kornblith2019CKA} to the overlap of nearest-neighbor sets~\cite{Huh2024Platonic}, and from whether one model can be stitched into or translated onto another~\cite{Lenc2015Equivariance,Bansal2021Stitching,Moschella2023Relative,Jha2025Universal}. These statistics fall as the number of images grows at a fixed neighbor count~\cite{Koepke2026PlatosCave}. They also rise with the width and depth of a network~\cite{Groeger2026Aristotelian}. Public models differ in their data, architecture, size, and objective. A comparison between them therefore cannot name the cause. To our knowledge, clinical supervision~\cite{Tiu2022CheXzero} and image-only self-supervision~\cite{PerezGarcia2025RadDino} have not been compared as causes of convergence in one controlled design.

\begin{figure*}[p]
\centering
\includegraphics[width=\textwidth]{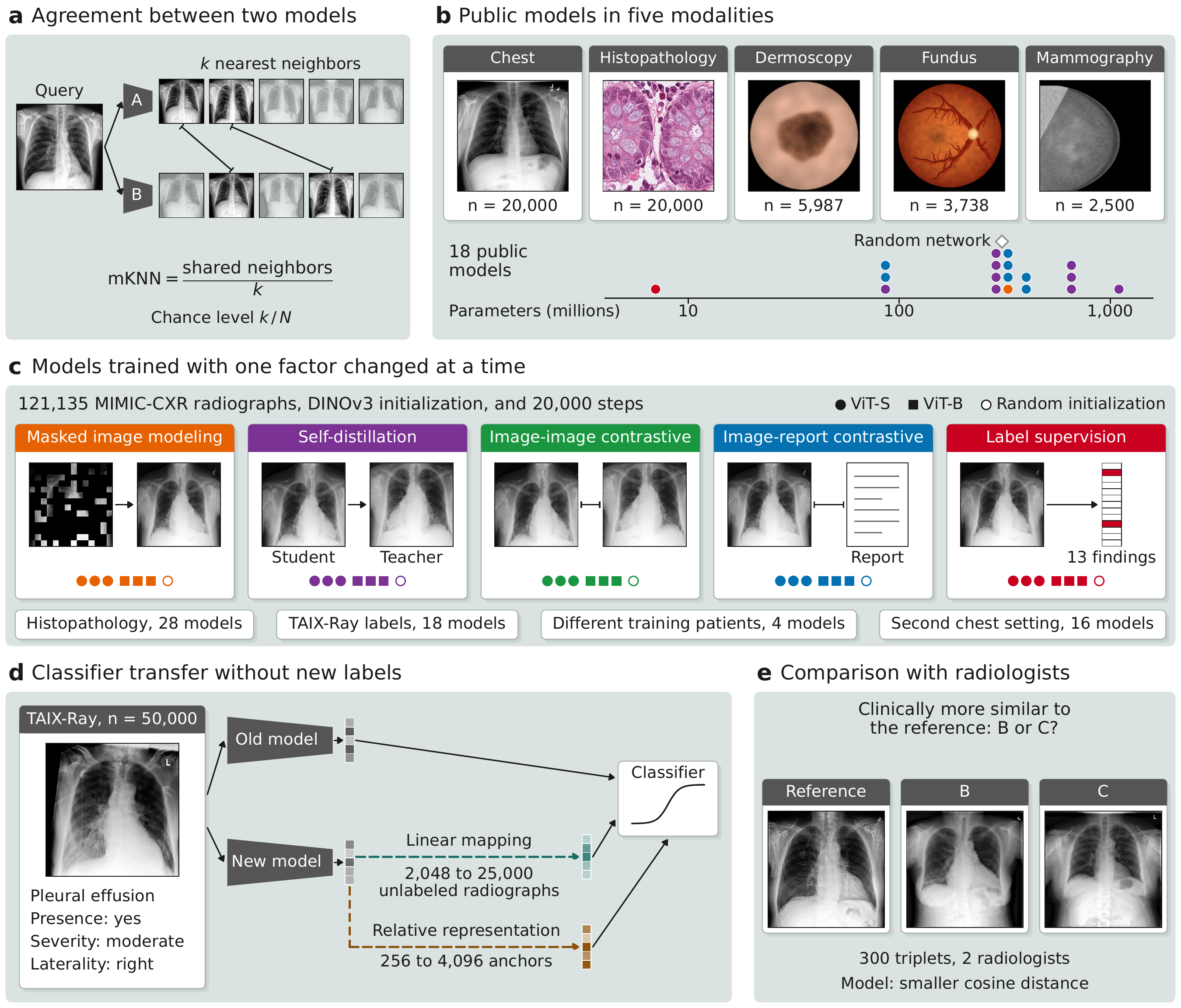}
\caption{Study design. Color marks the pretraining objective in \textbf{b} and \textbf{c}. \textbf{a}, Agreement between two models is the mKNN, the share of an image's $k$ nearest neighbors common to both models, averaged over the $N$ images. Lines join the neighbors common to both models. Faded radiographs are neighbors of one model only. \textbf{b}, The 18 public models are placed at their number of parameters. Blue marks image-text contrastive learning. The random network is a randomly initialized vision transformer. \textbf{c}, Each objective was trained with three random seeds at each model size and once from random initialization. Further models were trained on histopathology patches, on TAIX-Ray labels, on two disjoint sets of MIMIC-CXR training patients, and with a second training recipe on MIMIC-CXR. \textbf{d}, A classifier fitted on an old model is applied to a new model through a linear mapping or through the relative representation, the cosine similarities to anchor radiographs. Transfer is tested for 14 findings on 6 chest datasets and for presence, severity, and laterality on TAIX-Ray. \textbf{e}, Two radiologists chose which of two candidate radiographs is clinically more similar to a reference radiograph. A model chooses the candidate at the smaller cosine distance. The radiograph in \textbf{d} is from TAIX-Ray. The other chest radiographs are from NIH ChestX-ray14. The histopathology patch is from NCT-CRC. The dermoscopy, fundus, and mammography images are drawn. mKNN, mutual $k$-nearest-neighbor overlap; ViT-S and ViT-B, small and base vision transformers.}
\label{fig:overview}
\end{figure*}

Here we measured agreement as the mutual $k$-nearest-neighbor overlap (mKNN), the share of an image's nearest neighbors common to two models (Fig.~\ref{fig:overview}). We compared 18 public models of 7 million to 1.1 billion parameters on chest radiographs, histopathology patches, dermoscopy images, fundus photographs, and mammograms, with a randomly initialized network as the reference. To identify the cause of convergence, we trained 101 models that differ in pretraining objective, label type, model size, random seed, initialization, training patients, or training recipe. They were trained on MIMIC-CXR radiographs~\cite{Johnson2019MIMICCXR}, NCT-CRC histopathology patches~\cite{Kather2018NCT}, and TAIX-Ray radiographs~\cite{Truhn2026TAIXRay}. The objectives are masked image modeling~\cite{He2022MAE}, self-distillation~\cite{Caron2021DINO}, image-image contrastive learning~\cite{Chen2020SimCLR}, image-report contrastive learning~\cite{Radford2021CLIP}, and label supervision. On six chest datasets, we tested whether agreement depends on the findings or on the acquisition, and whether a classifier can be transferred between models without new labels. We compared the distances of the models with the choices of two radiologists who judged which of two radiographs is clinically more similar to a reference radiograph. Two chest datasets lie outside the documented pretraining data of every model. TAIX-Ray~\cite{Truhn2026TAIXRay} ($n = 50{,}000$ bedside radiographs of 11,475 patients) records graded and lateralized findings and was released after every model. ReXGradient-160K~\cite{Zhang2025Rexgradient} ($n = 20{,}000$ radiographs of 13,721 patients from 79 sites) records the scanner manufacturer. The study analyzes 755,207 images from 14 datasets of institutions in at least 7 countries, with 245,023 patients in the 6 datasets that release patient identifiers. It compares 126 models, 25 public and 101 trained under controlled conditions. Four hypotheses were fixed as decision rules before the results were computed. Table~\ref{etab:glossary} defines the terms.

\begin{table}[p]
\centering
\caption{Glossary of the terms of this study. The terms are grouped by data, models, pretraining objectives, agreement, classifier transfer, radiologists, and reporting. Each has the meaning given here in the main text, the figures, the tables, and the supplement. AUROC, area under the receiver operating characteristic curve; CI, confidence interval; SD, standard deviation.}
\label{etab:glossary}
{\scriptsize\hyphenpenalty=10000\exhyphenpenalty=10000
\setlength{\tabcolsep}{3pt}
\renewcommand{\arraystretch}{1.0}
\begin{tabular*}{\textwidth}{@{\extracolsep{\fill}}>{\raggedright\arraybackslash}p{0.235\textwidth}>{\raggedright\arraybackslash}p{0.75\textwidth}@{}}
\toprule
Term & Definition \\
\midrule
\multicolumn{2}{@{}l}{\textbf{Data}} \\
\midrule
Chest pool & 650{,}982 frontal radiographs of 6 chest datasets. Public models are compared on 20{,}000. \\
Held-out pool & The validation and test images of a pool without any patient of a training list. \\
TAIX-Ray pool & 50{,}000 bedside radiographs of 11{,}475 patients with the presence, severity grade, and laterality of each finding. \\
ReXGradient-160K pool & 20{,}000 radiographs of 13{,}721 patients from 79 sites, with the scanner manufacturer. \\
Anchor radiographs & 4{,}096 MIMIC-CXR training radiographs defining the relative representation, 1{,}024 in the primary set. \\
\midrule
\multicolumn{2}{@{}l}{\textbf{Models}} \\
\midrule
Public models & The 18 image and 7 text models of Table~\ref{etab:panel}, open-weight and used frozen. \\
Random network & A randomly initialized ViT-L/16 on the same images, the untrained reference of each pool. \\
Controlled study & The 101 models of Table~\ref{tab:design}, which differ in objective, label type, size, seed, initialization, training patients, or training recipe. \\
Setting & A group of controlled models that share their training images, or the models of such a group at one model size. \\
Second chest setting & The 16 chest models trained with a second training recipe on 180{,}300 MIMIC-CXR radiographs. \\
ViT-S and ViT-B & The small and the base vision transformer with 16-pixel patches. \\
Initialization & The DINOv3 checkpoint or the random weights at the start of training. \\
Seed pair & Two models that differ only in their random seed. \\
Size pair & A ViT-S model and a ViT-B model of one objective and setting. \\
\midrule
\multicolumn{2}{@{}l}{\textbf{Pretraining objectives}} \\
\midrule
Masked image modeling & Reconstruction of the pixels of 75\% of the patches, which are zeroed before the encoder. \\
Self-distillation & DINO, a student network trained to match a teacher network that is its moving average. \\
Image-image contrastive learning & A loss that makes two augmented views of an image more similar than views of other images, as in SimCLR. \\
Image-report contrastive learning & The CLIP loss between a radiograph and its report, with both encoders trained. \\
Contrastive pretraining & Image-image and image-report contrastive learning together. \\
Label supervision & A linear head on the 13 chest findings, the 9 tissue classes, or the TAIX-Ray labels. \\
\midrule
\multicolumn{2}{@{}l}{\textbf{Agreement}} \\
\midrule
Mutual $k$-nearest-neighbor overlap (mKNN) & The share of the $k$ nearest neighbors of an image that are common to two models, averaged over the images. \\
Centered kernel nearest-neighbor alignment (CKNNA) & Kernel alignment restricted to the image pairs that are nearest neighbors in both models. \\
Neighbor count $k$ & 0.4\% of the number of images, which is 10 at 2{,}500 images. \\
Chance level & The mKNN of two random neighbor sets, 0.004 at 2{,}500 images. \\
Convergence & Agreement between models trained separately, measured on seed pairs and size pairs. \\
\midrule
\multicolumn{2}{@{}l}{\textbf{Classifier transfer}} \\
\midrule
Old model and new model & The model used to fit a classifier, and the model that replaces it. \\
Linear mapping & A least-squares map from new to old embeddings, fitted on unlabeled radiographs (Eq.~\ref{eq:map}). \\
Relative representation & The cosine similarities of an image to the anchor radiographs in one model (Eq.~\ref{eq:anchor}). \\
Retention & The AUROC after transfer divided by the reference AUROC of Eq.~\ref{eq:retention}. \\
Refit & A classifier fitted on the new model with its own labels, on 50 to 25{,}000 radiographs. \\
\midrule
\multicolumn{2}{@{}l}{\textbf{Radiologists}} \\
\midrule
Triplet & A reference chest radiograph and two candidates. A radiologist chose the candidate clinically more similar to the reference. \\
Well-posed triplet & A triplet in which only one candidate shares a finding with the reference. \\
\midrule
\multicolumn{2}{@{}l}{\textbf{Reporting}} \\
\midrule
Delete-one-model jackknife & A summary over model pairs recomputed with each model left out in turn. Its standard error is the SD, and the 95\% CI is the mean $\pm$ 1.96 SD. \\
$p_{\mathrm{FDR}}$ & A $p$-value corrected by the Benjamini-Hochberg false discovery rate in a named family. \\
Pre-specified rules & The decision rules of the four hypotheses, fixed before any result (Supplementary Table~\ref{stab:rules}). \\
\bottomrule
\end{tabular*}}
\end{table}

Public models agree more than a random network in four of the five modalities, but they share few nearest neighbors. Convergence depends on the pretraining objective. Label supervision yields the least convergent models in every controlled setting. On chest radiographs, two label-supervised models that differ only in their random seed share few nearest neighbors, while models trained with the other four objectives share many. Changing the training patients of an image-image contrastive model lowers agreement less than changing its random seed. On chest radiographs, agreement depends more on the source dataset and the radiographic view than on the findings. It is not significantly associated with accuracy. A linear mapping fitted on unlabeled radiographs transfers classifiers of chest findings between models with no average loss of AUROC. Severity classifiers keep their performance after transfer. Laterality classifiers keep most of it. No model agrees with the radiologists' choice of the clinically more similar radiograph significantly above chance. Supplementary Table~\ref{stab:summary} summarizes every finding.

\section*{Results}

Agreement between two models is the mKNN~\cite{Huh2024Platonic}, the share of an image's $k$ nearest neighbors under cosine distance that are common to the two models, averaged over the images. With $k$ set to 0.4\% of the number of images, two random neighbor sets share 0.004 of their neighbors. The centered kernel nearest-neighbor alignment (CKNNA) of the same authors is reported beside mKNN. Agreement, the area under the receiver operating characteristic curve (AUROC), and accuracy are given as the mean $\pm$ standard deviation (SD) with the 95\% confidence interval (CI) in brackets. Correlations and coefficients of determination are given with a $p$-value. Tests are two-sided at 0.05. The exception is the one-sided comparison of an observed value with a permutation reference. Its $p$ is $(b + 1)/(m + 1)$ when $b$ of the $m$ reference values are at least as large as the observed value. A $p$-value corrected with the Benjamini-Hochberg false discovery rate (FDR)~\cite{BenjaminiHochberg1995} within a named family is written $p_{\mathrm{FDR}}$.

\subsection*{Agreement between public medical foundation models across five imaging modalities}

We measured agreement between 18 public image models of 7 million to 1.1 billion parameters (Table~\ref{etab:panel}) on 20,000 chest radiographs from 6 datasets, 20,000 histopathology patches, 5,987 dermoscopy images, 3,738 fundus photographs, and 2,500 mammograms (Table~\ref{tab:sources}). The random network is a randomly initialized vision transformer (ViT) embedding the same images.

\begin{table}[p]
\centering
\caption{The public models. The 18 image models are grouped by the objective of their pretraining. The randomly initialized ViT-L/16 used as the reference and the 7 text models that embed the reports follow them. For the three vision-language models, the model is the vision encoder. Lineage is the recipe or the initialization common to two or more models. A model without a shared lineage is marked None. Parameters are in millions. Width is the embedding dimension, and year is the release year used in the size analysis. Revision is the first seven characters of the Hugging Face commit that was recorded at extraction. For the text models, it is the latest commit of the repository, which predates every extraction. N/A, not applicable; ViT-L/16, large vision transformer with 16-pixel patches.}
\label{etab:panel}
{\footnotesize\hyphenpenalty=10000\exhyphenpenalty=10000
\setlength{\tabcolsep}{3.5pt}
\renewcommand{\arraystretch}{1.0}
\begin{tabular*}{\textwidth}{@{\extracolsep{\fill}}>{\raggedright\arraybackslash}p{0.285\textwidth}llrrrl@{}}
\toprule
Model & Domain & Lineage & Parameters & Width & Year & Revision \\
\midrule
\multicolumn{7}{@{}l}{\textbf{Image models, self-distillation}} \\
\midrule
DINOv2 ViT-L~\cite{Oquab2024DINOv2} & Natural images & DINOv2 recipe & 307 & 1{,}024 & 2023 & 47b73ee \\
DINOv3 ViT-L~\cite{Simenoni2025DINOv3} & Natural images & None & 307 & 1{,}024 & 2025 & ea8dc28 \\
RAD-DINO~\cite{PerezGarcia2025RadDino} & Chest radiographs & DINOv2 recipe & 86 & 768 & 2024 & 110cbc1 \\
Phikon-v2~\cite{Filiot2024Phikon} & Histopathology & DINOv2 recipe & 307 & 1{,}024 & 2024 & 2ae989a \\
Prov-GigaPath~\cite{Xu2024GigaPath} & Histopathology & DINOv2 recipe & 1{,}100 & 1{,}536 & 2024 & 64f9e26 \\
UNI~\cite{Chen2024UNI} & Histopathology & DINOv2 recipe & 307 & 1{,}024 & 2024 & b55a5ec \\
UNI2-h~\cite{Chen2024UNI} & Histopathology & DINOv2 recipe & 681 & 1{,}536 & 2025 & d517a8d \\
Virchow~\cite{Vorontsov2024Virchow} & Histopathology & DINOv2 recipe & 632 & 1{,}280 & 2024 & 19eebc8 \\
Virchow2~\cite{Zimmermann2024Virchow2} & Histopathology & DINOv2 recipe & 632 & 1{,}280 & 2024 & 3158645 \\
\midrule
\multicolumn{7}{@{}l}{\textbf{Image models, masked image modeling}} \\
\midrule
RETFound~\cite{Zhou2023RETFound} & Retinal images & None & 307 & 1{,}024 & 2023 & 42a93fe \\
\midrule
\multicolumn{7}{@{}l}{\textbf{Image models, image-text contrastive}} \\
\midrule
CLIP ViT-L/14~\cite{Radford2021CLIP} & Natural images & CLIP ViT-L/14 & 307 & 768 & 2021 & 32bd642 \\
SigLIP 2 ViT-L~\cite{Tschannen2025SigLIP2} & Natural images & None & 307 & 1{,}024 & 2025 & 4948821 \\
LLaVA-Med~\cite{Li2023LLaVAMed} & Natural images & CLIP ViT-L/14 & 307 & 1{,}024 & 2024 & 627be53 \\
LLaVA-OneVision~\cite{Li2024LLaVAOneVision} & General images & SigLIP & 400 & 1{,}152 & 2024 & 0d50680 \\
MedGemma~\cite{Sellergren2025MedGemma} & Medical images & SigLIP & 400 & 1{,}152 & 2026 & 9185054 \\
BiomedCLIP~\cite{Zhang2023BiomedCLIP} & Biomedical figures & None & 86 & 512 & 2023 & 9f341de \\
CONCH~\cite{Lu2024CONCH} & Histopathology & None & 86 & 512 & 2024 & f9ca9f8 \\
\midrule
\multicolumn{7}{@{}l}{\textbf{Image models, label supervision}} \\
\midrule
TorchXRayVision~\cite{Cohen2022TorchXRayVision} & Chest radiographs & None & 7 & 1{,}024 & 2020 & N/A \\
\midrule
\multicolumn{7}{@{}l}{\textbf{Reference}} \\
\midrule
Random network~\cite{Dosovitskiy2021ViT} & None & None & 307 & 1{,}024 & N/A & N/A \\
\midrule
\multicolumn{7}{@{}l}{\textbf{Text models}} \\
\midrule
Model & \multicolumn{3}{l}{Pretraining data and objective} & Width & & Revision \\
\cmidrule{1-7}
PubMedBERT~\cite{Gu2021PubMedBERT} & \multicolumn{3}{l}{Biomedical text, masked language modeling} & 768 & & e1354b7 \\
MedCPT query encoder~\cite{Jin2023MedCPT} & \multicolumn{3}{l}{Biomedical retrieval, contrastive} & 768 & & d83a36c \\
MedCPT article encoder~\cite{Jin2023MedCPT} & \multicolumn{3}{l}{Biomedical retrieval, contrastive} & 768 & & d05a736 \\
BiomedCLIP text encoder~\cite{Zhang2023BiomedCLIP} & \multicolumn{3}{l}{Biomedical figures, image-text contrastive} & 512 & & 9f341de \\
CONCH text encoder~\cite{Lu2024CONCH} & \multicolumn{3}{l}{Histopathology, image-text contrastive} & 512 & & f9ca9f8 \\
MedGemma 27B~\cite{Sellergren2025MedGemma} & \multicolumn{3}{l}{Medical language model, last-token state} & 5{,}376 & & 2d3e00e \\
SapBERT~\cite{Liu2021SapBERT} & \multicolumn{3}{l}{Biomedical entity linking, contrastive} & 768 & & 090663c \\
\bottomrule
\end{tabular*}}
\end{table}

\begin{table}[p]
\centering
\caption{The 14 data sources, grouped by imaging modality. Images and patients are the counts in the pools of this study. Among the roles, agreement marks the images of an agreement pool, and training marks the training images of the controlled study. Anchors marks the source of the anchors of the relative representation. Reports marks the radiographs compared with their reports, and cross-site marks the scoring of classifiers fitted on MIMIC-CXR. The model cards of MedGemma, RAD-DINO, and TorchXRayVision name MIMIC-CXR as pretraining data. The model cards of RAD-DINO and TorchXRayVision also name CheXpert, NIH ChestX-ray14, and PadChest. No model card names another source. TAIX-Ray and ReXGradient-160K are treated as unseen, and only TAIX-Ray was released after every model. N/A, not applicable, since the source releases no patient identifier.}
\label{tab:sources}
{\scriptsize\hyphenpenalty=10000\exhyphenpenalty=10000
\setlength{\tabcolsep}{3pt}
\renewcommand{\arraystretch}{1.1}
\begin{tabular*}{\textwidth}{@{\extracolsep{\fill}}>{\raggedright\arraybackslash}p{0.16\textwidth}>{\raggedright\arraybackslash}p{0.22\textwidth}rrrrr>{\raggedright\arraybackslash}p{0.15\textwidth}@{}}
\toprule
& & & & \multicolumn{3}{c}{Images per split} & \\
\cmidrule(lr){5-7}
Dataset & Institution and country & Images & Patients & Training & Validation & Test & Roles \\
\midrule
\multicolumn{8}{@{}l}{\textbf{Chest radiography}} \\
\midrule
MIMIC-CXR \cite{Johnson2019MIMICCXR} & Beth Israel Deaconess Medical Center, Boston, USA & 243{,}334 & 63{,}945 & 180{,}289 & 18{,}409 & 44{,}636 & Agreement, anchors, training, reports \\
CheXpert \cite{Irvin2019CheXpert} & Stanford Health Care, USA & 157{,}878 & 57{,}872 & 115{,}458 & 13{,}099 & 29{,}321 & Agreement, reports, cross-site \\
NIH ChestX-ray14 \cite{Wang2017ChestXray14} & NIH Clinical Center, USA & 112{,}120 & 30{,}805 & 77{,}870 & 8{,}654 & 25{,}596 & Agreement, cross-site \\
PadChest \cite{Bustos2020PadChest} & Hospital San Juan, Alicante, Spain & 110{,}525 & 67{,}205 & 79{,}697 & 8{,}783 & 22{,}045 & Agreement, cross-site \\
VinDr-CXR \cite{Nguyen2022VinDrCXR} & Hospital 108 and Hanoi Medical University Hospital, Vietnam & 18{,}000 & N/A & 15{,}000 & 0 & 3{,}000 & Agreement, cross-site \\
VinDr-PCXR \cite{Pham2022VinDrPCXR} & Phu Tho Obstetric and Pediatric Hospital, Vietnam & 9{,}125 & N/A & 6{,}955 & 773 & 1{,}397 & Agreement, cross-site \\
TAIX-Ray \cite{Truhn2026TAIXRay} & University Hospital RWTH Aachen, Germany & 50{,}000 & 11{,}475 & 31{,}942 & 8{,}093 & 9{,}965 & Agreement, training \\
ReXGradient-160K \cite{Zhang2025Rexgradient} & 79 sites of 3 health systems, USA & 20{,}000 & 13{,}721 & 17{,}400 & 1{,}272 & 1{,}328 & Agreement, reports \\
\midrule
\multicolumn{8}{@{}l}{\textbf{Histopathology}} \\
\midrule
NCT-CRC \cite{Kather2018NCT} & NCT Biobank Heidelberg and University Medical Center Mannheim, Germany & 18{,}000 & N/A & 16{,}784 & 1{,}216 & 0 & Agreement, training \\
PatchCamelyon \cite{Veeling2018PCam} & Radboud University Medical Center and University Medical Center Utrecht, the Netherlands & 4{,}000 & N/A & 0 & 0 & 4{,}000 & Agreement \\
\midrule
\multicolumn{8}{@{}l}{\textbf{Dermoscopy}} \\
\midrule
ISIC 2019 \cite{Tschandl2018HAM10000,Combalia2019BCN20000,Codella2019ISIC} & Dermatology centers in several countries & 5{,}987 & N/A & 5{,}987 & 0 & 0 & Agreement \\
\midrule
\multicolumn{8}{@{}l}{\textbf{Fundus photography}} \\
\midrule
APTOS \cite{Karthik2019APTOS} & Aravind Eye Hospital, India & 2{,}974 & N/A & 2{,}974 & 0 & 0 & Agreement \\
Messidor-2 \cite{Decenciere2014Messidor} & Ophthalmology departments in France & 764 & N/A & 0 & 0 & 764 & Agreement \\
\midrule
\multicolumn{8}{@{}l}{\textbf{Mammography}} \\
\midrule
VinDr-Mammo \cite{Nguyen2023VinDrMammo} & Hanoi Medical University Hospital and Hospital 108, Vietnam & 2{,}500 & N/A & 2{,}500 & 0 & 0 & Agreement \\
\bottomrule
\end{tabular*}}
\end{table}

\begin{figure*}[p]
\centering
\includegraphics[width=\textwidth]{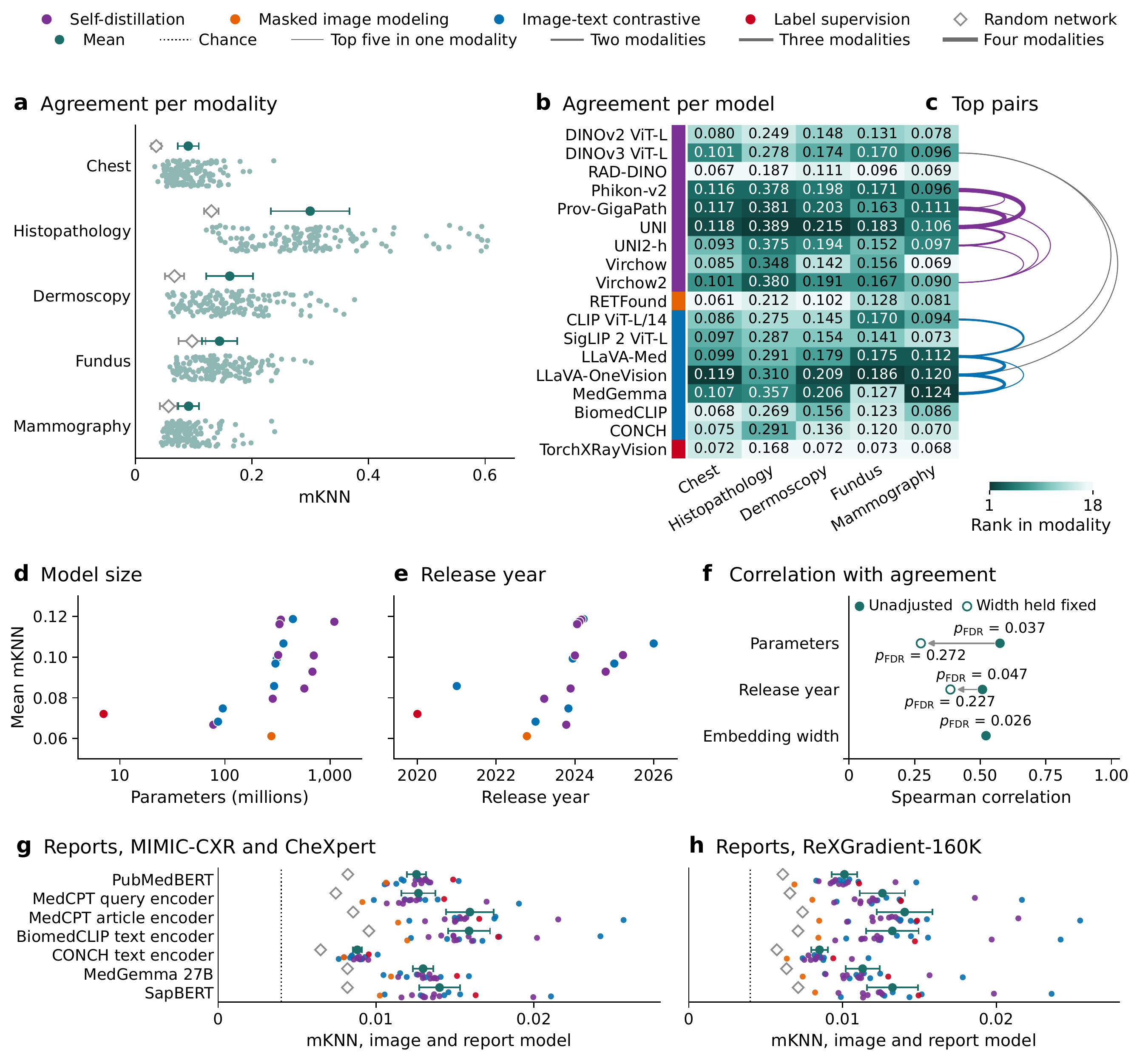}
\caption{Agreement between 18 public medical foundation models in 5 imaging modalities. Color marks the pretraining objective, and a diamond marks a randomly initialized network. In \textbf{a}, \textbf{g}, and \textbf{h}, a teal marker is the mean over pairs, with the 95\% confidence interval of the delete-one-model jackknife (mean $\pm$ 1.96 standard errors). Counts per panel: 153 model pairs and 18 random-network pairs per modality (\textbf{a}), 18 models (\textbf{b}, \textbf{d}--\textbf{f}), 13 model pairs (\textbf{c}), and 126 image-report pairs and 7 random-network pairs per dataset (\textbf{g}, \textbf{h}). \textbf{a}, mKNN of each pair (light dots) on 2{,}500 images per modality at $k = 10$. \textbf{b}, Mean mKNN of each model with the other 17 on each full image set, with $k$ at 0.4\% of the images. Shading marks the rank within a modality. \textbf{c}, Arcs join the pairs of \textbf{b} among the five most aligned in at least one modality. Arc width counts these modalities, and a gray arc joins two models of different objectives. \textbf{d},\textbf{e}, The chest column of \textbf{b} against parameter count and release year. \textbf{f}, Spearman correlations of the chest column of \textbf{b}, unadjusted (filled) and with the log width held fixed (open), with $p_{\mathrm{FDR}}$ within each test family. \textbf{g},\textbf{h}, mKNN between 7 text models on the reports and 18 image models on 2{,}500 radiographs (dots) at $k = 10$, on MIMIC-CXR and CheXpert (\textbf{g}) and ReXGradient-160K (\textbf{h}), with chance as a dotted line. mKNN, mutual $k$-nearest-neighbor overlap; $p_{\mathrm{FDR}}$, $p$-value corrected for the false discovery rate.}
\label{fig:panel}
\end{figure*}

\begin{figure*}[p]
\centering
\includegraphics[width=\textwidth]{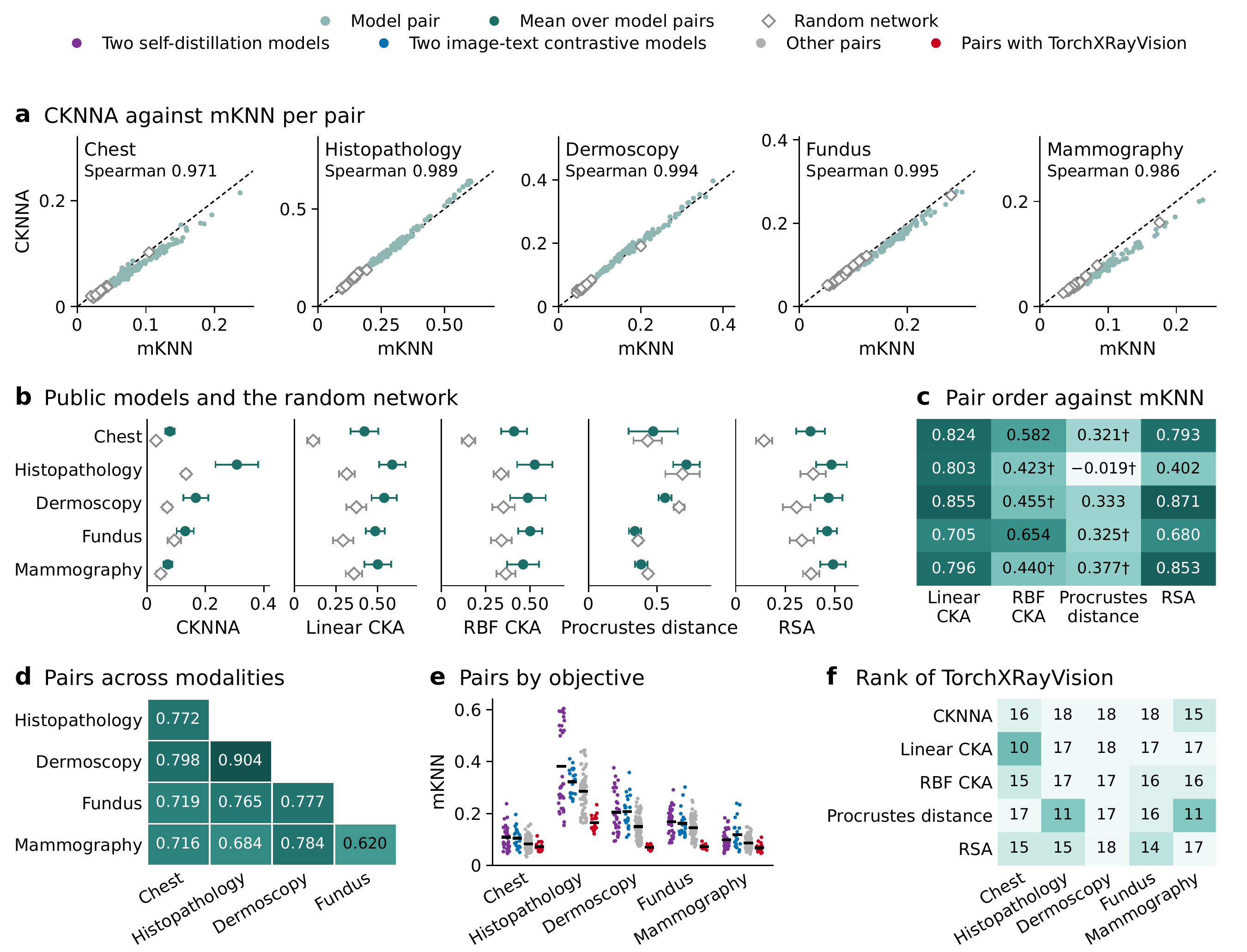}
\caption{Agreement between the 18 public models under 5 further statistics and across the 5 modalities, on 2{,}500 images per modality with $k = 10$ ($n$ = 153 model pairs per modality). Light teal dots are model pairs, and a teal marker is their mean. A diamond marks a pair with the randomly initialized network in \textbf{a} and the mean over its 18 pairs in \textbf{b}. In \textbf{e}, purple and blue dots are pairs of two self-distillation and of two image-text contrastive models, red dots are the pairs with TorchXRayVision, the only label-supervised model, and gray dots are the other pairs. Darker shading marks a larger coefficient in \textbf{c} and \textbf{d} and a rank closer to 1 in \textbf{f}. Only \textbf{b} draws an error bar, which spans the 95\% confidence interval, the mean $\pm$ 1.96 standard errors of the delete-one-model jackknife. Counts per panel: 153 model pairs and 18 random-network pairs per modality (\textbf{a}, \textbf{b}), 36, 21, 79, and 17 pairs per group and modality (\textbf{e}), and 18 models (\textbf{f}). \textbf{a}, CKNNA against mKNN, one facet per modality, with the Spearman correlation over the model pairs and the identity as a dashed line. \textbf{b}, Mean of each statistic over the model pairs and over the random-network pairs. A smaller Procrustes distance means more similar. \textbf{c}, Spearman correlation over the model pairs between mKNN and the statistic of each column, with the Procrustes distance negated. \textbf{d}, Spearman correlation of the mKNN of the model pairs between two modalities. \textbf{e}, mKNN of every pair, with a black line at the mean of each group. \textbf{f}, Rank of TorchXRayVision among the 18 models by its mean over its 17 pairs, 1 being the most similar. The $p$ of a coefficient comes from the delete-one-model jackknife, with the FDR over the 35 coefficients of \textbf{a}, \textbf{c}, and \textbf{d}. A dagger in \textbf{c} marks $p_{\mathrm{FDR}} \geq 0.05$. CKA, centered kernel alignment; CKNNA, centered kernel nearest-neighbor alignment; FDR, false discovery rate; mKNN, mutual $k$-nearest-neighbor overlap; RBF, radial basis function; RSA, representational similarity analysis.}
\label{efig:pairs}
\end{figure*}

Agreement between public models is low (Fig.~\ref{fig:panel}a). On 2,500 chest radiographs at $k = 10$, two models share on average 0.091 $\pm$ 0.009 [0.072, 0.109] of the nearest neighbors of a radiograph, fewer than one of 10. For the random network, the value is 0.036 $\pm$ 0.004 [0.027, 0.044]. In the other modalities, the mean mKNN at the same size ranges from 0.091 on mammograms to 0.300 on histopathology patches. On the full image sets, the pre-specified rule is met in every modality except fundus photographs, where the CIs of the 18 models and of the random network overlap (Supplementary Table~\ref{stab:rules}). Under CKNNA, the CIs also overlap for mammograms (Supplementary Note~\ref{snote:statistics}). The mean mKNN of each modality changes by at most 0.003 between the full image set and 2,500 images (Supplementary Fig.~\ref{sfig:fixedk}b). On 2,500 images, CKNNA orders the pairs almost as mKNN does (Spearman 0.971 to 0.995 over the 153 pairs of a modality, $p_{\mathrm{FDR}} < 0.001$; Fig.~\ref{efig:pairs}a). Linear and radial-basis-function (RBF) CKA, representational similarity analysis (RSA), and the Procrustes distance order the pairs less like mKNN does (Spearman $-0.019$ to 0.871; Fig.~\ref{efig:pairs}c). Under these four statistics, the models agree more than the random network in one to five of the modalities (nonoverlapping CIs; Fig.~\ref{efig:pairs}b; Supplementary Note~\ref{snote:statistics}).

In every modality, the most aligned pair of models shares a training recipe or an initialization (Fig.~\ref{fig:panel}b,c; Supplementary Table~\ref{stab:pair_types}). UNI~\cite{Chen2024UNI} and Phikon-v2~\cite{Filiot2024Phikon} are both histopathology models trained with the DINOv2 recipe. This pair is among the five most aligned pairs in four of the five modalities (mKNN 0.245 on chest radiographs and 0.603 on histopathology patches). In four of the five modalities, at least one pair drawn from CLIP~\cite{Radford2021CLIP} and the vision encoders of LLaVA-Med~\cite{Li2023LLaVAMed}, LLaVA-OneVision~\cite{Li2024LLaVAOneVision}, and MedGemma~\cite{Sellergren2025MedGemma} is also among the five most aligned pairs (mKNN 0.163 to 0.357). The vision encoder of LLaVA-Med has the weights of the CLIP ViT-L/14 encoder for $336 \times 336$ pixels. The agreement of a pair correlates between modalities (Spearman 0.620 to 0.904, $p_{\mathrm{FDR}} < 0.001$; Fig.~\ref{efig:pairs}d). In every modality, pairs of two self-distillation models and pairs of two image-text contrastive models agree more on average than pairs of models with different objectives (Fig.~\ref{efig:pairs}e). TorchXRayVision~\cite{Cohen2022TorchXRayVision} is the only label-supervised model. It is the least aligned model in the four modalities other than chest radiographs, on which it was trained, and the fourth least aligned on chest radiographs (Fig.~\ref{fig:panel}b; Fig.~\ref{efig:pairs}f). The mean agreement of a model with the other 17 on chest radiographs is correlated with its number of parameters (Spearman 0.575, $p_{\mathrm{FDR}} = 0.037$), its release year (0.508, $p_{\mathrm{FDR}} = 0.047$), and its embedding width (0.522, $p_{\mathrm{FDR}} = 0.026$; Fig.~\ref{fig:panel}d--f). The width correlates with the parameter count at 0.902 ($p_{\mathrm{FDR}} < 0.001$). With the width held fixed, neither the parameter count (partial Spearman 0.274, $p_{\mathrm{FDR}} = 0.272$) nor the release year (0.386, $p_{\mathrm{FDR}} = 0.227$) remains significantly correlated with agreement. An image model embedding a radiograph and a text model embedding its report share 0.013 $\pm$ 0.001 [0.011, 0.015] of the nearest neighbors on MIMIC-CXR and CheXpert and 0.012 $\pm$ 0.001 [0.010, 0.014] on ReXGradient-160K (Fig.~\ref{fig:panel}g,h). For the random network, the values are 0.008 and 0.007.

\subsection*{Label supervision yields the least convergent models}

\begin{figure*}[p]
\centering
\includegraphics[width=\textwidth]{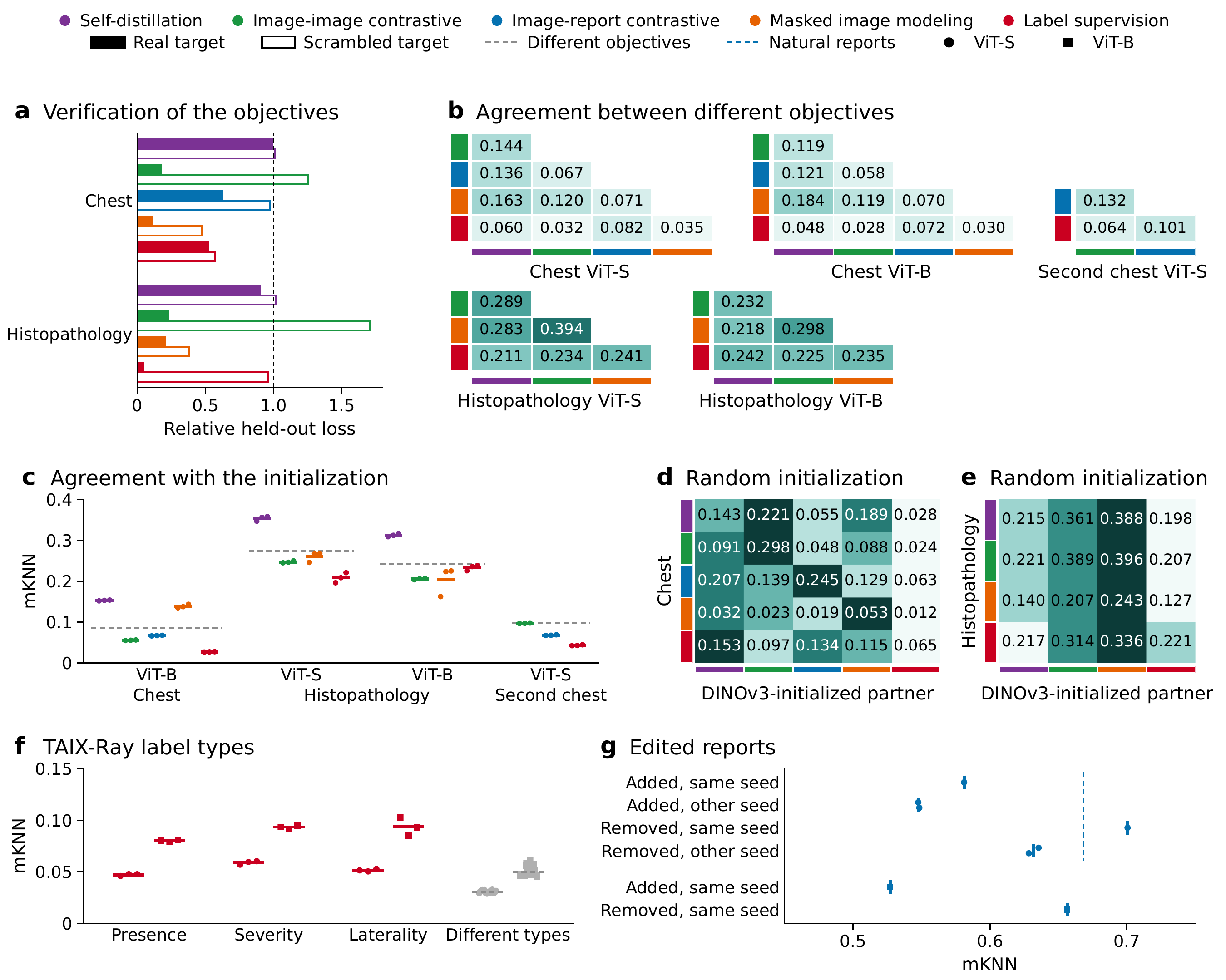}
\caption{Verification, agreement between objectives, initialization, label type, and report content in the controlled study, on 2{,}500 held-out images per setting at $k = 10$ ($n$ = 95 trained models and 18 verification models). Color marks the pretraining objective, and colored cells beside and below a matrix mark the objectives of its rows and columns. Circles mark ViT-S and squares ViT-B models. A dot is one model or pair, and a short line marks their mean. No panel draws an error bar. Counts per panel: 18 verification models, 1 per bar (\textbf{a}), 9 pairs of models per cell (\textbf{b}), 3 models per objective and setting (\textbf{c}), 3 pairs per cell (\textbf{d}, \textbf{e}), 3 seed pairs per label type and 27 pairs of different label types per size (\textbf{f}), and 1 or 2 pairs per row (\textbf{g}). \textbf{a}, Held-out loss of a model trained for 300 steps on the real target (filled) or on a scrambled target (open), relative to the untrained DINOv3 initialization (dashed line at 1). \textbf{b}, Mean mKNN between models of two different objectives per setting, shaded from 0 to 0.5. \textbf{c}, mKNN of each model with its DINOv3 initialization, in the settings of Fig.~\ref{fig:matrix}a other than chest ViT-S. The dashed line is the mean over pairs of different objectives. \textbf{d},\textbf{e}, mKNN of the randomly initialized ViT-S model of each objective (row) with the DINOv3-initialized models of each objective (column), on chest radiographs (\textbf{d}) and histopathology patches (\textbf{e}), shaded by rank in the row. \textbf{f}, mKNN between two TAIX-Ray models of one label type and different seeds, and between models of different label types (gray). \textbf{g}, mKNN between an image-report model of the second chest setting trained on edited reports, with a demographic sentence added or the demographic words removed, and a natural-report model of the same or another seed. The dashed line is the mean of the three natural-report ViT-S seed pairs. mKNN, mutual $k$-nearest-neighbor overlap; ViT-S and ViT-B, small and base vision transformers.}
\label{efig:start}
\end{figure*}

To identify the cause of convergence, we trained models that differ in one factor at a time (Table~\ref{tab:design}). The five pretraining objectives were each trained on the same 121,135 MIMIC-CXR training radiographs~\cite{Johnson2019MIMICCXR}, with label supervision on the 13 CheXpert findings other than no finding~\cite{Irvin2019CheXpert}. Each objective was trained at 2 model sizes, the small and the base ViT (ViT-S and ViT-B), from the same DINOv3 initialization~\cite{Simenoni2025DINOv3} for 20,000 optimizer steps with 3 random seeds. Histopathology models were trained in the same way on 16,784 NCT-CRC patches~\cite{Kather2018NCT}, without the report objective and for 19,712 optimizer steps. Agreement was measured on 2,500 held-out images of patients outside every training set. In a check of each objective, a model trained for 300 steps on its real target had a lower held-out loss than a model trained on a scrambled target (paired $t$ test, $p < 0.001$ for each objective; Fig.~\ref{efig:start}a). With a second chest setting of 16 models, we tested whether the effect of the objective depends on the training recipe. Its models were trained with label supervision and the two contrastive objectives on 180,300 MIMIC-CXR radiographs. The training schedule, the augmentation, the image-image contrastive loss, and the text encoder differ from those of the setting above. Within this setting, the label-supervised models had as many optimizer steps as the image-image contrastive models and the same augmentation as the image-report models.

\begin{table}[p]
\centering
\caption{Design of the controlled study. Each group heading names a setting with its training images and its budget. Every model uses the AdamW optimizer, the learning rate of $10^{-4}$, and the weight decay of 0.05. Outside the second chest setting, every model also uses the warm-up over the first 10\% of steps and the effective batch of 512 images. Training on the first set of patients stopped at 19{,}950 steps. Seeds is the number of seeds per size, 3 at ViT-S and 1 at ViT-B where two numbers are given. The second chest setting has a different training schedule, augmentation, image-image contrastive loss, and text encoder. Its image-report models were updated after every batch of 256 radiographs. ViT, vision transformer.}
\label{tab:design}
{\footnotesize\hyphenpenalty=10000\exhyphenpenalty=10000
\setlength{\tabcolsep}{6pt}
\renewcommand{\arraystretch}{1.08}
\begin{tabular*}{\textwidth}{@{\extracolsep{\fill}}>{\raggedright\arraybackslash}p{0.42\textwidth}lllr@{}}
\toprule
Objective & Model sizes & Initialization & Seeds & Models \\
\midrule
\multicolumn{5}{@{}l}{\textbf{Chest radiographs, 121{,}135 MIMIC-CXR radiographs, 20{,}000 steps}} \\
\midrule
Masked image modeling & ViT-S, ViT-B & DINOv3 & 3 & 6 \\
Self-distillation & ViT-S, ViT-B & DINOv3 & 3 & 6 \\
Image-image contrastive & ViT-S, ViT-B & DINOv3 & 3 & 6 \\
Image-report contrastive & ViT-S, ViT-B & DINOv3 & 3 & 6 \\
Label supervision, 13 findings & ViT-S, ViT-B & DINOv3 & 3 & 6 \\
Each of the 5 objectives & ViT-S & Random weights & 1 & 5 \\
\midrule
\multicolumn{5}{@{}l}{\textbf{Chest radiographs, two disjoint sets of training patients, 20{,}000 steps}} \\
\midrule
Image-image contrastive, first set of patients & ViT-S & DINOv3 & 2 & 2 \\
Image-image contrastive, second set of patients & ViT-S & DINOv3 & 2 & 2 \\
\midrule
\multicolumn{5}{@{}l}{\textbf{Histopathology, 16{,}784 NCT-CRC patches, 19{,}712 steps}} \\
\midrule
Masked image modeling & ViT-S, ViT-B & DINOv3 & 3 & 6 \\
Self-distillation & ViT-S, ViT-B & DINOv3 & 3 & 6 \\
Image-image contrastive & ViT-S, ViT-B & DINOv3 & 3 & 6 \\
Label supervision, 9 tissue classes & ViT-S, ViT-B & DINOv3 & 3 & 6 \\
Each of the 4 objectives & ViT-S & Random weights & 1 & 4 \\
\midrule
\multicolumn{5}{@{}l}{\textbf{TAIX-Ray, 31{,}942 radiographs, 20{,}000 steps}} \\
\midrule
Label supervision, presence labels & ViT-S, ViT-B & DINOv3 & 3 & 6 \\
Label supervision, severity labels & ViT-S, ViT-B & DINOv3 & 3 & 6 \\
Label supervision, laterality labels & ViT-S, ViT-B & DINOv3 & 3 & 6 \\
\midrule
\multicolumn{5}{@{}l}{\textbf{Second chest setting, 180{,}300 MIMIC-CXR radiographs with their reports, 50 epochs}} \\
\midrule
Image-image contrastive & ViT-S, ViT-B & DINOv3 & 3 and 1 & 4 \\
Image-report contrastive & ViT-S, ViT-B & DINOv3 & 3 and 1 & 4 \\
Label supervision, 13 findings & ViT-S, ViT-B & DINOv3 & 3 and 1 & 4 \\
Image-report contrastive, demographic sentence added & ViT-S, ViT-B & DINOv3 & 1 & 2 \\
Image-report contrastive, demographic words removed & ViT-S, ViT-B & DINOv3 & 1 & 2 \\
\midrule
All settings & & & & 101 \\
\bottomrule
\end{tabular*}}
\end{table}

\begin{figure*}[p]
\centering
\includegraphics[width=\textwidth]{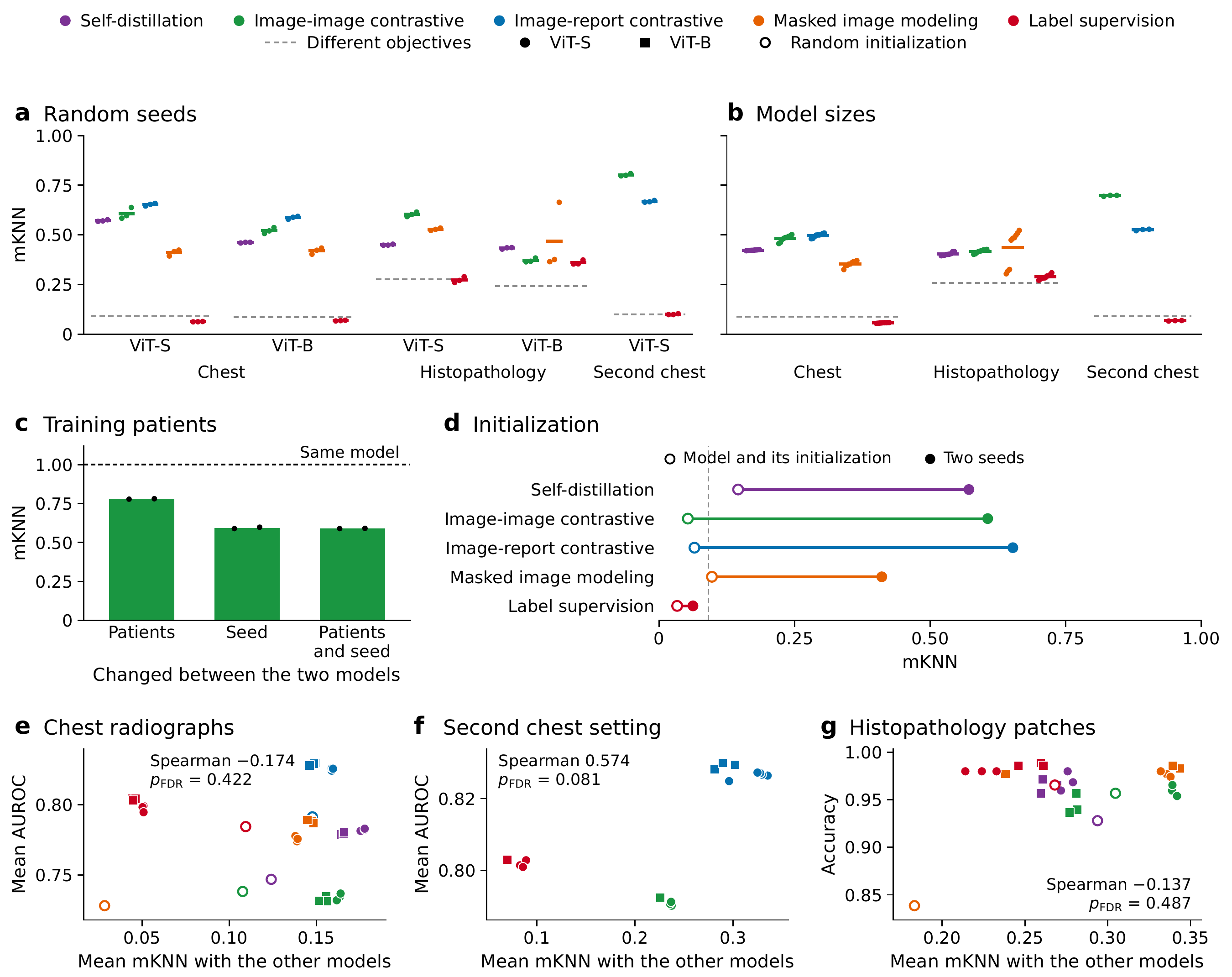}
\caption{Agreement and accuracy by pretraining objective for 83 models of the controlled study (Table~\ref{tab:design}). Color marks the pretraining objective, and agreement is the mKNN on 2{,}500 held-out images at $k = 10$. A horizontal line or a bar is the mean over the pairs drawn as dots. No panel draws an error bar. Counts per panel: 3 seed pairs per objective and setting (\textbf{a}), 9 size pairs per objective and 3 in the second chest setting (\textbf{b}), 2 model pairs per type (\textbf{c}), 3 models and 3 seed pairs per objective (\textbf{d}), and 35, 16, and 28 models (\textbf{e}--\textbf{g}). \textbf{a}, Agreement between two models that differ only in their random seed, in the chest and histopathology settings at ViT-S and ViT-B and in the second chest setting at ViT-S. The dashed line is the mean over pairs of models of different objectives. \textbf{b}, Agreement between the ViT-S and ViT-B models of one objective. \textbf{c}, Agreement between two image-image contrastive models at ViT-S that differ in their training patients, in their seed, or in the patients and the seed. Each of these models was trained on one of two disjoint sets of MIMIC-CXR training patients. \textbf{d}, Agreement of a chest ViT-S model with its DINOv3 initialization (open markers) and between two seeds of its objective (filled markers). \textbf{e}--\textbf{g}, Mean agreement of each model with the other models of its setting against its accuracy, the mean AUROC over the 14 findings (\textbf{e}, \textbf{f}) or the nine-class accuracy (\textbf{g}), with Spearman correlations corrected for the false discovery rate over the four settings. Open markers are randomly initialized models. Four image-report models in \textbf{f} were trained on reports with demographic information added or removed. mKNN, mutual $k$-nearest-neighbor overlap; AUROC, area under the receiver operating characteristic curve; $p_{\mathrm{FDR}}$, $p$-value corrected for the false discovery rate; ViT-S and ViT-B, small and base vision transformers.}
\label{fig:matrix}
\end{figure*}

Two models that differ only in their random seed agree least under label supervision in every setting (Fig.~\ref{fig:matrix}a). On chest radiographs at ViT-S, the seed-pair mKNN is 0.062 under label supervision. Two label-supervised models therefore share fewer than one of the 10 nearest neighbors of an image. The seed-pair mKNN is 0.411 under masked image modeling, 0.571 under self-distillation, 0.606 under image-image contrastive learning, and 0.653 under image-report contrastive learning. Two models trained with one of these objectives share four to seven of the 10 nearest neighbors. Two models of different objectives agree at 0.091 on average (Fig.~\ref{efig:start}b; Supplementary Note~\ref{snote:controlled}). At ViT-B, the seed-pair mKNN is 0.068 under label supervision and 0.419 to 0.587 under the other objectives. In the second chest setting at ViT-S, it is 0.100 under label supervision, 0.668 under image-report contrastive learning, and 0.802 under image-image contrastive learning. On histopathology patches at ViT-S, it is 0.273 under label supervision and 0.450 to 0.603 under the other three objectives. The pre-specified rule is met at both sizes on chest radiographs, at ViT-S in the second chest setting, and at ViT-S on histopathology patches, for mKNN and for CKNNA (Supplementary Table~\ref{stab:rules}). At ViT-B on histopathology patches, the seed-pair mKNN is 0.360 under label supervision and 0.371 to 0.468 under the other objectives. The seed pairs of label supervision (0.353 to 0.373) overlap those of image-image contrastive learning (0.364 to 0.383). The rule is not met in this setting, because one masked-image-modeling model agrees with the other two at 0.365 and 0.376, while those two agree with each other at 0.664. The largest spread of the seed pairs inside one objective (0.299) therefore exceeds the spread between the objective means (0.108).

Between the two model sizes of one objective, label supervision also agrees least, in chest radiography, in histopathology, and in the second chest setting (Fig.~\ref{fig:matrix}b). On chest radiographs, the ViT-S and ViT-B models of one objective agree at 0.354 to 0.497 under the other four objectives and at 0.057 under label supervision (Supplementary Note~\ref{snote:controlled}). For image-image contrastive models, the training patients affect agreement less than the random seed (Fig.~\ref{fig:matrix}c). Two image-image contrastive models trained with the same seed on two disjoint sets of MIMIC-CXR training patients agree at 0.780. Two models trained on the same set of patients with different seeds agree at 0.594. Two models that differ in both the patients and the seed agree at 0.590.

In every setting, a model agrees less with its DINOv3 initialization than two models of its objective agree with each other (Fig.~\ref{fig:matrix}d; Fig.~\ref{efig:start}c). On chest radiographs at ViT-S, a model agrees with its DINOv3 initialization at 0.033 under label supervision, at 0.053 under image-image and 0.065 under image-report contrastive learning, at 0.097 under masked image modeling, and at 0.146 under self-distillation. Two models of different objectives agree at 0.091. Two models of one objective other than label supervision agree with each other far more, at 0.411 to 0.653. We also trained one model per objective from random initialization. Under both contrastive objectives and under masked image modeling, this model agrees more with the DINOv3-initialized models of its own objective than with those of the other objectives. Under self-distillation, the difference is within one standard error of zero on chest radiographs and negative on histopathology patches. Under label supervision, it is negative in both modalities (Fig.~\ref{efig:start}d,e; Supplementary Table~\ref{stab:cross_init}). Pooled over the objectives, this difference is not significant ($p_{\mathrm{FDR}} = 0.531$ on chest radiographs and 0.908 on histopathology patches). The comparison is descriptive, since each randomly initialized model has a single seed and the architecture of the standard ViT, without the register tokens, rotary position embeddings, and layer scale of DINOv3. Except under image-image contrastive learning on chest radiographs, the randomly initialized model is also less accurate than the mean of the DINOv3-initialized models of its objective (Fig.~\ref{fig:matrix}e,g). Models trained with label supervision on the presence, severity, or laterality labels of TAIX-Ray agree between seeds at 0.047 to 0.094 (Fig.~\ref{efig:start}f). In the second chest setting, editing the demographic content of the reports changes an image-report model about as much as a new seed does (Fig.~\ref{efig:start}g).

Convergence and accuracy are different properties (Fig.~\ref{fig:matrix}e--g; Supplementary Table~\ref{stab:utility}). On chest radiographs at ViT-S, label supervision is the second most accurate objective (mean AUROC over the 14 findings 0.797) and the least convergent. Image-report contrastive learning is the most accurate (0.825) and the most convergent. Image-image contrastive learning is the least accurate (0.734) and the second most convergent. In histopathology, label supervision is among the most accurate objectives at both sizes (nine-class accuracy 0.980 and 0.987). The mean agreement of a model with the other models of its setting is not significantly correlated with its accuracy among the 35 chest models (Spearman $-0.174$, $p_{\mathrm{FDR}} = 0.422$), the 28 histopathology models ($-0.137$, $p_{\mathrm{FDR}} = 0.487$), the 18 TAIX-Ray models ($-0.476$, $p_{\mathrm{FDR}} = 0.092$), or the 16 models of the second chest setting (0.574, $p_{\mathrm{FDR}} = 0.081$).

\subsection*{Classifier transfer between public models through a linear mapping fitted on unlabeled radiographs}

\begin{figure*}[p]
\centering
\includegraphics[width=\textwidth]{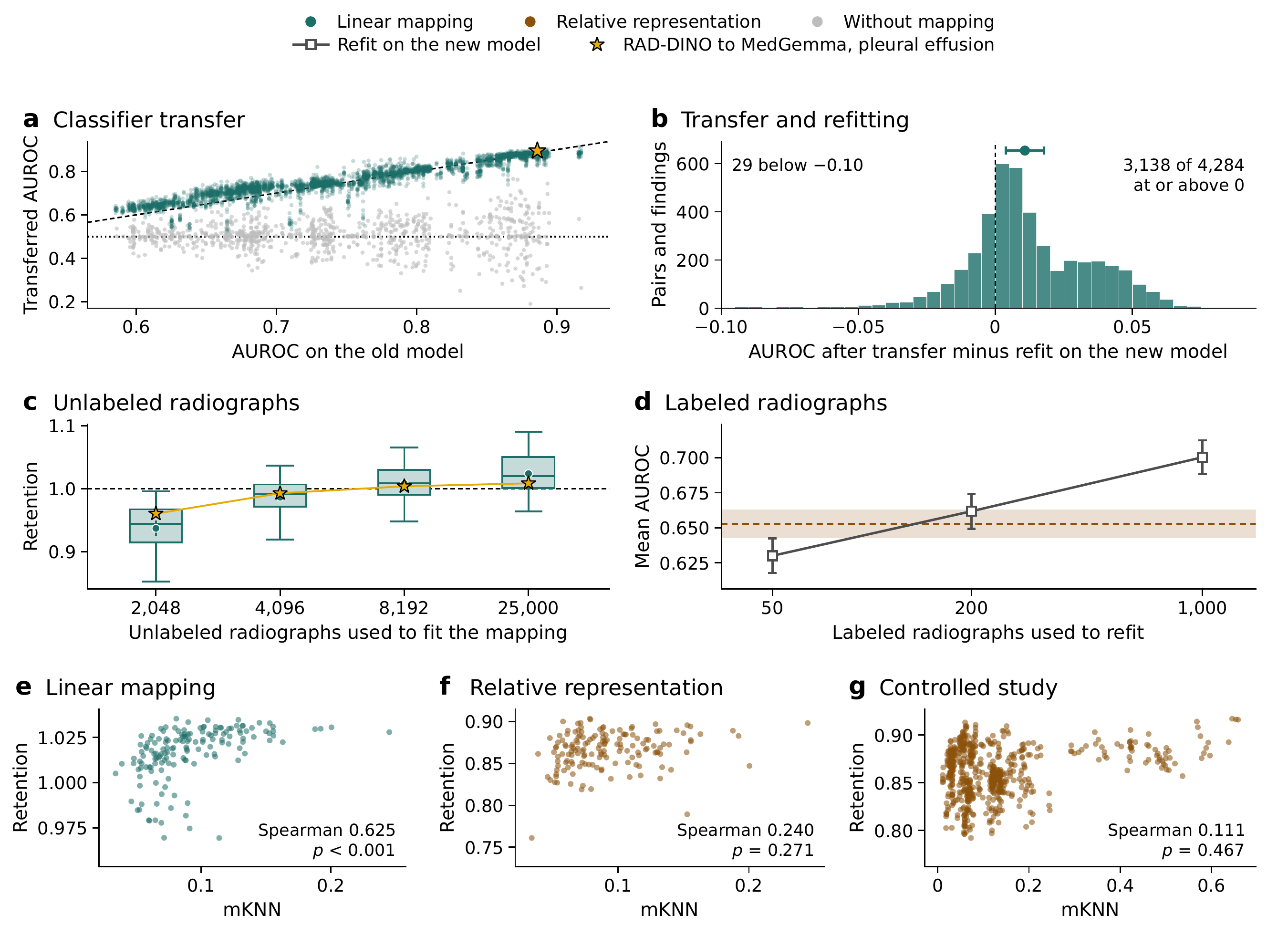}
\caption{Classifier transfer between the 18 public models on chest radiographs ($n$ = 306 ordered model pairs and 14 findings, 4{,}284 combinations). Retention is the AUROC after transfer to a new model divided by the AUROC on the old model, the model used to fit the classifier. For the relative representation (\textbf{f}, \textbf{g}), the denominator is the AUROC of a classifier fitted on the relative representation of the new model. Color marks the transfer route. In \textbf{a} and \textbf{e}--\textbf{g}, a dot is one combination or one model pair. In \textbf{b}--\textbf{d}, a marker with an error bar, or a dashed line inside a band, is the mean over the combinations. The error bar and the band span the 95\% confidence interval, the mean $\pm$ 1.96 standard errors of the delete-one-model jackknife. In \textbf{c}, a box spans the quartiles around the median, and its whiskers span the 5th to the 95th percentile. Counts per panel: 4{,}284 combinations (\textbf{a}--\textbf{d}) and 924 more without a mapping (\textbf{a}), 153 model pairs (\textbf{e}, \textbf{f}), and 595 model pairs (\textbf{g}). \textbf{a}, AUROC on the old model against AUROC after transfer through a linear mapping fitted on up to 25{,}000 unlabeled radiographs, or without a mapping for the 66 ordered pairs of equal width. The dashed line is the identity, and the dotted line is chance. \textbf{b}, AUROC after transfer minus that of a refit on the new model with up to 25{,}000 labels. \textbf{c}, Retention by the number of unlabeled radiographs used to fit the mapping. \textbf{d}, AUROC through the relative representation over 1{,}024 anchor radiographs (band) against refits with 50, 200, and 1{,}000 labels, averaged over 20 draws. \textbf{e}--\textbf{g}, Retention of a model pair, averaged over both directions and the findings, against its mKNN, for the linear mapping (\textbf{e}) and the relative representation (\textbf{f}) over the 153 pairs of public models, and for the relative representation over the 595 pairs of the 35 controlled chest models on 13 findings (\textbf{g}). The $p$-values of the Spearman correlations are from the delete-one-model jackknife. The star marks a pleural effusion classifier transferred from RAD-DINO to MedGemma. AUROC, area under the receiver operating characteristic curve; mKNN, mutual $k$-nearest-neighbor overlap.}
\label{fig:transfer}
\end{figure*}

We tested classifier transfer without new labels on chest radiographs, for the 14 findings and the 306 ordered pairs of the 18 public models, on the training and test partitions of the chest pool (Fig.~\ref{fig:transfer}a). A linear mapping from the embedding of the new model to the embedding of the old model is fitted by least squares on unlabeled training radiographs embedded by both models. The classifier of the old model is then applied to the mapped embeddings. Retention is the AUROC of the transferred classifier divided by the AUROC of the same classifier on the old model.

\begin{figure*}[p]
\centering
\includegraphics[width=\textwidth]{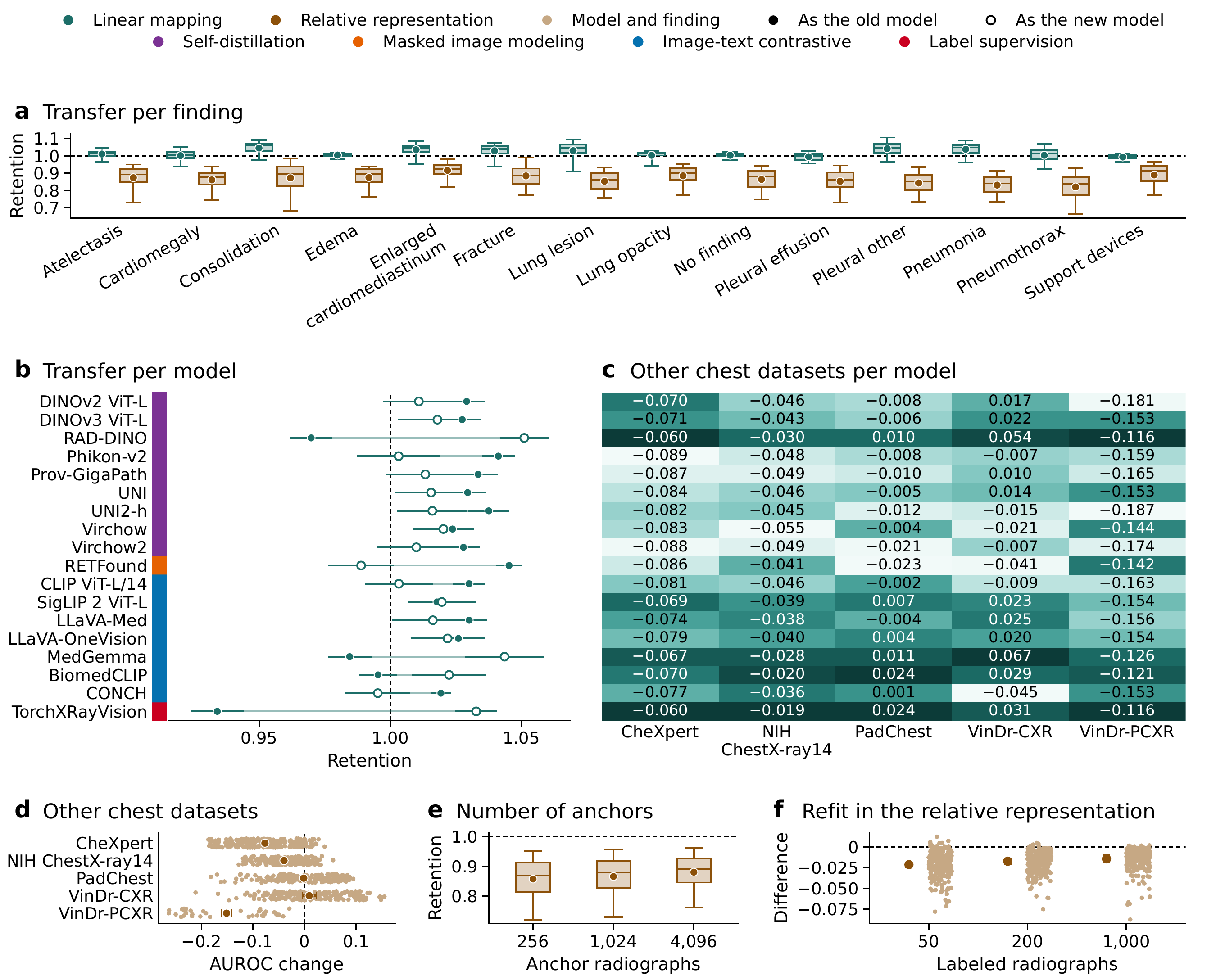}
\caption{Classifier transfer between the 18 public models on chest radiographs by finding, model, dataset, and number of anchors or labels ($n$ = 306 ordered model pairs and 14 findings). Retention is the AUROC after transfer divided by that on the old model for the linear mapping and by that of a classifier fitted on the relative representation of the new model for the relative representation. Color marks the transfer route, and the strip in \textbf{b} marks the pretraining objective. In \textbf{a}, \textbf{b}, and \textbf{d}--\textbf{f}, a dot with an error bar is a mean. The error bar spans the 95\% confidence interval, the mean $\pm$ 1.96 standard errors of the delete-one-model jackknife. A box spans the quartiles around the median, and its whiskers span the 5th to the 95th percentile. Counts per panel: 306 ordered model pairs per box (\textbf{a}), 17 other models and 14 findings per dot (\textbf{b}), 2 to 14 findings per cell (\textbf{c}), 36 to 252 dots per dataset (\textbf{d}), 4{,}284 combinations per box (\textbf{e}), and 252 dots per label count (\textbf{f}). \textbf{a}, Retention per finding, with 1{,}024 anchor radiographs for the relative representation. \textbf{b}, Retention through the linear mapping per model. \textbf{c}, AUROC on every labeled radiograph of another chest dataset minus AUROC on the MIMIC-CXR test radiographs, for a classifier fitted on MIMIC-CXR in the relative representation and averaged over the findings with at least 20 positive radiographs in that dataset. Shading marks the rank of a model in its dataset. \textbf{d}, The difference of \textbf{c} per model and finding. \textbf{e}, Retention through the relative representation by the number of anchors. \textbf{f}, AUROC of a refit on the relative representation of the new model minus that on its embedding, each averaged over 20 draws of labeled radiographs. Dashed lines mark a retention of 1 and a difference of 0. AUROC, area under the receiver operating characteristic curve.}
\label{efig:transfer}
\end{figure*}

For example, a pleural effusion classifier fitted on RAD-DINO~\cite{PerezGarcia2025RadDino} embeddings has an AUROC of 0.886, and of 0.894 after transfer to MedGemma. A classifier fitted directly on MedGemma embeddings with up to 25,000 labeled radiographs has an AUROC of 0.891. Over all pairs and findings, retention is 1.017 $\pm$ 0.005 [1.007, 1.027]. Mean retention is 0.993 to 1.045 per finding, 0.934 to 1.045 per old model, and 0.989 to 1.051 per new model (Fig.~\ref{efig:transfer}a,b; Supplementary Note~\ref{snote:transfer}). The transferred classifier scores 0.011 $\pm$ 0.004 [0.004, 0.018] higher than a classifier fitted directly on the new model, and at least as high in 3,138 of the 4,284 combinations of a model pair and a finding (Fig.~\ref{fig:transfer}b). Without the linear mapping, the old classifier can be applied to the new embedding only for the 66 ordered pairs of equal embedding width. Its AUROC is then 0.506 $\pm$ 0.005 [0.496, 0.515], the chance level. Retention increases with the number of unlabeled radiographs used to fit the mapping (Fig.~\ref{fig:transfer}c). It is 0.937 with 2,048 radiographs, 0.987 $\pm$ 0.003 [0.980, 0.993] with 4,096, 1.008 with 8,192, and 1.024 with 25,000.

The second transfer route is the anchor-based relative representation~\cite{Moschella2023Relative}. Each radiograph is represented by its cosine similarities to 1,024 anchor radiographs embedded by the same model (Fig.~\ref{fig:transfer}d; Supplementary Table~\ref{stab:anchor_route}). It retains 0.866 $\pm$ 0.006 [0.853, 0.878] of the AUROC of a classifier fitted on the relative representation of the new model. Per finding, it retains 0.820 to 0.916 (Fig.~\ref{efig:transfer}a). With 256 and 4,096 anchors, it retains 0.857 and 0.880 (Fig.~\ref{efig:transfer}e). This route outperforms refitting on the embeddings of the new model only below about 200 labels (Fig.~\ref{fig:transfer}d). Its AUROC is 0.023 $\pm$ 0.004 [0.016, 0.030] higher than that of a refit with 50 labels and 0.009 $\pm$ 0.004 [0.001, 0.017] lower than that of a refit with 200 labels. A classifier fitted on MIMIC-CXR in the relative representation loses 0.077 in AUROC on CheXpert, 0.040 on NIH ChestX-ray14, and 0.151 on VinDr-PCXR (Fig.~\ref{efig:transfer}c,d; Supplementary Table~\ref{stab:cross_site}). On PadChest and VinDr-CXR~\cite{Nguyen2022VinDrCXR}, the changes of $-0.001$ and 0.009 are not significant ($p_{\mathrm{FDR}} = 0.660$ and 0.227).

Retention through the linear mapping is higher between models that agree more (Spearman 0.625 over the 153 model pairs, jackknife $p < 0.001$; Fig.~\ref{fig:transfer}e; Supplementary Table~\ref{stab:link}). This analysis was added after the rules were fixed. The pre-specified rule concerned the relative representation and is not met (Spearman 0.240, jackknife $p = 0.271$, $p_{\mathrm{FDR}} = 0.406$; Fig.~\ref{fig:transfer}f; Supplementary Table~\ref{stab:rules}). Among the 35 chest models of the controlled study, retention through the relative representation is not significantly associated with agreement (Spearman 0.111 over 595 pairs, jackknife $p = 0.467$; Fig.~\ref{fig:transfer}g). In the controlled study, we also computed the retention of the AUROC above 0.5, the chance-corrected retention. It is 0.597 to 0.629 when the new model is trained with label supervision, image-report contrastive learning, or self-distillation, and 0.408 and 0.422 when it is trained with image-image contrastive learning or masked image modeling. Over the pairs of models with one objective, it is 0.582, and over the pairs with two objectives, it is 0.526 (Supplementary Note~\ref{snote:transfer}).

\subsection*{Severity and laterality classification after classifier transfer on TAIX-Ray}

Radiology reports record the severity and the laterality of a finding as well as its presence. In a review of 1,065,322 radiology reports, most of the laterality errors found were graded as clinically important~\cite{Sangwaiya2009Laterality}. We therefore tested classifier transfer for severity and laterality on 50,000 TAIX-Ray radiographs of 11,475 patients~\cite{Truhn2026TAIXRay}, whose structured reports record the presence, a severity grade from 0 to 4, and the laterality of each finding.

\begin{figure*}[p]
\centering
\includegraphics[width=\textwidth]{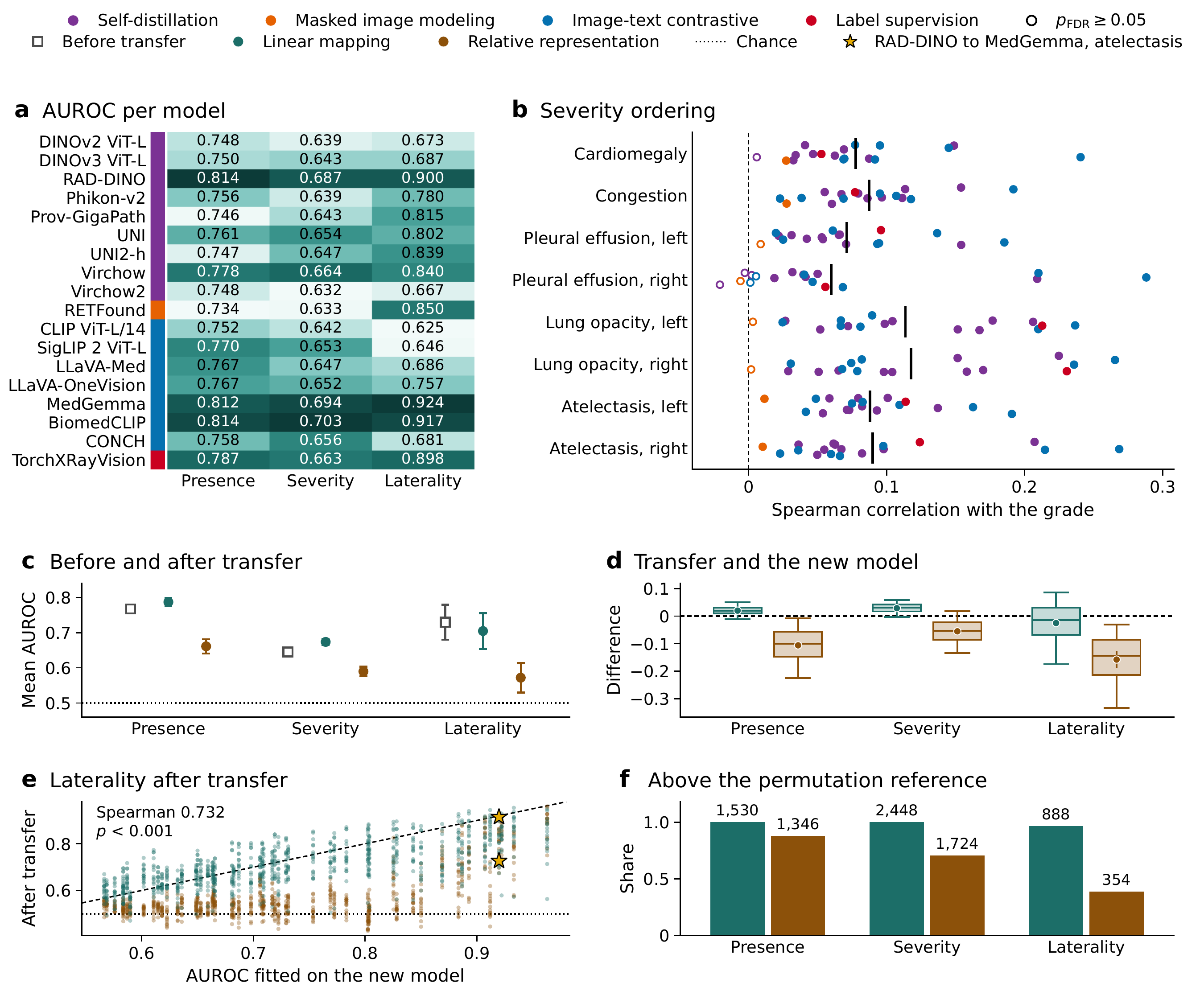}
\caption{Severity and laterality on TAIX-Ray before and after transfer between the 18 public models ($n$ = 50{,}000 radiographs of 11{,}475 patients). Color marks the pretraining objective in \textbf{a} and \textbf{b} and the route in \textbf{c}--\textbf{f}. In \textbf{c} and \textbf{d}, a marker is the mean over the combinations of an ordered model pair and a target, with the 95\% confidence interval of the delete-one-model jackknife (mean $\pm$ 1.96 standard errors). In \textbf{d}, a box spans the quartiles around the median, and its whiskers span the 5th to the 95th percentile. Counts per panel: 18 models (\textbf{a}), 144 combinations of a model and a graded item (\textbf{b}), and 1{,}530 presence, 2{,}448 severity, and 918 laterality combinations per route (\textbf{c}--\textbf{f}). \textbf{a}, Mean AUROC of each model over five presence findings, eight graded items, and three laterality findings. Shading marks the rank within a column. \textbf{b}, Spearman correlation between the grade and the cosine distance from the grade-0 centroid over the radiographs with a grade of 1 to 4, per model and graded item (dots, open where $p_{\mathrm{FDR}} \geq 0.05$), with the mean over models as a vertical line. \textbf{c}, Mean AUROC before transfer (squares) and after transfer (circles). \textbf{d}, AUROC after transfer minus that of a classifier fitted on the new model. \textbf{e}, Laterality AUROC after transfer against that of the classifier fitted on the new model, with the identity as a dashed line. Stars mark the atelectasis laterality classifier transferred from RAD-DINO to MedGemma through each route. The Spearman correlation, with its delete-one-model jackknife $p$, is for the linear mapping. \textbf{f}, Share of combinations above the 97.5th percentile of 20 label permutations, with the count above each bar. In \textbf{c} and \textbf{e}, the dotted line is chance. AUROC, area under the receiver operating characteristic curve; $p_{\mathrm{FDR}}$, $p$-value corrected for the false discovery rate over the 144 tests.}
\label{fig:detail}
\end{figure*}

On the embeddings of each of the 18 models, linear classifiers have a mean AUROC of 0.767 for presence, 0.655 for severity, and 0.777 for laterality (Fig.~\ref{fig:detail}a). Presence is present against absent for five findings, severity is grade 2 or above against grade 1 for eight graded items, and laterality is right-only against left-only at matched grade for three findings. Each of the 288 classifiers exceeds its label-permutation reference (Supplementary Table~\ref{stab:rules}). Laterality AUROC is 0.898 to 0.924 for RAD-DINO, TorchXRayVision, MedGemma, and BiomedCLIP. The model cards of RAD-DINO, TorchXRayVision, and MedGemma name chest radiograph datasets of this study as pretraining data. BiomedCLIP was pretrained on figures from biomedical articles. It is 0.625 to 0.687 for the four general-purpose models pretrained on natural images. For a radiograph with a grade of 1 to 4, the cosine distance from the centroid of the grade-0 radiographs correlates positively with the grade at $p_{\mathrm{FDR}} < 0.05$ in 134 of the 144 combinations of a model and a graded item (mean Spearman 0.088; Fig.~\ref{fig:detail}b).

Classifiers transferred through the linear mapping retain presence, severity, and most of laterality (Fig.~\ref{fig:detail}c). The mean AUROC is 0.787 after transfer against 0.767 before transfer for presence, 0.674 against 0.645 for severity, and 0.705 against 0.730 for laterality. The transferred classifier scores at least as high as a classifier fitted on the new model in 1,353 of the 1,530 presence, 2,269 of the 2,448 severity, and 386 of the 918 laterality combinations of an ordered model pair and a target (Fig.~\ref{fig:detail}d). Classifiers transferred through the relative representation keep little of their laterality performance. Their mean laterality AUROC is 0.680 before transfer and 0.572 $\pm$ 0.022 [0.530, 0.614] after transfer, where the chance level is 0.5. It is at least as high as that of a classifier fitted on the new model in 4 of the 918 combinations. For presence and severity, their mean AUROC is 0.661 and 0.590. For example, a laterality classifier for atelectasis fitted on RAD-DINO embeddings has an AUROC of 0.893, and of 0.915 after transfer to MedGemma through the linear mapping. Fitted on the relative representation of RAD-DINO, the classifier has an AUROC of 0.930, and of 0.727 after transfer to MedGemma (Fig.~\ref{fig:detail}e). The transferred laterality classifiers exceed their label-permutation reference in 888 of the 918 combinations through the linear mapping and in 354 through the relative representation (Fig.~\ref{fig:detail}f; Supplementary Note~\ref{snote:transfer}).

\subsection*{Agreement depends on the source dataset and radiographic view more than on the findings}

\begin{figure*}[p]
\centering
\includegraphics[width=\textwidth]{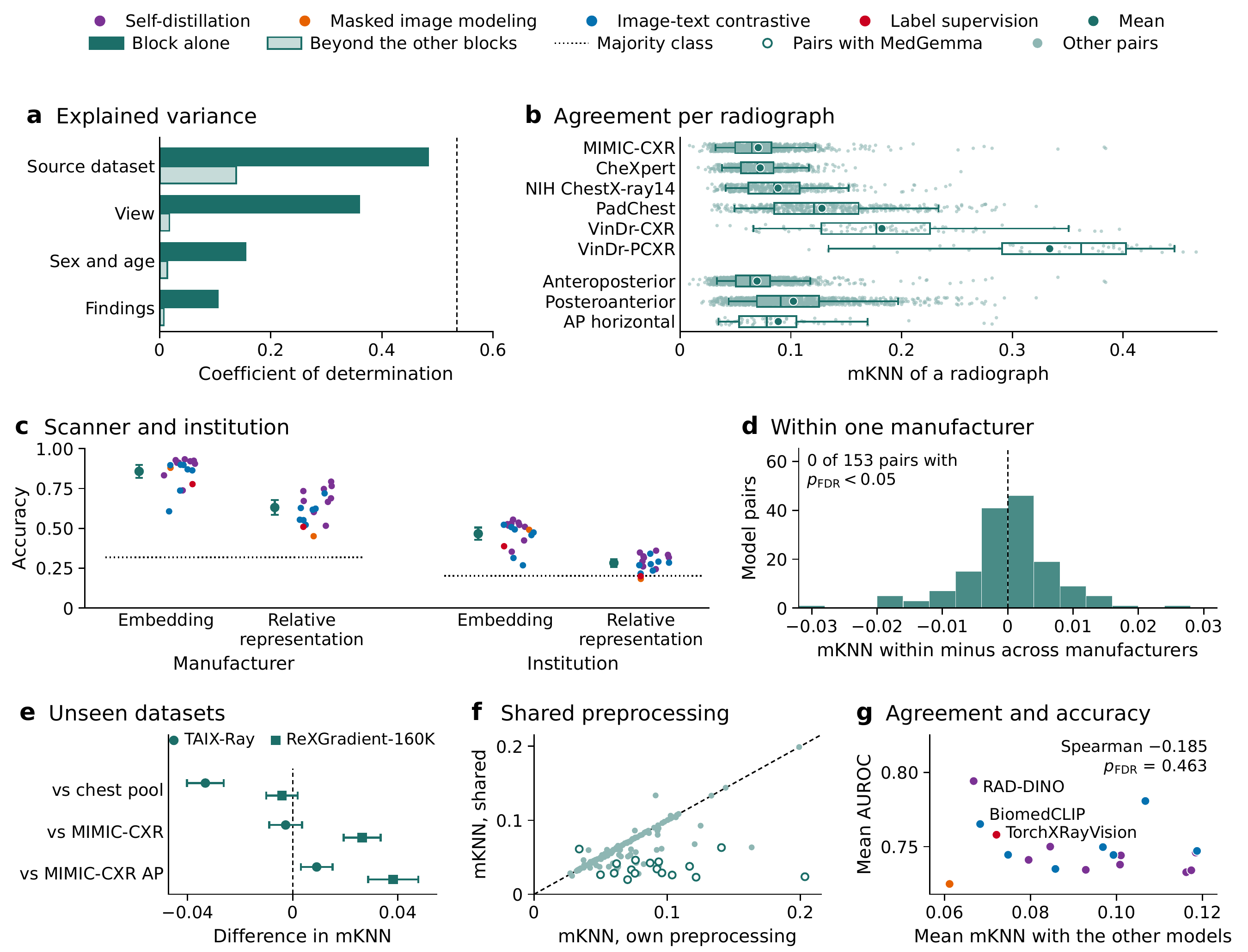}
\caption{Agreement between the 18 public models on chest radiographs against source dataset, view, scanner, pretraining exposure, preprocessing, and accuracy. Color marks the pretraining objective. In \textbf{b}, \textbf{c}, and \textbf{e}, a teal marker is a mean. In \textbf{c} and \textbf{e}, its error bar is the 95\% confidence interval of the delete-one-model jackknife (mean $\pm$ 1.96 standard errors). In \textbf{b}, a box spans the quartiles around the median, and its whiskers span the 5th to the 95th percentile. Counts per panel: 2{,}500 radiographs (\textbf{a}), 971, 549, 443, 428, 71, and 38 radiographs per dataset and 1{,}264, 1{,}059, and 68 per view, with no view for the 109 VinDr radiographs (\textbf{b}), 18 models per group (\textbf{c}), 153 model pairs (\textbf{d}--\textbf{f}), and 18 models (\textbf{g}). \textbf{a}, Coefficient of determination of the per-radiograph mean mKNN over the 153 pairs for each block of variables alone and beyond the other three, with all four blocks as a dashed line. \textbf{b}, The same mKNN by source dataset and by view, with one dot per radiograph. \textbf{c}, Accuracy of a linear classifier predicting the manufacturer (15 classes) or the institution (37 classes) of a ReXGradient-160K radiograph from a model's embedding or relative representation. \textbf{d}, mKNN within one manufacturer minus that across manufacturers on 6 draws of 1{,}000 ReXGradient-160K radiographs at $k = 4$, with a permutation $p_{\mathrm{FDR}}$ over the 153 pairs. \textbf{e}, mKNN on 2{,}500 TAIX-Ray or ReXGradient-160K radiographs minus that on 2{,}500 radiographs of the chest pool, MIMIC-CXR, or anteroposterior MIMIC-CXR. \textbf{f}, Mean mKNN over 4 draws of 2{,}500 of 5{,}000 CheXpert radiographs under each model's own and under one shared preprocessing, with the identity dashed. \textbf{g}, Mean mKNN of each model with the other 17 on 20{,}000 radiographs against the mean AUROC of its linear classifiers over the 14 findings, with $p_{\mathrm{FDR}}$ over the tests of size, release year, and accuracy. AUROC, area under the receiver operating characteristic curve; mKNN, mutual $k$-nearest-neighbor overlap; $p_{\mathrm{FDR}}$, $p$-value corrected for the false discovery rate.}
\label{fig:content}
\end{figure*}

We decomposed the per-radiograph agreement of the 153 model pairs on 2,500 chest radiographs into 4 blocks of variables, namely the source dataset, the radiographic view, sex and age, and the 14 findings (Fig.~\ref{fig:content}a). The coefficient of determination is 0.483 for the source dataset alone, 0.359 for the view, 0.154 for sex and age, and 0.104 for the findings. The variance explained by a block beyond the other three is 0.138 for the source dataset, 0.017 for the view, 0.014 for sex and age, and 0.007 for the findings (every block above its permutation reference, $p_{\mathrm{FDR}} \leq 0.002$). The mean agreement ranges from 0.071 on MIMIC-CXR to 0.334 on the pediatric radiographs of VinDr-PCXR and is 0.069 on anteroposterior and 0.102 on posteroanterior radiographs (Fig.~\ref{fig:content}b). On ReXGradient-160K radiographs, a linear classifier on the embeddings predicts the scanner manufacturer among 15 with an accuracy of 0.857 and the institution among 37 with an accuracy of 0.466 (Fig.~\ref{fig:content}c). On the relative representation, the two accuracies are 0.631 and 0.283. The most frequent class is correct for 0.319 of the radiographs for the manufacturer and 0.203 for the institution. After FDR correction, no model pair agrees differently within one manufacturer than across manufacturers (Fig.~\ref{fig:content}d).

Agreement on chest radiographs outside the pretraining data of every model is not significantly lower than on MIMIC-CXR, which is in the documented pretraining data of three models (Fig.~\ref{fig:content}e). TAIX-Ray was released after every model. No model card names ReXGradient-160K. Against MIMIC-CXR, agreement differs by $-0.003$ $\pm$ 0.003 [$-0.009$, 0.004] on TAIX-Ray and by 0.026 $\pm$ 0.004 [0.019, 0.033] on ReXGradient-160K. Against the anteroposterior MIMIC-CXR radiographs, which match the bedside view of TAIX-Ray, the differences are 0.009 and 0.038. On the whole chest pool, agreement is 0.033 $\pm$ 0.004 [0.026, 0.040] higher than on TAIX-Ray and differs by 0.004 $\pm$ 0.003 [$-0.002$, 0.010] from that on ReXGradient-160K. Re-extracting all models on 5,000 CheXpert radiographs with a shared preprocessing changes agreement by $-0.011$ $\pm$ 0.007 [$-0.024$, 0.003] on average over the 153 pairs, by $-0.056$ for the 17 pairs involving MedGemma, and by $-0.005$ for the other 136 pairs (Fig.~\ref{fig:content}f).

Among the public models, agreement does not indicate accuracy (Fig.~\ref{fig:content}g). The mean agreement of a model with the other models is not significantly correlated with the AUROC of a linear classifier on its embeddings (Spearman $-0.185$, $p_{\mathrm{FDR}} = 0.463$). RAD-DINO, BiomedCLIP~\cite{Zhang2023BiomedCLIP}, and TorchXRayVision were developed for medical images and are among the four least aligned and the four most accurate models (AUROC 0.794, 0.765, and 0.758, against 0.725 to 0.781 for the other models). The consensus of the 18 models arranges the 14 findings by their co-occurrence (Supplementary Fig.~\ref{sfig:consensus}d; Supplementary Note~\ref{snote:consensus}). For the co-occurrence, its Mantel statistic is 0.713 against 0.175 for a reference that preserves the co-occurrence ($p_{\mathrm{FDR}} = 0.010$). For the ICD-10 hierarchy, it is 0.085 against 0.055 ($p_{\mathrm{FDR}} = 0.338$). On single-label dermoscopy images, no model's class centroids correlate significantly with the lesion lineage or with malignancy (Supplementary Fig.~\ref{sfig:consensus}e,f).

\subsection*{Comparison with radiologists}

We compared the distances between radiographs in the embedding of each model with the judgment of two board-certified radiologists, SZ with 6 years of experience and LA with 10 years. In each triplet, a radiologist chose which of two candidate chest radiographs is clinically more similar to a reference radiograph. SZ judged 300 triplets, and LA judged 100 of them. On these 100, the two radiologists chose the same candidate in 73 (0.730 $\pm$ 0.045 [0.640, 0.810]; Cohen's kappa 0.459, $p < 0.001$; Table~\ref{etab:readers}). A model chooses the candidate at the smaller cosine distance from the reference. Against the 300 choices of SZ, the accuracy of the 18 public models is 0.427 to 0.513. Since a radiologist may judge similarity by the findings, we also scored the seven models of the chest setting trained with label supervision on the findings. Their accuracy is 0.442 to 0.522 on the 276 triplets inside their held-out pool. Because the triplets were not conditioned on a finding, we also scored the 77 triplets in which only one candidate shares a finding with the reference. On these, the accuracy is 0.390 to 0.558 for the public models and 0.429 to 0.529 for the label-supervised models on the 70 inside their held-out pool. No model has a 95\% CI above the chance level of 0.5 in any of these comparisons. On the 300 triplets, the CIs of MedGemma and CONCH lie below it.

\begin{table}[p]
\centering
\caption{Comparison of the models with the radiologists SZ and LA. In each triplet, a radiologist chose which of two candidate chest radiographs is clinically more similar to a reference radiograph. A model is correct when the candidate at the smaller cosine distance from the reference is the candidate chosen by SZ. No model produced a tie. A triplet is well posed when only one candidate shares a finding with the reference and ambiguous otherwise. The models of the chest setting are scored on the triplets inside their held-out pool. Each value is the mean $\pm$ SD [95\% CI] over 1{,}000 bootstrap resamples of the triplets, except Cohen's kappa, which is given with a two-sided $p$ from 1{,}000 permutations. The chance level is 0.5. CI, confidence interval; SD, standard deviation.}
\label{etab:readers}
{\scriptsize\hyphenpenalty=10000\exhyphenpenalty=10000
\setlength{\tabcolsep}{3pt}
\renewcommand{\arraystretch}{1.08}
\begin{tabular*}{\textwidth}{@{\extracolsep{\fill}}>{\raggedright\arraybackslash}p{0.24\textwidth}rrr@{}}
\toprule
\multicolumn{4}{@{}l}{\textbf{Agreement of the two radiologists on the 100 triplets judged by both radiologists}} \\
\midrule
Share with the same choice & \multicolumn{3}{r}{0.730 $\pm$ 0.045 [0.640, 0.810]} \\
Cohen's kappa & \multicolumn{3}{r}{0.459, $p$ $<$ 0.001} \\
\midrule
\multicolumn{4}{@{}l}{\textbf{Public models, 300 triplets (223 ambiguous, 77 well posed)}} \\
\midrule
Model & All triplets & Ambiguous & Well posed \\
\cmidrule{1-4}
DINOv2 ViT-L & 0.483 $\pm$ 0.028 [0.430, 0.540] & 0.484 $\pm$ 0.034 [0.422, 0.556] & 0.481 $\pm$ 0.058 [0.377, 0.597] \\
DINOv3 ViT-L & 0.460 $\pm$ 0.029 [0.403, 0.517] & 0.475 $\pm$ 0.032 [0.413, 0.534] & 0.416 $\pm$ 0.057 [0.312, 0.532] \\
RAD-DINO & 0.497 $\pm$ 0.029 [0.440, 0.553] & 0.511 $\pm$ 0.035 [0.439, 0.578] & 0.455 $\pm$ 0.059 [0.351, 0.571] \\
Phikon-v2 & 0.470 $\pm$ 0.029 [0.413, 0.527] & 0.484 $\pm$ 0.033 [0.417, 0.547] & 0.429 $\pm$ 0.057 [0.324, 0.545] \\
Prov-GigaPath & 0.460 $\pm$ 0.028 [0.403, 0.517] & 0.466 $\pm$ 0.033 [0.399, 0.534] & 0.442 $\pm$ 0.058 [0.338, 0.571] \\
UNI & 0.490 $\pm$ 0.028 [0.437, 0.543] & 0.489 $\pm$ 0.032 [0.422, 0.547] & 0.494 $\pm$ 0.056 [0.390, 0.610] \\
UNI2-h & 0.463 $\pm$ 0.028 [0.407, 0.517] & 0.484 $\pm$ 0.033 [0.417, 0.552] & 0.403 $\pm$ 0.057 [0.299, 0.519] \\
Virchow & 0.447 $\pm$ 0.028 [0.393, 0.503] & 0.466 $\pm$ 0.033 [0.403, 0.534] & 0.390 $\pm$ 0.056 [0.286, 0.506] \\
Virchow2 & 0.510 $\pm$ 0.030 [0.450, 0.567] & 0.529 $\pm$ 0.034 [0.466, 0.596] & 0.455 $\pm$ 0.057 [0.351, 0.571] \\
RETFound & 0.493 $\pm$ 0.027 [0.440, 0.547] & 0.471 $\pm$ 0.034 [0.408, 0.538] & 0.558 $\pm$ 0.055 [0.455, 0.675] \\
CLIP ViT-L/14 & 0.500 $\pm$ 0.029 [0.440, 0.557] & 0.520 $\pm$ 0.033 [0.453, 0.583] & 0.442 $\pm$ 0.056 [0.338, 0.546] \\
SigLIP 2 ViT-L & 0.510 $\pm$ 0.029 [0.450, 0.567] & 0.538 $\pm$ 0.033 [0.475, 0.605] & 0.429 $\pm$ 0.058 [0.324, 0.545] \\
LLaVA-Med & 0.490 $\pm$ 0.029 [0.430, 0.550] & 0.489 $\pm$ 0.033 [0.422, 0.552] & 0.494 $\pm$ 0.058 [0.389, 0.610] \\
LLaVA-OneVision & 0.477 $\pm$ 0.027 [0.423, 0.533] & 0.475 $\pm$ 0.033 [0.413, 0.538] & 0.481 $\pm$ 0.059 [0.377, 0.610] \\
MedGemma & 0.430 $\pm$ 0.029 [0.373, 0.487] & 0.426 $\pm$ 0.033 [0.359, 0.493] & 0.442 $\pm$ 0.056 [0.338, 0.558] \\
BiomedCLIP & 0.463 $\pm$ 0.030 [0.407, 0.520] & 0.484 $\pm$ 0.033 [0.417, 0.547] & 0.403 $\pm$ 0.058 [0.299, 0.519] \\
CONCH & 0.427 $\pm$ 0.028 [0.370, 0.483] & 0.439 $\pm$ 0.034 [0.372, 0.502] & 0.390 $\pm$ 0.057 [0.286, 0.506] \\
TorchXRayVision & 0.513 $\pm$ 0.030 [0.457, 0.570] & 0.511 $\pm$ 0.034 [0.444, 0.574] & 0.519 $\pm$ 0.056 [0.416, 0.636] \\
\midrule
\multicolumn{4}{@{}l}{\textbf{Label-supervised models of the chest setting, 276 triplets (206 ambiguous, 70 well posed)}} \\
\midrule
ViT-S, DINOv3, seed 0 & 0.507 $\pm$ 0.030 [0.449, 0.565] & 0.515 $\pm$ 0.036 [0.442, 0.583] & 0.486 $\pm$ 0.060 [0.371, 0.600] \\
ViT-S, DINOv3, seed 1 & 0.522 $\pm$ 0.030 [0.467, 0.583] & 0.519 $\pm$ 0.035 [0.451, 0.587] & 0.529 $\pm$ 0.058 [0.429, 0.643] \\
ViT-S, DINOv3, seed 2 & 0.511 $\pm$ 0.031 [0.453, 0.572] & 0.534 $\pm$ 0.036 [0.461, 0.602] & 0.443 $\pm$ 0.059 [0.329, 0.557] \\
ViT-B, DINOv3, seed 0 & 0.493 $\pm$ 0.030 [0.435, 0.551] & 0.500 $\pm$ 0.035 [0.437, 0.573] & 0.471 $\pm$ 0.058 [0.357, 0.586] \\
ViT-B, DINOv3, seed 1 & 0.442 $\pm$ 0.030 [0.384, 0.500] & 0.447 $\pm$ 0.035 [0.379, 0.515] & 0.429 $\pm$ 0.058 [0.314, 0.543] \\
ViT-B, DINOv3, seed 2 & 0.507 $\pm$ 0.030 [0.449, 0.569] & 0.524 $\pm$ 0.035 [0.456, 0.587] & 0.457 $\pm$ 0.061 [0.343, 0.586] \\
ViT-S, random weights & 0.475 $\pm$ 0.030 [0.413, 0.536] & 0.476 $\pm$ 0.036 [0.403, 0.544] & 0.471 $\pm$ 0.059 [0.357, 0.586] \\
\bottomrule
\end{tabular*}}
\end{table}

The conditioning of the triplets, the label-supervised models, and the classifiers on the chest pool use the dataset labels (Supplementary Note~\ref{snote:data}). We therefore compared these labels with the answers of the radiologists. For 480 radiographs, SZ answered whether one named finding was present. LA answered the same question for 150 of these radiographs and for 150 further radiographs. The dataset label agrees with SZ on 0.744 $\pm$ 0.020 [0.704, 0.783] of the 480 radiographs and with LA on 0.660 $\pm$ 0.039 [0.580, 0.733] of the 150 further radiographs (Supplementary Fig.~\ref{sfig:readers}a). Against SZ, its accuracy ranges from 0.634 on NIH ChestX-ray14 to 0.863 on VinDr-CXR across the source datasets (Supplementary Fig.~\ref{sfig:readers}c,d). On the 150 radiographs read by both radiologists, they gave the same answer on 111 (0.740 $\pm$ 0.037 [0.667, 0.813]; Cohen's kappa 0.462, $p < 0.001$; Supplementary Fig.~\ref{sfig:readers}b). No model output is used in these comparisons.

\section*{Discussion}

Public medical foundation models agree more than a randomly initialized network in four of the five modalities but share few nearest neighbors. Label supervision yields the least convergent models. On chest radiographs, two label-supervised models that differ only in their random seed share 0.062 to 0.100 of their nearest neighbors. Two models trained with self-distillation, masked image modeling, or contrastive pretraining with images or reports share 0.411 to 0.802. Label supervision is the least convergent objective at both model sizes in chest radiography and in histopathology, and in a second chest setting with a different training recipe. The pre-specified rule that the objective sets agreement is met in four of these five settings. For a developer of a medical foundation model, the pretraining objective is therefore a choice between a model whose nearest neighbors are interchangeable with those of other models of the same recipe and a model whose nearest neighbors are idiosyncratic. A retrained label-supervised chest model shares fewer than one of 10 nearest neighbors with its predecessor, while a chest model retrained with any of the other objectives shares four or more. Similar-case retrieval~\cite{Chen2022SISH,Choe2022CBIR} and other applications built on nearest neighbors depend on these neighbors.

For clinical deployment, a classifier validated on one model can be transferred to a new model through a linear mapping fitted on unlabeled radiographs embedded by both models. On 14 chest findings and 306 model pairs, transferred classifiers retained 0.987 of their AUROC with 4,096 radiographs. With up to 25,000 radiographs, they scored 0.011 higher than classifiers refitted on the new model with up to 25,000 labels. Methods that translate between embedding spaces were introduced on text and natural images~\cite{Jha2025Universal,Moschella2023Relative}. In a predetermined change control plan~\cite{FDA2024PCCP}, the linear mapping could make the replacement of a model a procedure that needs no labels for the transfer and can be specified in advance. The pre-specified rule concerned the relative representation. Through it, classifiers retained 0.866 of the AUROC of a refit on the relative representation of the new model, kept little of their laterality performance on TAIX-Ray, and outperformed refitting only below about 200 labels. Retention through the relative representation was not significantly associated with agreement, among the public models or among the controlled chest models. Agreement measured on unlabeled radiographs was correlated with retention through the linear mapping in an analysis added after the rules were fixed. If this association is confirmed, agreement could be measured before a model is replaced.

Agreement between models is not evidence that they are accurate. On chest radiographs, the source dataset and the radiographic view explain more of the variance in agreement than the findings do (coefficients of determination 0.483 and 0.359 against 0.104). The scanner manufacturer can be predicted with an accuracy of 0.857 from the embeddings and 0.631 from the relative representation. The majority class is correct for 0.319 of the radiographs. Histopathology foundation models also encode the medical center of a slide~\cite{DeJong2025PathologyCenter}. Site effects likewise limit chest radiograph classifiers across hospitals~\cite{TayebiArasteh2023DomainGeneralization}. RAD-DINO, BiomedCLIP, and TorchXRayVision were developed for medical images. All three are among the four most accurate and among the four least aligned models. In the controlled study, the label-supervised models are among the most accurate and the least convergent. Agreement is not significantly correlated with accuracy in any of its four groups of models. In this panel, selecting a model for its agreement with the others would not select the most accurate model. No model agreed with a radiologist's choice of the clinically more similar chest radiograph significantly above chance, although two radiologists agreed on 73 of 100 such choices.

The Platonic Representation Hypothesis credits scale and the diversity of the training data for convergence~\cite{Huh2024Platonic}. In medicine, clinical supervision is also expected to drive convergence. Neither account predicts the differences between the controlled models. With the data, the architecture, the size, and the initialization held fixed, the most and the least convergent objectives differ in seed-pair agreement by 0.590 on chest radiographs at ViT-S. Label supervision is the least convergent objective in all five settings. Image-report contrastive learning also uses clinical supervision. Two of its models agree between seeds at 0.587 to 0.668 on chest radiographs. Convergence therefore differs between label supervision and the other four objectives, and not between self-supervision and clinical supervision. In the public models, the association of agreement with size and release year is not significant after controlling for the embedding width. Gröger et al. describe the same width effect for these statistics~\cite{Groeger2026Aristotelian}. Ciernik et al. reported on natural images that the consistency of representational similarity across datasets depends more on the objective than on the architecture or the data~\cite{Ciernik2025Objective}. Our controlled models extend that observation to medical images and to agreement between independently trained models. One possible explanation is that a loss over 13 findings constrains only the directions of the embedding that separate the findings, while the other objectives constrain the position of every image relative to the other images of a batch or the reconstruction of its masked patches. In a synthetic generative model, two models trained to predict a label agree less than two models trained to reconstruct their input at every informativeness of the label (Supplementary Note~\ref{snote:synthetic}). We did not test this explanation in the trained models. In each modality, the most aligned pair of public models shares a DINOv2 recipe~\cite{Oquab2024DINOv2,Chen2024UNI,Filiot2024Phikon,Xu2024GigaPath} or a CLIP or SigLIP initialization~\cite{Radford2021CLIP,Tschannen2025SigLIP2}. Convergence measured between such models is in part a shared lineage.

This study has several limitations. First, the objective effect is established from a shared DINOv3 initialization and three seeds per objective. The randomly initialized control has one seed per objective, since a full second grid would have doubled the training. At the matched budget, it is less accurate than the mean of the initialized models under four of the five objectives on chest radiographs and under every objective on histopathology patches. With three seeds, a single divergent model can change the outcome of the pre-specified rule. On histopathology patches at ViT-B, one masked-image-modeling model agrees with the other two at 0.365 and 0.376, while those two agree with each other at 0.664. The rule is not met in that setting. Training more seeds per objective, also from random weights and with a longer schedule, would test the effect without a shared initialization and estimate the variation between seeds more precisely. Second, the objective effect is shown in chest radiography and histopathology. Classifier transfer and the content of agreement are shown in chest radiography alone. Dermoscopy, fundus photography, and mammography have at most 5,987 images here. Only the chest datasets share a vocabulary of labeled findings. The histopathology setting also lacks a report objective, since no captioned histopathology collection shares its images with a labeled collection. Larger labeled collections in the other modalities and captions for labeled histopathology images would allow these analyses. Third, the objective effect was measured with the local statistics mKNN and CKNNA, since Huh et al. measured the Platonic Representation Hypothesis with mKNN~\cite{Huh2024Platonic}. CKA, RSA, and the Procrustes distance order the pairs of public models differently from mKNN. The objective effect remains to be tested under these global statistics. Fourth, ReXGradient-160K counts as unseen only because no model card names it. The MedGemma data card also states that its training mix includes private data. Only TAIX-Ray is certainly unseen, since it was released after every model. The image lists of the pretraining data would be needed to check the other sources. Fifth, TorchXRayVision is the only label-supervised model among the public models. It is also the smallest public model and the only convolutional network among them. The claim about label supervision therefore comes from the controlled study. Testing it among public models requires more publicly released label-supervised transformers. Sixth, the human reference consists of two radiologists of one department. The second radiologist judged only 100 of the 300 triplets, which were used to measure the agreement between the two radiologists. The triplets were drawn at random before any conditioning was defined. Only 77 of them are well posed by the dataset labels. Drawing each triplet for one finding and having radiologists of several institutions judge it would separate the ambiguity of the question from the distances of the models. Seventh, we did not test whether disagreement between models flags a perturbed or unusual radiograph, since the test requires radiographs with known perturbations. A benchmark of such radiographs would allow the agreement statistics to be tested as a detector.

Medical foundation models converge less under label supervision. Public models agree more than a randomly initialized network in four of the five modalities but share few nearest neighbors. On chest radiographs, two label-supervised models that differ only in their random seed or their size share few nearest neighbors, while models trained with self-distillation, masked image modeling, or contrastive pretraining share many more. Agreement on chest radiographs depends more on the source dataset and the radiographic view than on the findings. It is not significantly correlated with accuracy. Classifiers can nevertheless be transferred between models through a linear mapping fitted on 8,192 or more unlabeled radiographs, with no loss of AUROC on average over 14 chest findings and 306 model pairs. Developers can therefore choose convergence when they choose the pretraining objective. A chest radiograph classifier can be transferred to a new model without new labels.

\section*{Methods}

\subsection*{Ethics statement}

This study was conducted in accordance with the relevant guidelines and regulations. It used previously collected, de-identified images, reports, and labels released for research, some openly and some under credentialed access. MIMIC-CXR~\cite{Johnson2019MIMICCXR} and MIMIC-IV~\cite{Johnson2023MIMICIV} were approved by the institutional review board of Beth Israel Deaconess Medical Center, which waived individual patient consent. CheXpert~\cite{Irvin2019CheXpert} was approved by the institutional review board of Stanford Hospital, which waived individual consent for the de-identified data. CheXpert Plus~\cite{Chambon2024CheXpertPlus} adds the de-identified reports of the same studies. The institutional review boards of Hanoi Medical University Hospital and Hospital 108 approved VinDr-CXR~\cite{Nguyen2022VinDrCXR} and VinDr-Mammo~\cite{Nguyen2023VinDrMammo} and waived patient consent. The institutional review board of Phu Tho Obstetric and Pediatric Hospital approved VinDr-PCXR~\cite{Pham2022VinDrPCXR} and waived patient consent. The ethics committee of the RWTH Aachen Faculty of Medicine approved TAIX-Ray~\cite{Truhn2026TAIXRay} and waived patient consent. An institutional research committee approved PadChest~\cite{Bustos2020PadChest}. The ethics committee of Heidelberg University approved the samples of NCT-CRC-HE-100K and CRC-VAL-HE-7K~\cite{Kather2018NCT} from the tissue bank of the National Center for Tumor Diseases, whose patients gave informed consent under the protocol of the tissue bank. Ethics Board II of University Medical Center Mannheim approved the NCT-CRC-HE-100K samples from its pathology archive and waived consent for them. The ethics committees of the Medical University of Vienna and the University of Queensland approved HAM10000~\cite{Tschandl2018HAM10000}. BCN20000~\cite{Combalia2019BCN20000} was collected with institutional ethics approval at Hospital Cl\'{\i}nic de Barcelona and with informed consent from every participant. The NIH Clinical Center released NIH ChestX-ray14~\cite{Wang2017ChestXray14} after removing all personally identifiable information. No approving committee and no consent procedure are reported for NIH ChestX-ray14, ReXGradient-160K~\cite{Zhang2025Rexgradient}, PatchCamelyon~\cite{Veeling2018PCam}, the MSK images of ISIC 2019~\cite{Codella2019ISIC}, APTOS~\cite{Karthik2019APTOS}, or Messidor-2~\cite{Decenciere2014Messidor}. Each dataset was used according to the conditions of its release. No new patient data were generated and no participants were recruited. This secondary analysis therefore required no further ethics approval or informed consent. No re-identification was attempted. No image or report is redistributed. Every model ran on local hardware. No image, report, or record was passed to an external service.

\subsection*{Data sources and pools}

\paragraph{Chest radiographs.} The chest pool contains $n = 650{,}982$ frontal radiographs from 6 sources: MIMIC-CXR~\cite{Johnson2019MIMICCXR} ($n = 243{,}334$ from 63,945 patients, Beth Israel Deaconess Medical Center, Boston, USA), CheXpert~\cite{Irvin2019CheXpert} ($n = 157{,}878$ from 57,872 patients, Stanford Health Care, USA), NIH ChestX-ray14~\cite{Wang2017ChestXray14} ($n = 112{,}120$ from 30,805 patients, NIH Clinical Center, USA), PadChest~\cite{Bustos2020PadChest} ($n = 110{,}525$ from 67,205 patients, Hospital San Juan, Alicante, Spain), VinDr-CXR~\cite{Nguyen2022VinDrCXR} ($n = 18{,}000$, Hospital 108 and Hanoi Medical University Hospital, Vietnam), and VinDr-PCXR~\cite{Pham2022VinDrPCXR} ($n = 9{,}125$, Phu Tho Obstetric and Pediatric Hospital, Vietnam). Only posteroanterior and anteroposterior views were kept. We also removed every frontal CheXpert radiograph with an uncertain label for cardiomegaly, lung opacity, lung lesion, edema, or pneumonia ($n = 33{,}351$). No such removal was applied to MIMIC-CXR. Each source is divided into training, validation, and test partitions. The test partition is the released test set for NIH ChestX-ray14, VinDr-CXR, and VinDr-PCXR. MIMIC-CXR, CheXpert, and PadChest were divided by patient, with 17\% to 20\% of the patients of each source in the test partition. The pool has $n = 475{,}269$ training, $n = 49{,}718$ validation, and $n = 125{,}995$ test radiographs (Table~\ref{tab:sources}). Every partition boundary was verified to be a patient boundary for the four sources that release patient identifiers. Every radiograph was resized once to $224 \times 224$ pixels with Lanczos resampling. Agreement between two public models was measured on $n = 20{,}000$ radiographs drawn at random with a fixed seed. Every comparison across modalities was made on another such draw of $n = 2{,}500$ radiographs. Both draws were taken from the radiographs with a finite embedding in both models.

\paragraph{TAIX-Ray and ReXGradient-160K.} TAIX-Ray~\cite{Truhn2026TAIXRay} is the bedside chest radiography collection of University Hospital RWTH Aachen, released in April 2026, after every public model. For each radiograph, the presence of each finding is recorded in a structured report, with a grade on an ordinal scale from 0 to 4 and a side where it applies. The identity of the reporting radiologist is released with each report. The pool contains $n = 50{,}000$ radiographs of 11,475 patients, drawn by patient from the released partitions in their released proportions ($n = 31{,}942$ training, $n = 8{,}093$ validation and $n = 9{,}965$ test radiographs). The presence labels are cardiomegaly, pleural effusion, lung opacity, atelectasis, and congestion, which is used in place of edema. The graded items are cardiomegaly, congestion, and the left and right sides of pleural effusion, lung opacity, and atelectasis, eight in all. ReXGradient-160K~\cite{Zhang2025Rexgradient} contains radiographs from 79 sites of 3 health systems in the USA with the scanner manufacturer, the institution, and the report of every study. Its pool contains the frontal computed or digital radiographs with a recorded manufacturer and institution, capped by patient at $n = 20{,}000$ radiographs of 13,721 patients ($n = 17{,}400$ training, $n = 1{,}272$ validation and $n = 1{,}328$ test radiographs). No model card of a public model names either source as pretraining data. ReXGradient-160K was released in May 2025, before DINOv3 and MedGemma 1.5. It is therefore treated as unseen only on the strength of the model cards.

\paragraph{Histopathology, dermoscopy, fundus photography, and mammography.} The histopathology pool contains $n = 22{,}000$ patches at $224 \times 224$ pixels. Of these, $n = 18{,}000$ come from NCT-CRC-HE-100K and CRC-VAL-HE-7K~\cite{Kather2018NCT}, with $n = 2{,}000$ for each of the 9 tissue classes (adipose, background, debris, lymphocytes, mucus, smooth muscle, normal colonic mucosa, stroma, and tumor epithelium) and with $n = 16{,}784$ training and $n = 1{,}216$ validation patches. The other $n = 4{,}000$ are test patches from PatchCamelyon~\cite{Veeling2018PCam}, $n = 2{,}000$ tumor and $n = 2{,}000$ non-tumor. They were upsampled from $96 \times 96$ pixels with Lanczos resampling. Agreement was measured on $n = 20{,}000$ of these patches drawn at random. The dermoscopy pool contains $n = 5{,}987$ single-diagnosis images of ISIC 2019~\cite{Tschandl2018HAM10000,Combalia2019BCN20000,Codella2019ISIC}, at most $n = 1{,}000$ per diagnosis over melanocytic nevus, melanoma, basal cell carcinoma, benign keratosis, actinic keratosis, squamous cell carcinoma, vascular lesion, and dermatofibroma. The fundus pool contains $n = 3{,}738$ photographs, $n = 2{,}974$ from APTOS~\cite{Karthik2019APTOS} and $n = 764$ gradable photographs from Messidor-2~\cite{Decenciere2014Messidor}, sampled to balance referable diabetic retinopathy, defined as a grade of 2 or more, against non-referable photographs within each source. The mammography pool contains $n = 2{,}500$ images of VinDr-Mammo~\cite{Nguyen2023VinDrMammo}, drawn at random from all of its images and labeled for mass, suspicious calcification, focal asymmetry, architectural distortion, asymmetry, and no finding.

\paragraph{Reports and labels.} The report of a radiograph is the released report file for MIMIC-CXR, the findings and impression sections of CheXpert Plus~\cite{Chambon2024CheXpertPlus} for CheXpert, and the findings and impression fields for ReXGradient-160K. Where both CheXpert Plus sections are empty, the full report is used. A radiograph without a report text is left out of the agreement between images and reports. Every chest source's native labels were mapped onto the 14 findings of the CheXpert vocabulary, which are atelectasis, cardiomegaly, consolidation, edema, enlarged cardiomediastinum, fracture, lung lesion, lung opacity, no finding, pleural effusion, pleural other, pneumonia, pneumothorax, and support devices (Supplementary Note~\ref{snote:data}). A finding not labeled by a source is missing for every radiograph of that source (Supplementary Table~\ref{stab:label_mapping}). A finding was taken as present when the source labeled it present, as absent when the source labeled it negative or did not mention it, and as missing for that radiograph when the source marked the mention as uncertain. In the chest pool, sex and age are read from every source that records them, and race, ethnicity, and insurance from CheXpert alone.

\paragraph{Anchors.} The anchors of the relative representation are $n = 4{,}096$ radiographs of the MIMIC-CXR training partition, stored by case identifier. For each of the 14 findings, in the order listed above, $n = 292$ positive radiographs were chosen. The list was deduplicated and completed to 4,096 with radiographs drawn at random from the other MIMIC-CXR training radiographs. The $n = 1{,}024$ anchors of the primary count are the first 1,024 of that list, which are positive radiographs of atelectasis, cardiomegaly, consolidation, and part of edema. The 256-anchor variant is the first 256 radiographs of that list, and the 4,096-anchor variant is the full list.

\subsection*{Public models and feature extraction}

\paragraph{Models.} The public models are 18 open-weight image models and 7 text models (Table~\ref{etab:panel}). We loaded TorchXRayVision from its own package~\cite{Cohen2022TorchXRayVision} and the other models from the Hugging Face Hub. Every model ran locally as a frozen feature extractor. The image models ran in 16-bit floating point, except TorchXRayVision in 32-bit floating point. A DINO-family model whose output was not finite was moved to bfloat16, and to 32-bit floating point if its output was still not finite. The text models ran in 16-bit floating point, with the weights of MedGemma 27B quantized to 4 bits. The revision of each image model was recorded at extraction. Each text model was loaded at the latest commit of its repository, which predates every extraction. The image models are RAD-DINO~\cite{PerezGarcia2025RadDino}, TorchXRayVision, BiomedCLIP~\cite{Zhang2023BiomedCLIP}, DINOv3 ViT-L/16~\cite{Simenoni2025DINOv3}, DINOv2 ViT-L/14~\cite{Oquab2024DINOv2}, CLIP ViT-L/14~\cite{Radford2021CLIP}, SigLIP 2 ViT-L/16~\cite{Tschannen2025SigLIP2}, UNI~\cite{Chen2024UNI}, UNI2-h, Virchow~\cite{Vorontsov2024Virchow}, Virchow2~\cite{Zimmermann2024Virchow2}, Prov-GigaPath~\cite{Xu2024GigaPath}, CONCH~\cite{Lu2024CONCH}, Phikon-v2~\cite{Filiot2024Phikon}, and RETFound~\cite{Zhou2023RETFound}. The other three image models are the vision encoders of MedGemma 1.5~\cite{Sellergren2025MedGemma}, LLaVA-Med~\cite{Li2023LLaVAMed}, and LLaVA-OneVision~\cite{Li2024LLaVAOneVision}. The text models are PubMedBERT~\cite{Gu2021PubMedBERT}, the MedCPT query and article encoders~\cite{Jin2023MedCPT}, the text encoders of BiomedCLIP and CONCH, MedGemma 27B, and SapBERT~\cite{Liu2021SapBERT}.

\paragraph{Feature extraction.} Each image model received the $224 \times 224$ image through its own released preprocessing. CONCH received it at $224 \times 224$ pixels with the CLIP normalization, in place of its released preprocessing at $448 \times 448$ pixels. The feature vector is the class token of the last layer for the DINO-family models, the timm-loaded histopathology models, and RETFound. It is the projected pooled image embedding for CLIP, SigLIP 2, BiomedCLIP, and CONCH. For the vision encoders of the three vision-language models, it is the mean over all tokens of the last layer. Where the processor splits an image into tiles, the tile features are averaged. For TorchXRayVision, it is the globally pooled 1,024-dimensional feature map. A text is embedded as the class token for PubMedBERT and SapBERT, the mean over tokens for MedCPT, the projected text embedding for BiomedCLIP and CONCH, and the last attended token's final hidden state for MedGemma 27B. Texts were truncated at 512 tokens, at 256 for BiomedCLIP, and at 127 for CONCH. Every feature vector is L2-normalized. The random network is a ViT-L/16 of the timm library with randomly initialized weights, initialized anew for each pool, and embedded through the same pipeline. For the preprocessing control, every image model was run a second time on the first 5,000 radiographs of the chest pool in sorted order of their identifiers, which are all CheXpert training radiographs. In this run, every image was resized so that its shorter side has 224 pixels, center-cropped to $224 \times 224$ pixels, normalized with the ImageNet mean and SD, and then resampled to the input size of each model. TorchXRayVision received the same crop on its own single-channel intensity scale.

\subsection*{Agreement between models}

\paragraph{Agreement statistics.} Agreement between two models $f$ and $g$ embedding the same $N$ cases is the mKNN of Huh et al.~\cite{Huh2024Platonic}, computed on the feature vector of each model with no search over layers. For case $i$, the overlap is the share of its $k$ nearest neighbors under cosine distance in $f$ that are also among its $k$ nearest neighbors in $g$. The agreement is the mean of this share over the $N$ cases. Beside the overlap, we computed the CKNNA of the same authors as defined in their implementation. It masks the two models' cosine kernels by the intersection of the two $k$-nearest-neighbor sets and applies the unbiased estimator of the Hilbert-Schmidt independence criterion~\cite{Song2012HSIC}. A binary-graph variant of the kernel statistic is described only in the supplement, under its own name (Supplementary Note~\ref{snote:statistics}). Every comparison is made at one pool size with $k$ fixed at 0.4\% of it. Comparisons across pools, across modalities, and inside the controlled study use 2,500 cases with $k = 10$. A statement inside one pool uses the pool's own size of up to 20,000 cases, where $k$ is 80 on the chest and histopathology pools, 24 on the dermoscopy pool, 15 on the fundus pool, and 10 on the mammography pool. For a pair of public models, the agreement at 2,500 cases is the statistic on one draw of 2,500 cases with a fixed seed. For a pair of controlled models and for an image model with a text model, it is the mean over 200 draws of 2,500 cases from up to 10,000 held cases. Its interval comes from the same draws. The draws are without replacement, because a duplicated case would be its own nearest neighbor in both models. A summary over the 153 pairs of the 18 models is given with the delete-one-model jackknife, which recomputes the summary 18 times with one model and its 17 pairs removed. The jackknife standard error and the interval of $\pm 1.96$ standard errors are reported, since every model is in 17 pairs and a pair-level interval would understate the uncertainty. The randomly initialized network is the reference of each pool at the same size and the same $k$, as the mean over its 18 pairs with the public models. The rule fixed before the results were computed for the first hypothesis is that a pool has agreement above the random network only when the mean $\bar{A}_P$ of the public models and the random-network mean $\bar{R}_P$ of pool $P$ satisfy
\begin{equation}\label{eq:h1}
\bar{A}_P - 1.96\,\mathrm{SE}_{\mathrm{jk}}(\bar{A}_P) > \bar{R}_P + 1.96\,\mathrm{SE}_{\mathrm{jk}}(\bar{R}_P),
\end{equation}
where $\mathrm{SE}_{\mathrm{jk}}$ is the delete-one-model jackknife standard error. Linear and RBF CKA~\cite{Kornblith2019CKA}, the orthogonal Procrustes distance~\cite{Schonemann1966Procrustes}, and RSA~\cite{KriegeskorteKievit2013} were computed as secondary statistics on the same 2,500 cases as the mKNN of each pair (Supplementary Note~\ref{snote:statistics}). RBF CKA and RSA used a subsample of 2,000 of these cases. The order of the 153 pairs under each statistic was compared with their order under mKNN by the Spearman correlation over the pairs, with the Procrustes distance negated so that a positive coefficient means the same order. The order under mKNN in one modality was compared with that in another in the same way. The pairs were also grouped by the objectives of their two models. Each model was ranked by its mean over its 17 pairs under each statistic.

\paragraph{Size, release year, and width.} Each public model's mean agreement with the other 17 on the chest pool at 20,000 radiographs was correlated by Spearman's rank correlation with the logarithm of its parameter count, with its release year, and with its accuracy, with FDR correction over the three tests. The accuracy is the mean AUROC over the 14 findings of a logistic regression classifier on the model's features, with an inverse regularization strength of 0.1, at most 150 iterations, balanced class weights, and standardized inputs. It was fitted on up to 25,000 radiographs of the chest pool's training partition and scored on up to 20,000 of its validation and test radiographs. The logarithm of the embedding width was correlated with the agreement and with the logarithm of the parameter count, corrected over these two tests. The parameter count and the release year were then correlated again with the agreement while the logarithm of the embedding width was held fixed. This partial correlation is the Spearman correlation between the residuals of the agreement and of the covariate after a linear regression on that logarithm. The two partial tests are corrected by FDR.

\paragraph{Images and reports.} Agreement between an image model and a text model was measured at $k = 10$ with the 200-draw interval, on 2,500 radiographs of the chest pool and on 2,500 of the ReXGradient-160K pool. Each radiograph was embedded by the image model and its report by the text model. The random image network embedding the same radiographs is the reference. The chance level of two unrelated neighbor sets is $k/N = 0.004$.

\paragraph{The positive control.} We introduced a shared neighborhood structure at doses of 0, 0.2, 0.4, 0.6, 0.8, and 1, first between two random representations and then in the training data of two small models. Before any negative result about agreement was interpreted, the agreement had to be at the chance level or at the untrained reference at dose 0 and to increase with the dose (Supplementary Note~\ref{snote:synthetic}; Supplementary Fig.~\ref{sfig:synthetic}f,g). The tolerances of this rule are in Supplementary Table~\ref{stab:rules}. In the same Note, we describe a synthetic generative model with labels of controlled informativeness. No claim of this paper depends on it.

\subsection*{The controlled study}

\paragraph{Design and training.} The design is given in Table~\ref{tab:design}. Every controlled model is a ViT~\cite{Dosovitskiy2021ViT} with 16-pixel patches at $224 \times 224$ pixels, ViT-S/16 or ViT-B/16. The models were initialized either from the DINOv3 checkpoints pretrained on LVD-1689M~\cite{Simenoni2025DINOv3} or from random weights of the standard ViT of the timm library. Unlike that ViT, the DINOv3 checkpoints have four register tokens, rotary position embeddings, and layer scale. We characterized the transfer of these checkpoints to chest radiograph classification before~\cite{Arasteh2025Dinov3Resolution}. Outside the second chest setting, the models were trained with AdamW~\cite{LoshchilovHutter2019AdamW} at a learning rate of $10^{-4}$ and a weight decay of 0.05, with a linear warm-up over the first 10\% of steps followed by a cosine decay to zero. The gradient norm was clipped at 1. The gradients of 2 batches of 256 images were accumulated into an effective batch of 512. Training ran for 20,000 optimizer steps with seeds 0, 1, and 2. The forward and backward passes ran in bfloat16 mixed precision, while the weights, the optimizer states, and the teacher of self-distillation were kept in 32-bit floating point. Training stopped at 19,712 steps in histopathology and at 19,950 steps on the first set of training patients (Supplementary Note~\ref{snote:controlled}). In these settings, every objective used the same augmentation: a random resized crop to $224 \times 224$ pixels covering 70 to 100\% of the image area, a random horizontal flip, and normalization with the ImageNet mean and SD. The chest list contains the $n = 121{,}135$ MIMIC-CXR training radiographs that have a report and at least one of the 13 findings other than no finding labeled present, so that every objective trains on the same images. The histopathology list contains the $n = 16{,}784$ NCT-CRC training patches. The TAIX-Ray list contains the $n = 31{,}942$ training radiographs of its pool. No label of the three label forms is missing for them. For the comparison of training patients, the MIMIC-CXR training patients were split at random into two sets with no patient in common. Two image-image contrastive models at ViT-S, with seeds 0 and 1, were trained on the radiographs of each set.

\paragraph{Objectives.} Masked image modeling zeroes 75\% of the 196 patches in pixel space before the encoder runs, predicts the pixels of every patch from its patch token through a two-layer perceptron, and minimizes the mean squared error over the masked patches. It is a masked-input reconstruction and not the masked autoencoder of He et al.~\cite{He2022MAE}, whose encoder receives only the visible patches. Self-distillation follows DINO~\cite{Caron2021DINO} over two global views with no multi-crop schedule. The student and the teacher each have a three-layer projection head with a hidden width of 512 and an output width of 2,048. The teacher is an exponential moving average of the student with momentum 0.996 and receives no gradient. Its output is centered by a running mean with momentum 0.9 and sharpened at a temperature of 0.04. The running mean is updated after every batch of 256 images. The student's temperature is 0.1. Image-image contrastive learning is the normalized-temperature cross entropy of SimCLR~\cite{Chen2020SimCLR} over two augmented views of every image at a temperature of 0.2. Its projection head has two layers and an output width of 128. Image-report contrastive learning is the symmetric InfoNCE loss of CLIP~\cite{Radford2021CLIP} between the image and its report. PubMedBERT~\cite{Gu2021PubMedBERT} is the text encoder. Both encoders are trained, with 256-dimensional linear projections on both sides and a temperature of 0.07. Reports are truncated at 256 tokens. Both contrastive losses take their negatives from the 256 images of a batch. Label supervision is a linear head on the class token. On chest radiographs, it is trained with binary cross entropy over the 13 findings, with a missing label excluded from the loss. In histopathology, it is trained with cross entropy over the nine tissue classes. On TAIX-Ray, the same head predicts the five presence labels in the presence form and the right and left sides of pleural effusion, lung opacity, and atelectasis as six binary labels in the laterality form. In the severity form, it predicts the grade of each of the eight graded items through one softmax over the five grade levels per item. Each of these objectives was also trained twice for 300 steps at a batch of 32 from the DINOv3 ViT-S initialization in the precision of the controlled models, once on its real target and once on a target scrambled within each batch. The two models and the untrained initialization were scored against the real target on 640 held-out rows in 20 batches. An objective passed when the held-out loss of the scrambled-target model exceeded that of the real-target model by more than 1\% of the loss of the untrained initialization. The difference also had to be significant in a paired $t$ test over the 20 batches at $p < 0.05$ (Fig.~\ref{efig:start}a). The verification models were trained with AdamW at a learning rate of $10^{-4}$ and its default weight decay of 0.01.

\paragraph{The second chest setting.} With 16 further chest models, we tested whether the effect of the objective depends on the training recipe. These models share the DINOv3 initializations, the image normalization, the precision, the optimizer, the learning rate, and the weight decay with the other settings. They differ in the training list, the training schedule, the augmentation, the image-image contrastive loss, and the text encoder. The list contains the $n = 180{,}300$ MIMIC-CXR training radiographs with a recorded frontal view, including the radiographs without any of the 13 findings labeled present. Of these, 11 are decubitus or axial views, which are not in the chest pool. Every radiograph of the list has its report. Every model was trained for 50 epochs without gradient clipping. An epoch is 704 batches of 256 radiographs. The learning rate schedule is a linear warm-up over the first 3,520 optimizer steps and a cosine decay defined over 35,200 steps. The image-image contrastive and label-supervised models accumulated 2 batches into an effective batch of 512 and took 17,600 optimizer steps. The image-report models were updated after every batch and took 35,200 steps. For image-image contrastive learning, two augmented views of each radiograph were drawn. Each view is a random resized crop to $224 \times 224$ pixels covering 20 to 100\% of the image area with a random horizontal flip. A color jitter of brightness and contrast 0.4 was applied to a view with probability 0.5. The loss is the symmetric InfoNCE loss of CLIP~\cite{Radford2021CLIP} between the two views at a temperature of 0.1. The projection head has two layers, a hidden width of 2,048, and an output width of 256. Label supervision and image-report contrastive learning used a random resized crop covering 50 to 100\% of the image area with a random horizontal flip. Otherwise, they are implemented as described above, with Bio\_ClinicalBERT~\cite{Alsentzer2019ClinicalBERT} in place of PubMedBERT as the text encoder. Each objective was trained at ViT-S with seeds 0, 1, and 2 and at ViT-B with seed 0. Four further image-report models were trained at ViT-S and ViT-B with seed 0. Two of them were trained on reports that begin with a sentence on the race, sex, and age group of the patient. The race was taken from MIMIC-IV~\cite{Johnson2023MIMICIV}. The other two were trained on reports with a fixed list of words for race, sex, and age removed. The 16 models were embedded through the same pipeline as the other controlled models and are not compared with the models of the other settings.

\paragraph{Evaluation of the controlled models.} Every controlled model was embedded on the held-out pool of its setting, which contains the validation and test radiographs or patches of the pool with every patient of the training list removed. The chest held-out pool has $n = 175{,}713$ radiographs, $n = 63{,}045$ of them from MIMIC-CXR. The histopathology held-out pool has the $n = 1{,}216$ NCT-CRC validation patches and the $n = 4{,}000$ PatchCamelyon test patches. No patient of the training list of the second chest setting is in the held-out pool. The feature of every controlled model is the class token of its last layer. Agreement between two models was measured at $k = 10$ as the mean over 200 draws of 2,500 of those cases, with the interval of the draws. The rule for the second hypothesis compares, in one setting, the objective means $\bar{a}_o$ of the agreement between two models of objective $o$ that differ only in their seed with the seed pairs $a_{o,s}$ inside each objective,
\begin{equation}\label{eq:h2}
\max_{o} \bar{a}_o - \min_{o} \bar{a}_o \;>\; \max_{o}\Big(\max_{s} a_{o,s} - \min_{s} a_{o,s}\Big),
\end{equation}
and a permutation $p$ that relabels the objective within each seed 1,000 times is reported beside it. Agreement across model size is the mean over the nine combinations of a ViT-S seed and a ViT-B seed of one objective, or over three in the second chest setting. Agreement between two objectives is the mean over the nine combinations of a seed of each at one size. The agreement with the initialization is the agreement of a model with the DINOv3 checkpoint it was initialized from, embedded on the same held-out pool through the same transform and class token. The cross-initialization contrast compares the randomly initialized model of an objective at ViT-S with the three DINOv3-initialized ViT-S models of the same objective and with those of the other objectives. The difference has a jackknife standard error over the partner models. The pooled difference over objectives is tested with the delete-one-model jackknife. Accuracy was measured on the same held-out pool by one logistic regression classifier per target, fitted on 70\% of the held-out patients and scored on the other 30\%. In histopathology, which has no patient identifiers, the 1,216 held-out NCT-CRC patches were split at random into 869 training and 347 test patches. These classifiers have an inverse regularization strength of 0.1, at most 200 iterations, balanced class weights, and standardized features. Accuracy is the mean AUROC over the 14 findings on chest radiographs, the nine-class accuracy in histopathology, and the mean AUROC over the 5 presence findings on TAIX-Ray. The TAIX-Ray accuracy is computed on the presence findings for the models of every label form. The Spearman correlation between a model's accuracy and its mean agreement with the other models of its setting was corrected over the four settings.

\subsection*{Classifier transfer between models}

\paragraph{The linear mapping.} On the chest pool, every classifier and mapping of the transfer between models was fitted on the training partition and scored on the test partition. The anchor route and the refits are fitted on at most 25,000 training radiographs and scored on at most 20,000 test radiographs drawn with a fixed seed. The mapping route is scored on all test radiographs. In the mapping route, the classifier was fitted on the features of an old model $g$ and is applied to the features of a new model $f$ at deployment. For this ordered pair, the mapping from the new features $X_f$ to the old features $X_g$ of the training radiographs is the least-squares solution
\begin{equation}\label{eq:map}
\hat{W}, \hat{b} = \arg\min_{W, b}\, \big\lVert X_g - X_f W - \mathbf{1} b^{\top} \big\rVert_F^{2},
\end{equation}
fitted once per pair, without any label, on up to 25,000 training radiographs embedded finitely by both models. The classifier of $g$ for a finding, fitted on up to 25,000 labeled training radiographs of $g$, is then scored on $X_f \hat{W} + \mathbf{1}\hat{b}^{\top}$ over the test radiographs. The classifier of $f$ for the same finding, fitted in the same way on the features of $f$, is the refit with labels on the new model. Its AUROC is averaged over the pairs with $f$ as the old model. The classifier of $g$ scored on the unmapped features $X_f$ is the identity comparison, defined for the 66 ordered pairs of equal width. For the unlabeled-image budget, one fixed set of 25,000 training radiographs embedded finitely by every model was drawn. Each classifier was fitted on all of them. The mapping was fitted on their first 2,048, 4,096, 8,192, and 25,000.

\paragraph{The relative representation and retention.} In the anchor route, a radiograph $x$ is represented in model $f$ by its relative representation~\cite{Moschella2023Relative}, the cosine similarities to the $M$ anchor radiographs $a_1, \dots, a_M$ embedded by the same model,
\begin{equation}\label{eq:anchor}
r_f(x) = \big(\cos(f(x), f(a_1)), \dots, \cos(f(x), f(a_M))\big).
\end{equation}
The classifier for a finding is fitted on $r_g$ over the training radiographs of the old model $g$ and scored on $r_f$ over the test radiographs of the new model $f$, for every ordered pair and at $M$ of 256, 1,024, and 4,096. A transferred classifier with AUROC $a_{\mathrm{tr}}$ is compared with a reference AUROC $a_{\mathrm{ref}}$. In the mapping route, $a_{\mathrm{ref}}$ is the AUROC of the same classifier on the old model. In the anchor route, it is the AUROC of the classifier fitted on $r_f$ with the training labels. From the two AUROCs, we compute the retention and the chance-corrected retention
\begin{equation}\label{eq:retention}
\rho = \frac{a_{\mathrm{tr}}}{a_{\mathrm{ref}}}, \qquad \rho_{c} = \frac{a_{\mathrm{tr}} - 0.5}{a_{\mathrm{ref}} - 0.5}.
\end{equation}
Every classifier of these two routes is a logistic regression with an inverse regularization strength of 0.1, at most 150 iterations, balanced class weights, and standardized inputs. In the anchor route, the standardization fitted on $r_g$ is applied to $r_f$.

\paragraph{Refits and summaries per model.} The refit comparators are logistic regression classifiers with an inverse regularization strength of 1.0, at most 1,000 iterations, balanced class weights, and unstandardized inputs. They were fitted on the features of the new model and on its relative representation. Each fit used 50, 200, or 1,000 labeled training radiographs, drawn 20 times with the finding's prevalence and at least 2 radiographs of each class. The mean AUROC over the 20 draws, on the same test radiographs, was compared with the AUROC of the transferred classifier. The refit on the relative representation is computed over the 1,024 anchors. It was compared with the refit on the features of the same model through their difference per model, finding, and label count, with the delete-one-model jackknife over the 18 models. Retention through the linear mapping was averaged per model over its 17 partners and the 14 findings, once with the model as the old model and once with it as the new model. The interval of each average comes from the delete-one-model jackknife over the 17 partners.

\paragraph{The cross-site analysis.} The classifier of each model and finding was fitted on the MIMIC-CXR training partition in the relative representation over 1,024 anchors. It was scored on the MIMIC-CXR test partition and on every labeled radiograph of each other chest source that labels the finding, with a patient-level bootstrap interval per AUROC. A finding with fewer than 20 positive radiographs in a source was left out for that source. Only edema on VinDr-CXR, which has one positive radiograph, was left out. The paired difference against MIMIC-CXR was summarized over models and findings with the delete-one-model jackknife. Its two-sided $p$ against zero was computed from the jackknife standard error and corrected over the five datasets. The same difference was also averaged per model and dataset over the findings kept in that dataset.

\paragraph{Agreement and retention.} The rule fixed for the fourth hypothesis concerns the retention through 1,024 anchors, averaged over the 14 findings per unordered pair, and the pair's agreement on the chest pool at 20,000 radiographs. We may claim that convergence enables transfer only if retention and agreement are associated across the 153 pairs, with a positive Spearman coefficient and a linear fit with a coefficient of determination above 0. The coefficient is tested by its delete-one-model jackknife $p$. This $p$ is corrected by FDR over the three anchor counts and has to be below 0.05. The same association was computed for the mapping route, inside the controlled study over the pairs of the controlled chest models, and for the pairs of each objective family. Inside the controlled study, each classifier was fitted on at most 25,000 validation radiographs of the held-out pool and scored on at most 20,000 of its test radiographs. Both sets were drawn by patient. The controlled chest models embedded the anchor set beside their held-out pool.

\subsection*{Presence, severity, and laterality on TAIX-Ray}

\paragraph{Decodability.} For each public model and target, a logistic regression classifier was fitted on a patient-disjoint split of the TAIX-Ray pool, with 70\% of the patients for training and at most 8,000 training and 8,000 test radiographs. The classifier has the settings of the accuracy classifiers of the controlled study. It was scored by AUROC with a patient-level bootstrap interval against 200 refits on permuted labels at the same counts, with FDR correction within each target type. The 3 families have 90 presence, 144 severity, and 54 laterality classifiers. The presence target is present against absent for five findings. The severity target is grade 2 or above against grade 1 for eight graded items, with grade 0 excluded. The laterality target is right-only against left-only at matched grade for pleural effusion, lung opacity, and atelectasis, with 3,267, 1,971, and 3,123 training radiographs. Over the radiographs with a grade of 1 to 4, the ordering of the grades inside each model was tested by the Spearman correlation between the grade and the cosine distance from the centroid of the grade-0 radiographs. Each correlation was compared with 200 permutations of the grades across these radiographs and corrected over the 144 tests of a model and a graded item.

\paragraph{Transfer.} Classifiers were transferred between models by the two routes of the chest analysis: the anchor route on the standardized relative representation over the 1,024 anchors (Eq.~\ref{eq:anchor}) and the mapping route through the linear mapping of Eq.~\ref{eq:map}, fitted on the training radiographs. Every classifier on TAIX-Ray has the settings of the decodability classifiers. Transfer was measured on one patient-disjoint split of the TAIX-Ray radiographs embedded by all 18 models, with 70\% of the patients for training. At most 8,000 training and 8,000 test radiographs were drawn before the radiographs labeled for each target were kept. The AUROC before transfer is that of the classifier fitted on these radiographs. Each transferred AUROC was compared with the 97.5th percentile of 20 refits on permuted labels transferred the same way. Its retention and chance-corrected retention follow Eq.~\ref{eq:retention}. Each transferred AUROC was also compared with the AUROC of the classifier fitted on the embeddings of the new model for the same target. For laterality in the mapping route, the Spearman correlation between the transferred AUROC and the AUROC of the classifier fitted on the new model is reported over the combinations, with the delete-one-model jackknife $p$.

\subsection*{Analyses of agreement on chest radiographs}

\paragraph{Variance decomposition.} On 2,500 chest radiographs embedded finitely by every public model, the per-case agreement is the mean over the 153 pairs of the case's neighbor overlap at $k = 10$. It was regressed on 4 blocks of indicator columns: the radiographic view (3 columns), the source dataset (5), sex and age band (6), and the 14 findings. A missing view, sex, or age forms its own level. A finding without a label in a source counts as absent. The unique contribution of a block is the fit of all 4 blocks minus the fit without it. Each block's unique contribution was compared with 1,000 permutations of the block's rows across cases, with FDR correction over the 4 blocks.

\paragraph{Acquisition, exposure, and preprocessing.} On the ReXGradient-160K pool, a logistic regression classifier with the settings of the decodability classifiers predicted the scanner manufacturer and the institution, among the 15 manufacturers and the 37 institutions with at least 20 training radiographs. It was fitted on the features of each model and on its relative representation, on a patient-disjoint split with 70\% of the patients for training. Its accuracy was resampled by patient and compared with the majority-class accuracy of the same split. Agreement within a manufacturer was measured at $k = 4$ on 6 draws of 1,000 radiographs, one from each manufacturer with at least 1,000 radiographs, and compared with 6 draws of 1,000 radiographs across manufacturers. Each model pair was tested by permutation, with FDR correction over the 153 pairs. Exposure was measured per pair as the agreement on 2,500 TAIX-Ray radiographs and on 2,500 ReXGradient-160K radiographs against the agreement on 2,500 radiographs of the whole chest pool, of MIMIC-CXR alone, and of the anteroposterior radiographs of MIMIC-CXR alone. Each difference was summarized over the 153 pairs with the delete-one-model jackknife. In the preprocessing control, the agreement of each pair under each model's own preprocessing was compared with its agreement under the shared preprocessing, on 4 draws of 2,500 of the 5,000 re-extracted radiographs at $k = 10$. The change of a pair is its mean difference over the 4 draws. The mean change over the 153 pairs has the interval of the delete-one-model jackknife.

\paragraph{The consensus configuration and the structure of the findings.} A consensus configuration of the 18 public models was fitted by generalized Procrustes analysis~\cite{Gower1975GPA} over their relative representation. The distances between its 14 finding centroids were correlated by the Mantel statistic~\cite{Mantel1967} with the co-occurrence of the findings and with their distance in the ICD-10 hierarchy. Each correlation was compared with a reference that keeps the co-occurrence. The fit, the tests, and their replication on single-label dermoscopy images are given in Supplementary Note~\ref{snote:consensus} (Supplementary Fig.~\ref{sfig:consensus}).

\paragraph{Findings, subgroups, and the TAIX-Ray radiologists.} We measured agreement among the positive radiographs of each finding and inside demographic groups (Supplementary Fig.~\ref{sfig:subsets}). We also analyzed the share of the variance of the TAIX-Ray grades explained by the identity of the reporting radiologists (Supplementary Fig.~\ref{sfig:readers}e,f). Each of these analyses is exploratory and is described in Supplementary Note~\ref{snote:readers}.

\subsection*{Reader study}

Two board-certified radiologists took part, SZ with 6 years of experience and LA with 10 years. All radiographs were drawn from the chest pool with a fixed seed, stratified by 20 findings. These are the 13 CheXpert findings other than no finding and 7 findings outside the CheXpert vocabulary, namely emphysema, pulmonary fibrosis, bronchiectasis, tuberculosis, interstitial lung disease, cavitation, and lung tumor. For each finding, radiographs labeled present and radiographs labeled absent were drawn in equal numbers and evenly over the source datasets. The number of radiographs of each label was at most 20 for a CheXpert finding and at most 7 for a further finding. The set of SZ consists of $n = 480$ of these radiographs, drawn at random, of which 244 are labeled present and 236 are labeled absent for their finding. They come from CheXpert ($n = 102$), VinDr-CXR ($n = 102$), PadChest ($n = 97$), MIMIC-CXR ($n = 96$), NIH ChestX-ray14 ($n = 71$), and VinDr-PCXR ($n = 12$). LA read $n = 150$ radiographs of this set, drawn at random. LA also read $n = 150$ further radiographs, drawn in the same way from the radiographs outside the set of SZ. For these, the number of radiographs of each label was at most 7 for a CheXpert finding and at most 3 for a further finding. Of these, 74 are labeled present and 76 are labeled absent.

Each radiograph was provided as an image of $512 \times 512$ pixels with a question about the finding selected for it, such as ``Is pleural effusion present in this chest X-ray?''. The radiologist answered yes or no. The triplets are 300 sets of 3 different radiographs drawn at random from the set of SZ, with no set drawn twice. In each set, the first radiograph is the reference and the other two are the candidates B and C. The triplets were split into 3 sessions of 100 in random order. For each triplet, SZ chose the candidate clinically more similar to the reference, as judged by the findings and their clinical picture and not by incidental features of the image. LA also judged the 100 triplets of the first session, split into 3 sessions of 34, 33, and 33 in a new order. Each of these triplets had the same reference and the same candidates at the same positions for both radiologists. Each radiologist read alone, without references or consultation. Each packet contained only the radiographs and an answer sheet with the questions. The radiologists therefore saw no source dataset, no dataset label, and no model output. In the packet of LA, the radiographs and triplets shared with SZ were renamed and reordered. The packet contained no answer of SZ. The models embedded the same radiographs at $224 \times 224$ pixels.

A model chose the candidate at the smaller cosine distance from the reference and was scored correct when it chose the candidate of the radiologist. A tie between the two distances would have been excluded and counted. No model produced a tie. The 18 public models were scored against the 300 choices of SZ. The seven label-supervised models of the chest setting were scored on the triplets with all three radiographs in their held-out pool. Six of them were initialized from DINOv3 with three seeds at each size. The seventh model was initialized from random weights. A candidate shares a finding with the reference when both radiographs are labeled present for at least one of the 14 findings. After the reading, a triplet was called well posed when only one candidate shares a finding with the reference and ambiguous otherwise. Of the 300 triplets, 77 are well posed and 223 are ambiguous. Of the 276 triplets inside the held-out pool of these models, 70 are well posed and 206 are ambiguous. Each accuracy is reported with 1,000 bootstrap resamples of the triplets and compared with the chance level of 0.5.

For the finding questions, the answer of each radiologist was compared with the dataset label of the finding in the question by accuracy, sensitivity, and specificity, with 1,000 bootstrap resamples of the radiographs. Sensitivity is the share of the radiographs labeled present that were called present. Specificity is the share of those labeled absent that were called absent. After the reading, the accuracy of the dataset label against SZ was also computed per source dataset and per asked finding. On the 150 radiographs and the 100 triplets read by both radiologists, their agreement is measured by the share with the same answer and by Cohen's kappa for two raters~\cite{Cohen1960Kappa}. The share is reported with 1,000 bootstrap resamples. The two-sided $p$ of the kappa is computed from 1,000 permutations that pair the answers of one radiologist with those of the other at random. The reader analyses are exploratory.

\subsection*{Statistical analysis}

\paragraph{Metrics and intervals.} Agreement statistics, AUROC, accuracy, and retention are performance metrics, reported as the mean $\pm$ SD [95\% CI]. A summary over model pairs or over the rows of a model and a finding takes its SD and CI from the delete-one-model jackknife, whose standard error is the SD. A per-pair agreement takes them from 200 draws of the reported size without replacement. These draws vary the images and not the training run. An AUROC on a fixed test set takes them from 1,000 bootstrap resamples of its test patients, which keep every radiograph of a drawn patient together. A difference against a refit takes them from the same jackknife as a summary over model pairs. The bootstrap resamples use the fixed seed 0. The draws of cases and the permutations also use fixed seeds.

\paragraph{Statistical measures and tests.} Correlations, Cohen's kappa~\cite{Cohen1960Kappa}, the Mantel statistic~\cite{Mantel1967}, and the coefficient of determination are statistical measures, reported with a $p$-value and no interval. A Spearman correlation over model pairs or over the combinations of an ordered model pair and a target is reported with the delete-one-model jackknife $p$. This $p$ is computed from the jackknife standard error of the coefficient against zero. For the link tests, the naive $p$ over pairs is given beside it in the supplement. A Spearman correlation over models or findings is reported with the asymptotic $p$. Permutation tests use 1,000 permutations. The structural reference of the centroid tests, the label-permutation reference of the decodability classifiers, and the grade-ordering test use 200. The transfer references use 20 and are taken at their 97.5th percentile. A resampled $p$ is reported no finer than its resolution. A value at the resolution of a 1,000-permutation test is therefore written as $p < 0.001$. The paired verification test of the objectives is a paired $t$ test over 20 held-out batches. A comparison of an observed value with a permutation reference is one-sided. Its $p$ is $(b + 1)/(m + 1)$ when $b$ of the $m$ reference values are at least as large as the observed value. The permutation tests of the manufacturer comparison and of the kappa are two-sided. They count the permuted statistics at least as large in absolute value as the observed statistic. Every other test is two-sided at 0.05.

\paragraph{Multiplicity and pre-specification.} Multiplicity is controlled by FDR correction~\cite{BenjaminiHochberg1995} within each named family, written $p_{\mathrm{FDR}}$. For the public models, the families are the 35 correlations over the pairs between statistics and between modalities, the 3 covariates of the size analysis, its 2 width tests, and its 2 partial tests. In the controlled study, they are the 4 settings of the accuracy correlation and the 2 cross-initialization settings. In the transfer analyses, they are the 5 datasets of the cross-site analysis, the 3 anchor counts of the link through the relative representation, and the 4 objective-family pairs of that link at the 3 anchor counts. On TAIX-Ray, they are the 90 presence, 144 severity, and 54 laterality classifiers of the decodability analysis and the 144 grade-ordering tests. In the analyses of agreement on chest radiographs, they are the 4 blocks of the variance decomposition, the 153 pairs of the manufacturer test, the 2 chest references and the 36 dermoscopy tests of the centroid analysis, and the 5 demographic attributes. The shares of the TAIX-Ray radiologists, the reader study, the cross-initialization comparisons per objective, the demographic and per-finding analyses, and the synthetic model are descriptive or exploratory and are reported as such. The four hypotheses were fixed as decision rules before the results were computed. Two rules were amended after their first computation. After the rules were fixed, we added the linear mapping on TAIX-Ray with its comparison against the classifier fitted on the new model, the fits of the mapping on fewer unlabeled radiographs, and the size analysis with the width held fixed. We also compared the public models between statistics, between modalities, and between groups of pairs, summarized retention and the cross-site change per model, summarized the transfer between controlled models by objective, and compared the two refits. The post hoc analyses further include the Spearman correlations of the agreement among the positive radiographs of each finding with the AUROC of its classifier, of each model's distance to the consensus with its mean agreement with the other models, and of the Procrustes distance with the mKNN in the synthetic model. In the reader study, the agreement of the two radiologists on the 100 shared triplets and the accuracy of the dataset labels per source dataset and per asked finding are post hoc as well. Supplementary Table~\ref{stab:rules} lists every rule with the values that decide it, each amendment, and each outcome.

\section*{Data availability}

All 14 datasets of this study are existing public releases. None of their images, reports, or derived records is redistributed here.

\begingroup
\footnotesize
\sloppy
\textbf{Chest radiography:}
\begin{itemize}
  \item MIMIC-CXR~\cite{Johnson2019MIMICCXR}, radiographs and reports, available from PhysioNet under credentialed access: \url{https://physionet.org/content/mimic-cxr-jpg/} and \url{https://physionet.org/content/mimic-cxr/}.
  \item MIMIC-IV~\cite{Johnson2023MIMICIV}, patient demographics, available from PhysioNet under credentialed access: \url{https://physionet.org/content/mimiciv/}.
  \item CheXpert~\cite{Irvin2019CheXpert} and CheXpert Plus~\cite{Chambon2024CheXpertPlus}, available from the Stanford AIMI portal after registration and acceptance of its terms: \url{https://aimi.stanford.edu/}.
  \item NIH ChestX-ray14~\cite{Wang2017ChestXray14}, openly available: \url{https://nihcc.app.box.com/v/ChestXray-NIHCC}.
  \item PadChest~\cite{Bustos2020PadChest}, available on request from the BIMCV repository: \url{https://bimcv.cipf.es/bimcv-projects/padchest/}.
  \item VinDr-CXR~\cite{Nguyen2022VinDrCXR}, available from PhysioNet under credentialed access: \url{https://physionet.org/content/vindr-cxr/}.
  \item VinDr-PCXR~\cite{Pham2022VinDrPCXR}, available from PhysioNet under credentialed access: \url{https://physionet.org/content/vindr-pcxr/}.
  \item TAIX-Ray~\cite{Truhn2026TAIXRay}, openly available on the Hugging Face Hub: \url{https://huggingface.co/datasets/TLAIM/TAIX-Ray}.
  \item ReXGradient-160K~\cite{Zhang2025Rexgradient}, provided by Gradient Health and available on the Hugging Face Hub after acceptance of its terms: \url{https://huggingface.co/datasets/rajpurkarlab/ReXGradient-160K}.
\end{itemize}
\textbf{Histopathology:}
\begin{itemize}
  \item NCT-CRC-HE-100K and CRC-VAL-HE-7K~\cite{Kather2018NCT}, openly available from Zenodo: \url{https://doi.org/10.5281/zenodo.1214456}.
  \item PatchCamelyon~\cite{Veeling2018PCam}, openly available: \url{https://github.com/basveeling/pcam}.
\end{itemize}
\textbf{Dermoscopy:}
\begin{itemize}
  \item ISIC 2019~\cite{Tschandl2018HAM10000,Combalia2019BCN20000,Codella2019ISIC}, openly available from the ISIC Archive: \url{https://challenge.isic-archive.com/}.
\end{itemize}
\textbf{Fundus photography:}
\begin{itemize}
  \item APTOS~\cite{Karthik2019APTOS}, available from Kaggle after acceptance of the competition rules: \url{https://www.kaggle.com/c/aptos2019-blindness-detection}.
  \item Messidor-2~\cite{Decenciere2014Messidor}, available on request from ADCIS: \url{https://www.adcis.net/en/third-party/messidor2/}.
\end{itemize}
\textbf{Mammography:}
\begin{itemize}
  \item VinDr-Mammo~\cite{Nguyen2023VinDrMammo}, available from PhysioNet under credentialed access: \url{https://physionet.org/content/vindr-mammo/}.
\end{itemize}
\textbf{Reference structure:}
\begin{itemize}
  \item The ICD-10 hierarchy, available from the US Centers for Medicare and Medicaid Services: \url{https://www.cms.gov/medicare/coding-billing/icd-10-codes}.
\end{itemize}
\endgroup

\section*{Code availability}

The analysis code is publicly available at \url{https://github.com/tayebiarasteh/convergence}. The repository contains the code for building the datasets, extracting the features, training the controlled models, computing the agreement and the classifier transfer between models, and computing the reported statistics and the outcomes of the pre-specified decision rules. The bootstrap and permutation seeds are fixed. It does not redistribute the public model weights or the underlying datasets.

All 25 public models are open-weight checkpoints and were run on local hardware, without any cloud service or third-party application programming interface. No closed model was used. Some of the released checkpoints require accepting their terms on the Hugging Face Hub. The controlled models were initialized from the DINOv3 ViT-S/16 and ViT-B/16 checkpoints or from random weights. The random network is a ViT-L/16 with random weights and has no released checkpoint. The models were accessed and the experiments were run between June and October 2026. The URLs of the checkpoints used in this study are:

\begingroup
\footnotesize
\sloppy
\textbf{Image models:}
\begin{itemize}
  \item RAD-DINO: \url{https://huggingface.co/microsoft/rad-dino}
  \item TorchXRayVision DenseNet-121 (densenet121-res224-all): \url{https://github.com/mlmed/torchxrayvision}
  \item BiomedCLIP (image and text encoders): \url{https://huggingface.co/microsoft/BiomedCLIP-PubMedBERT_256-vit_base_patch16_224}
  \item MedGemma 1.5 vision encoder: \url{https://huggingface.co/google/medgemma-1.5-4b-it}
  \item LLaVA-Med vision encoder: \url{https://huggingface.co/chaoyinshe/llava-med-v1.5-mistral-7b-hf}
  \item LLaVA-OneVision vision encoder: \url{https://huggingface.co/llava-hf/llava-onevision-qwen2-7b-ov-hf}
  \item DINOv3 ViT-L/16: \url{https://huggingface.co/facebook/dinov3-vitl16-pretrain-lvd1689m}
  \item DINOv2 ViT-L/14: \url{https://huggingface.co/facebook/dinov2-large}
  \item CLIP ViT-L/14: \url{https://huggingface.co/openai/clip-vit-large-patch14}
  \item SigLIP 2 ViT-L/16: \url{https://huggingface.co/google/siglip2-large-patch16-512}
  \item UNI: \url{https://huggingface.co/MahmoodLab/UNI}
  \item UNI2-h: \url{https://huggingface.co/MahmoodLab/UNI2-h}
  \item Virchow: \url{https://huggingface.co/paige-ai/Virchow}
  \item Virchow2: \url{https://huggingface.co/paige-ai/Virchow2}
  \item Prov-GigaPath: \url{https://huggingface.co/prov-gigapath/prov-gigapath}
  \item Phikon-v2: \url{https://huggingface.co/owkin/phikon-v2}
  \item CONCH (image and text encoders): \url{https://huggingface.co/MahmoodLab/CONCH}
  \item RETFound: \url{https://huggingface.co/iszt/RETFound_mae_meh}
\end{itemize}
\textbf{Text models:}
\begin{itemize}
  \item PubMedBERT: \url{https://huggingface.co/microsoft/BiomedNLP-BiomedBERT-base-uncased-abstract-fulltext}
  \item MedCPT query encoder: \url{https://huggingface.co/ncbi/MedCPT-Query-Encoder}
  \item MedCPT article encoder: \url{https://huggingface.co/ncbi/MedCPT-Article-Encoder}
  \item MedGemma 27B: \url{https://huggingface.co/google/medgemma-27b-it}
  \item SapBERT: \url{https://huggingface.co/cambridgeltl/SapBERT-from-PubMedBERT-fulltext}
\end{itemize}
\textbf{Starting checkpoints of the controlled models:}
\begin{itemize}
  \item DINOv3 ViT-S/16: \url{https://huggingface.co/facebook/dinov3-vits16-pretrain-lvd1689m}
  \item DINOv3 ViT-B/16: \url{https://huggingface.co/facebook/dinov3-vitb16-pretrain-lvd1689m}
  \item Bio\_ClinicalBERT, the text encoder of the image-report models of the second chest setting: \url{https://huggingface.co/emilyalsentzer/Bio_ClinicalBERT}
\end{itemize}
\endgroup

The study used Python~3.11 with PyTorch~2.9 (CUDA~13.0) and torchvision~0.24. The models were loaded with Hugging Face Transformers~5.0, huggingface-hub~1.3, safetensors~0.7, timm~1.0, OpenCLIP~3.3, and TorchXRayVision~1.0. MedGemma 27B was run in 4-bit quantization with bitsandbytes~0.49. The nearest-neighbor searches used faiss~1.9. The data handling, the statistical procedures, and the figures used NumPy~1.26, SciPy~1.17, pandas~3.0, scikit-learn~1.8, h5py~3.16, OpenCV~4.13, Pillow~12.2, Matplotlib~3.10, PyYAML~6.0, and tqdm~4.67. Feature extraction and the controlled training ran on NVIDIA L40S GPUs with 48~GB of memory and Intel Xeon Silver 4310 CPUs.

\section*{Acknowledgements}

{\emergencystretch=2em STA is supported by the German Federal Ministry of Research, Technology and Space (ARISTOTLE, 01ZU2602B) and the Excellence Strategy of the German Federal Government, the L\"ander, and RWTH ERS (START\_526-26). SN is supported by the DFG (701010997, 517243167). JNK is supported by the German Cancer Aid (DECADE, 70115166), the German Federal Ministry of Education and Research (PEARL, 01KD2104C; CAMINO, 01EO2101; SWAG, 01KD2215A; TRANSFORM LIVER, 031L0312A; TANGERINE, 01KT2302 through ERA-NET Transcan; Come 2Data, 16DKZ2044A; DEEP-HCC, 031L0315A), the German Academic Exchange Service (SECAI, 57616814), the German Federal Joint Committee (TransplantKI, 01VSF21048) the European Union's Horizon Europe and innovation programme (ODELIA, 101057091; GENIAL, 101096312), the European Research Council (ERC; NADIR, 101114631), the National Institutes of Health (EPICO, R01 CA263318) and the National Institute for Health and Care Research (NIHR, NIHR 203331) Leeds Biomedical Research Centre. The views expressed are those of the author(s) and not necessarily those of the NHS, the NIHR or the Department of Health and Social Care. This work was funded by the European Union. Views and opinions expressed are however those of the author(s) only and do not necessarily reflect those of the European Union. Neither the European Union nor the granting authority can be held responsible for them. DT is supported by the German Ministry of Research, Technology and Space (TRANSFORM LIVER - 031L0312C, DECIPHER-M - 01KD2420B), DFG (515639690), and the European Union (Horizon Europe, ODELIA - GA 101057091, ERC Starting Grant SAGMA -- GA 101222556).\par}

\section*{Author contributions}

The formal analysis was conducted by STA and DT. The original draft was written by STA and edited by STA and DT. STA and ML developed the code. The experiments were performed by STA. The statistical analyses were performed by STA and DT. SZ and LA performed the reader studies. SZ, LA, SN, JNK, and DT provided clinical expertise. STA, ML, JNK, and DT provided technical expertise. The study was defined by STA. All authors read the manuscript and agreed to the submission of this paper.

\section*{Competing interests}

STA is on the editorial board of Communications Medicine and of European Radiology Experimental, and on the trainee editorial board of Radiology: Artificial Intelligence. ML is employed by Generali Deutschland Services GmbH, Germany, and is on the editorial board of European Radiology Experimental. LA is on the trainee editorial board of Radiology: Artificial Intelligence. JNK holds shares in StratifAI, Synagen, Spira Labs, Tremont AI, and Saterra AI; is Co-PI on institutional research grants from GSK and AstraZeneca; and declares honoraria or consulting fees from AstraZeneca, Bayer, Bioptimus, Daiichi Sankyo, Eisai, Janssen, Merck, MSD, Novartis, BMS, Roche, and Pfizer. DT received honoraria for lectures from Bayer, GE, Roche, AstraZeneca, and Philips and holds shares in StratifAI GmbH, Germany, and in Synagen GmbH, Germany. The authors were not involved in the journal's review of, or decisions related to, this manuscript. The authors declare no other competing financial or non-financial interests.

\bibliographystyle{splncs04}
\bibliography{bibliography}

\setcounter{table}{0}
\setcounter{figure}{0}
\setcounter{equation}{0}
\renewcommand{\tablename}{Supplementary Table}
\renewcommand{\figurename}{Supplementary Fig.}
\floatname{algorithm}{Supplementary Algorithm}
\renewcommand{\thealgorithm}{\arabic{algorithm}}
\renewcommand{\theequation}{S\arabic{equation}}
\renewcommand{\theHequation}{S\arabic{equation}}

\section*{Supplementary information}

\newcounter{snote}\setcounter{snote}{0}

\refstepcounter{snote}
\section*{Supplementary Note \thesnote: The agreement statistics and their behavior with the neighbor count}
\label{snote:statistics}

The mutual $k$-nearest-neighbor overlap (mKNN)~\cite{Huh2024Platonic} is computed on L2-normalized embeddings under cosine distance, with each case excluded from its own neighborhood. The centered kernel nearest-neighbor alignment (CKNNA)~\cite{Huh2024Platonic} applies the unbiased estimator of the Hilbert-Schmidt independence criterion~\cite{Song2012HSIC} to the two cosine kernels $K = X_f X_f^{\top}$ and $L = X_g X_g^{\top}$, each masked by the indicator of the intersection of the two $k$-nearest-neighbor sets. The value is normalized by the square root of the product of the same estimator applied to each kernel masked by its own neighborhood. The binary-graph variant of CKNNA uses each model's binary neighbor graph in place of its masked kernel, no intersection mask, and the centered trace of the product of the two graphs. It is a function of the same binary graphs as mKNN. Its correlation with mKNN is therefore close to one by construction. It enters no result of this paper.

On the full image sets, CKNNA orders the five modalities as mKNN does. Under CKNNA, the 95\% confidence interval of the 18 models lies above that of the random network in chest radiographs, histopathology, and dermoscopy. The two intervals overlap for fundus photographs (0.133 $\pm$ 0.015 [0.104, 0.163] against 0.094 $\pm$ 0.012 [0.071, 0.117]) and for mammograms (0.071 $\pm$ 0.008 [0.055, 0.087] against 0.047 $\pm$ 0.007 [0.032, 0.061]).

\begin{figure*}[p]
\centering
\includegraphics[width=\textwidth]{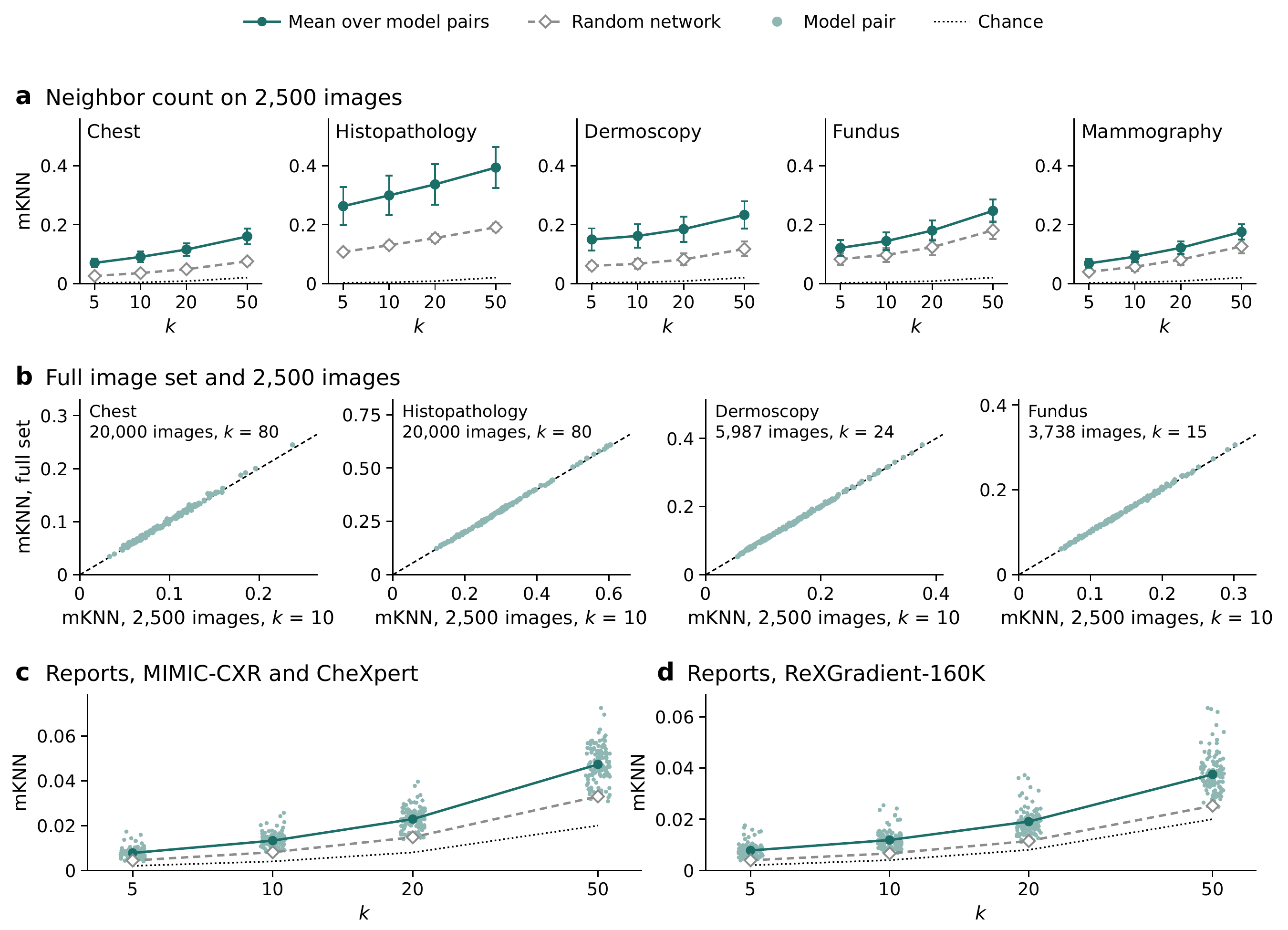}
\caption{Agreement between the 18 public models against the neighbor count $k$ and the number of images. Teal markers on a solid line are means over the model pairs, and diamonds on a dashed line mark the random network. The dotted line marks the chance level, $k$ divided by the 2{,}500 images. In \textbf{a}, the error bar spans the 95\% confidence interval, the mean $\pm$ 1.96 standard errors of the delete-one-model jackknife. No other panel draws an error bar. Counts per panel: 153 model pairs and 18 random-network pairs per modality and $k$ (\textbf{a}, \textbf{b}), and 126 image-report pairs and 7 random-network pairs per $k$ (\textbf{c}, \textbf{d}). \textbf{a}, mKNN on 2{,}500 images at $k$ of 5, 10, 20, and 50, one facet per modality. \textbf{b}, mKNN of each model pair on the full image set, with $k$ at 0.4\% of the images, against its mKNN on 2{,}500 images at $k = 10$. The dashed line is the identity. The mammography set has 2{,}500 images and is not drawn, since its two values coincide. \textbf{c},\textbf{d}, mKNN between each image model on 2{,}500 radiographs and each text model on their reports at each $k$, on MIMIC-CXR and CheXpert (\textbf{c}) and on ReXGradient-160K (\textbf{d}), with the random image network for comparison. mKNN, mutual $k$-nearest-neighbor overlap.}
\label{sfig:fixedk}
\end{figure*}

At a fixed pool size, the mKNN rises from $k$ = 10 to $k$ = 50 in every pool (Supplementary Fig.~\ref{sfig:fixedk}a). At a fixed $k$, a neighborhood statistic falls as the pool grows while the relation between the two models is unchanged~\cite{Koepke2026PlatosCave}, because the $k$ neighbors of a case are drawn from a denser sample. The image-report agreement is above that of the random image network at every $k$ from 5 to 50 (Supplementary Fig.~\ref{sfig:fixedk}c,d). Four further statistics were computed on the same 2,500 cases as the mKNN of each pair: linear centered kernel alignment (CKA) through the feature-space identity on column-centered features~\cite{Kornblith2019CKA}, radial-basis-function (RBF) CKA on a 2,000-case subsample~\cite{Kornblith2019CKA}, the orthogonal Procrustes distance~\cite{Schonemann1966Procrustes}, and representational similarity analysis (RSA)~\cite{KriegeskorteKievit2013} as the Spearman correlation~\cite{Spearman1904Correlation} of the two pairwise-distance vectors on a 2,000-case subsample. RBF CKA uses the median distance between the features of the first model as the bandwidth of both kernels and sets the diagonals of the kernels to zero. The Procrustes distance is the root mean square difference between the features of one model and the orthogonally rotated features of the other model. The features are not centered. The narrower features are padded with zeros. Their means per modality are in Fig.~\ref{efig:pairs}b. None of them orders the five modalities as mKNN does. With nonoverlapping 95\% confidence intervals, the 18 models agree more than the random network in five modalities under linear CKA, three under RBF CKA, four under RSA, and one under the Procrustes distance. Over the 153 pairs of a modality, the Spearman correlation with mKNN is 0.705 to 0.855 for linear CKA, 0.402 to 0.871 for RSA, 0.423 to 0.654 for RBF CKA, and $-0.019$ to 0.377 for the negated Procrustes distance (Fig.~\ref{efig:pairs}c). For RBF CKA and the Procrustes distance, the $p$ after Benjamini-Hochberg false discovery rate (FDR) correction, $p_{\mathrm{FDR}}$, is at least 0.05 in three and in four of the five modalities. Under the five statistics other than mKNN, TorchXRayVision is among the four least aligned models in 21 of the 25 combinations of statistic and modality. In every modality, its 17 pairs have a lower mean mKNN than the pairs without it (Fig.~\ref{efig:pairs}e,f).

For a pair of controlled models and for an image model with a text model, the point estimate is the mean over the 200 draws. It therefore lies inside its own interval by construction. The delete-one-model jackknife~\cite{Efron1979Bootstrap} recomputes a summary over model pairs $E$ times, each time without one model and its pairs. Its standard error is $\sqrt{(E - 1) / E \cdot \sum_e (\theta_{(e)} - \bar{\theta})^2}$ over the recomputed values $\theta_{(e)}$.

\refstepcounter{snote}
\section*{Supplementary Note \thesnote: Controlled training, its verification, and agreement across objectives, sizes, and reports}
\label{snote:controlled}

Outside the second chest setting, every model of the controlled study (Table~\ref{tab:design}) has a budget of 20,000 optimizer steps. The number of epochs is the number of batches for 20,000 steps divided by the batches per epoch, rounded up. Because the gradient of an odd last batch is discarded at the start of every epoch, training stopped at 19,712 steps in histopathology and at 19,950 steps on the first set of training patients.

In the verification of the objectives (Fig.~\ref{efig:start}a), the scrambled target is the labels for label supervision, the report for image-report contrastive learning~\cite{Radford2021CLIP}, the second view for self-distillation~\cite{Caron2021DINO} and image-image contrastive learning~\cite{Chen2020SimCLR}, and the reconstruction target for masked image modeling. Each is permuted within the batch while the images stay unchanged. The seed was reset before scoring, so that the two models and the untrained initialization were scored on identical batches, crops, and masks. For masked image modeling, we also checked for every model that the output of its encoder does not change when the pixels of the masked patches are replaced. The masked pixels therefore cannot reach the encoder.

Between the models of two different objectives, the mean agreement is 0.028 to 0.184 on chest radiographs, 0.211 to 0.394 on histopathology patches, and 0.064 to 0.132 in the second chest setting (Fig.~\ref{efig:start}b). On chest radiographs, the two contrastive objectives agree with each other at 0.067 with the small vision transformer (ViT-S) and at 0.058 with the base vision transformer (ViT-B). Image-report contrastive learning and label supervision agree at 0.082 and 0.072 at the two sizes. Between the ViT-S and the ViT-B model of one objective, the agreement on chest radiographs is 0.497 under image-report contrastive learning, 0.482 under image-image contrastive learning, 0.423 under self-distillation, 0.354 under masked image modeling, and 0.057 under label supervision (Fig.~\ref{fig:matrix}b). On histopathology patches, it is 0.435 under masked image modeling, 0.416 under image-image contrastive learning, 0.403 under self-distillation, and 0.288 under label supervision. In the second chest setting, it is 0.697 under image-image contrastive learning, 0.525 under image-report contrastive learning, and 0.067 under label supervision. On chest radiographs, a ViT-S and a ViT-B model of different objectives agree at 0.087 on average.

In the second chest setting, image-report models were also trained on reports with a demographic sentence added or with the demographic words removed. At ViT-S, the model with the sentence added agrees with the natural-report model of the same seed at 0.581. The model with the words removed agrees with it at 0.700. Two natural-report models of different seeds agree at 0.668. At ViT-B, the two edited models agree with the natural-report model of the same seed at 0.527 and 0.656 (Fig.~\ref{efig:start}g).

\refstepcounter{snote}
\section*{Supplementary Note \thesnote: Classifier transfer per model, refits on the relative representation, and permutation references}
\label{snote:transfer}

Through the linear mapping, mean retention is 0.934 to 1.045 when a model is the old model and 0.989 to 1.051 when it is the new model (Fig.~\ref{efig:transfer}b). The classifiers of TorchXRayVision have the lowest retention. It is below 1 after transfer to each of the other 17 models. Refitting a classifier on the relative representation~\cite{Moschella2023Relative} of the new model instead of its embedding lowers the area under the receiver operating characteristic curve (AUROC) by 0.021 $\pm$ 0.001 [0.019, 0.024] with 50 labels, by 0.017 $\pm$ 0.002 [0.014, 0.021] with 200 labels, and by 0.014 $\pm$ 0.002 [0.010, 0.019] with 1,000 labels (Fig.~\ref{efig:transfer}f). Among the classifiers fitted on MIMIC-CXR in the relative representation, those of RAD-DINO, TorchXRayVision, MedGemma, and BiomedCLIP lose the least AUROC or gain the most on four of the five other chest datasets (Fig.~\ref{efig:transfer}c).

Inside the controlled study, classifiers for the 13 findings other than no finding were transferred through the relative representation over 1,024 anchors between the 35 chest models, over 1,190 ordered pairs. The chance-corrected retention (Eq.~\ref{eq:retention}) depends on the objective of the model that receives the classifier. It is 0.629 $\pm$ 0.034 [0.562, 0.696] when the receiving model was trained with self-distillation, 0.621 $\pm$ 0.040 [0.543, 0.699] when it was trained with image-report contrastive learning, and 0.597 $\pm$ 0.028 [0.543, 0.652] when it was trained with label supervision. With masked image modeling and image-image contrastive learning, it is 0.422 $\pm$ 0.045 [0.334, 0.510] and 0.408 $\pm$ 0.029 [0.351, 0.465]. Over the pairs of two models of one objective, it is 0.582 $\pm$ 0.023 [0.536, 0.628], and over the pairs of two objectives, it is 0.526 $\pm$ 0.019 [0.488, 0.563]. Two models of one objective retain 0.525 and 0.541 under image-image contrastive learning and masked image modeling, and 0.599 to 0.626 under self-distillation, image-report contrastive learning, and label supervision. Each value is a mean over the ordered pairs and findings with the delete-one-model jackknife.

On TAIX-Ray, the laterality AUROC of a classifier transferred through the linear mapping is correlated with that of the classifier fitted on the new model (Spearman 0.732 over the 918 combinations of an ordered model pair and a finding, jackknife $p < 0.001$; Fig.~\ref{fig:detail}e). Through the linear mapping, the transferred classifiers exceed their label-permutation reference in every presence and severity combination and in 888 of the 918 laterality combinations. Through the relative representation, they exceed it in 1,346 of the 1,530 presence, 1,724 of the 2,448 severity, and 354 of the 918 laterality combinations (Fig.~\ref{fig:detail}f).

\refstepcounter{snote}
\section*{Supplementary Note \thesnote: The consensus configuration and the reference for the co-occurrence test}
\label{snote:consensus}

\begin{figure*}[p]
\centering
\includegraphics[width=\textwidth]{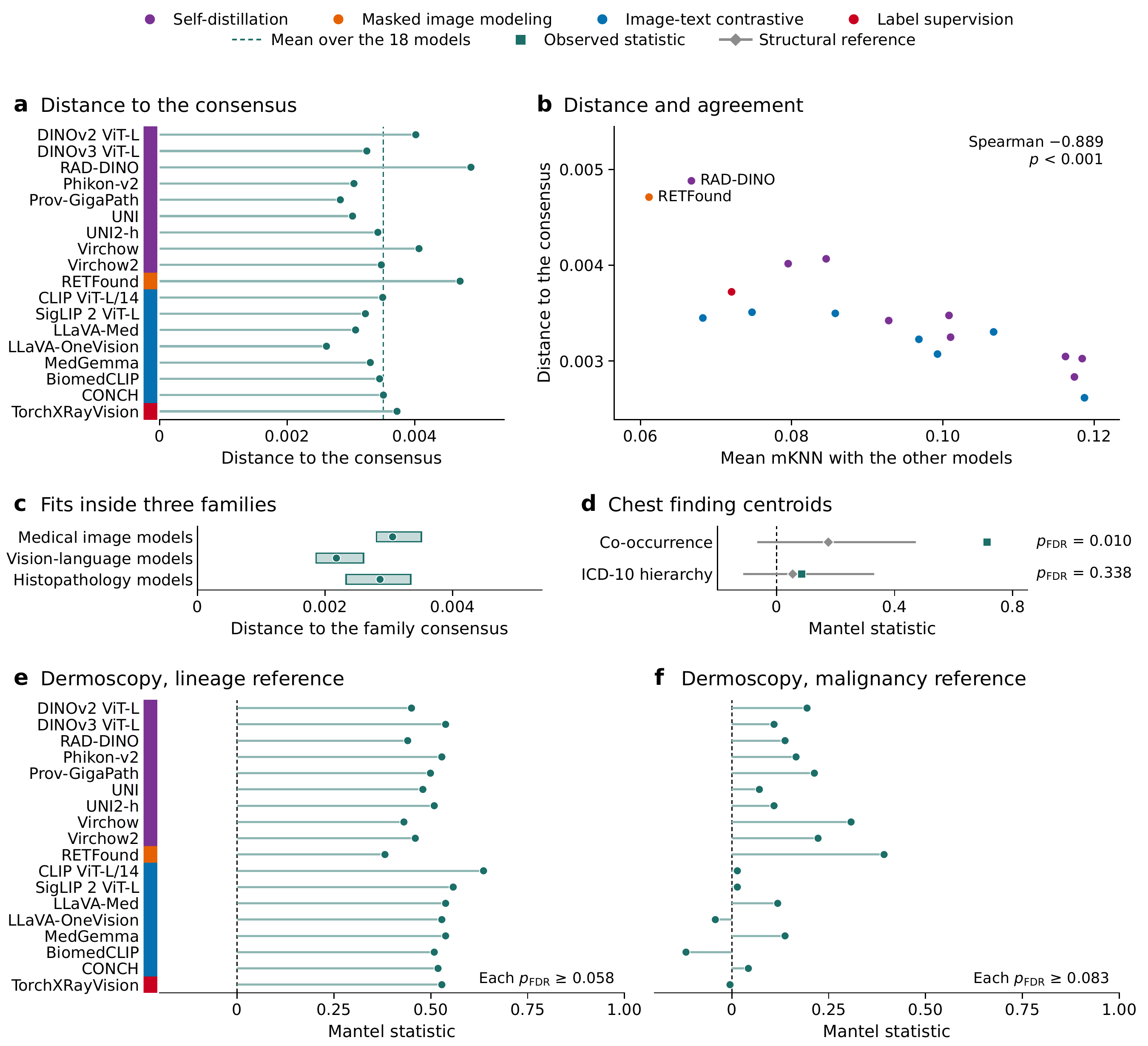}
\caption{The consensus configuration of the 18 public models on 48{,}554 chest radiographs and the structure of its class centroids. Color marks the pretraining objective, in the strip beside each name and in \textbf{b}. The teal dashed line marks the mean over the 18 models (\textbf{a}) and the black dashed line marks zero. Counts per panel: 18 models (\textbf{a}, \textbf{b}, \textbf{e}, \textbf{f}) and families of 3, 3, and 7 models (\textbf{c}). \textbf{a}, Procrustes distance of each model to the consensus, divided by the square root of the number of radiographs. \textbf{b}, Distance to the consensus against the mean mKNN of the model with the other 17 on 20{,}000 chest radiographs at $k = 80$. \textbf{c}, Separate fits inside three families: RAD-DINO, BiomedCLIP, and TorchXRayVision; MedGemma, LLaVA-Med, and LLaVA-OneVision; and the seven histopathology models. The bar spans the range of the distances to the family consensus and the dot marks their mean. \textbf{d}, Mantel statistic between the distances of the 14 finding centroids and each reference (square) and its distribution under 200 permutations of whole label rows (diamond at the mean, bar from the 2.5th to the 97.5th percentile). Each $p_{\mathrm{FDR}}$ compares the statistic with this reference, corrected over the two references. \textbf{e},\textbf{f}, Mantel statistic between the distances of the 8 diagnosis centroids of each model on 5{,}987 dermoscopy images and the lineage (\textbf{e}) or malignancy reference (\textbf{f}), with $p_{\mathrm{FDR}}$ from 1{,}000 permutations over the 36 tests. FDR, false discovery rate; mKNN, mutual $k$-nearest-neighbor overlap.}
\label{sfig:consensus}
\end{figure*}

The consensus configuration is a generalized Procrustes fit~\cite{Gower1975GPA} over the relative representation (Eq.~\ref{eq:anchor}) of the 18 public models over 1,024 anchors on 48,554 chest radiographs, the finite rows of a 50,000-radiograph subsample that contains the 4,096 anchor radiographs. Each configuration and the consensus are column-centered and scaled to unit Frobenius norm before the fit. The residual of a model is its Procrustes distance to the consensus divided by the square root of the number of radiographs. The fit is iterated until the change in the mean distance of the models to the consensus falls below $10^{-6}$, for at most 2,000 iterations. It converged at the fourth iteration, with a change of $3.2 \times 10^{-7}$. The distance of a model to the consensus is 0.0026 to 0.0049 (Supplementary Fig.~\ref{sfig:consensus}a). In a post hoc analysis, the models farther from the consensus agree less with the other 17 models on the chest pool (Spearman $-0.889$, $p < 0.001$; Supplementary Fig.~\ref{sfig:consensus}b). At every iteration, the mean of the rotated configurations is aligned back onto the previous consensus, so that a free rotation of the mean does not count as a change. The same fit was run alone for the three families of at least three models, which are RAD-DINO, BiomedCLIP, and TorchXRayVision; the vision encoders of the three vision-language models; and the seven histopathology models. Each converged in three iterations, with mean distances to the family consensus of 0.0031, 0.0022, and 0.0029 (Supplementary Fig.~\ref{sfig:consensus}c).

In the co-occurrence test, we correlated the 91 pairwise cosine distances between the 14 finding centroids of the consensus with the distance of the findings in the ICD-10 hierarchy and with one minus their Jaccard co-occurrence in the MIMIC-CXR radiographs of the pool. Each centroid is the mean over the positive radiographs of a finding. The Mantel statistic~\cite{Mantel1967} was tested with 1,000 permutations of the findings. A multilabel radiograph enters the centroid of each of its positive findings, and the same labels define the co-occurrence. A correlation between the centroid distances and the co-occurrence could therefore arise from case overlap without any image content. Under the structural reference, only this overlap can produce a correlation. Whole label rows are permuted across radiographs 200 times, which keeps the prevalence and the co-occurrence of the findings and breaks only the tie between a radiograph and its label set. The Mantel statistic is recomputed on each permutation, and the observed value is compared with the reference distribution (Supplementary Fig.~\ref{sfig:consensus}d). For the co-occurrence, the Mantel statistic of the consensus is 0.713, and its mean under the structural reference is 0.175 ($p_{\mathrm{FDR}} = 0.010$). For the ICD-10 hierarchy, the Mantel statistic of the consensus is 0.085, and its mean under the structural reference is 0.055 ($p_{\mathrm{FDR}} = 0.338$). On the dermoscopy pool of ISIC 2019~\cite{Tschandl2018HAM10000,Combalia2019BCN20000,Codella2019ISIC}, every image has one diagnosis. No image enters two centroids, and case overlap is absent by construction. There, the eight class centroids of each model's own embedding were correlated in the same way with two references declared before the analysis, with FDR correction over the 36 tests (Supplementary Fig.~\ref{sfig:consensus}e,f). In the lineage reference, melanoma and nevus are melanocytic; basal cell carcinoma, squamous cell carcinoma, actinic keratosis, and benign keratosis are keratinocytic; and dermatofibroma and vascular lesion are other. In the malignancy reference, melanoma, basal cell carcinoma, and squamous cell carcinoma are malignant; actinic keratosis is premalignant; and nevus, benign keratosis, dermatofibroma, and vascular lesion are benign. Two classes of one group are at distance 1, and two classes of different groups are at distance 2. The Mantel statistic of a model is 0.382 to 0.636 for the lineage reference and $-0.118$ to 0.393 for the malignancy reference. None is significant after FDR correction, with a smallest $p_{\mathrm{FDR}}$ of 0.058 and 0.083.

\refstepcounter{snote}
\section*{Supplementary Note \thesnote: Label harmonization}
\label{snote:data}

MIMIC-CXR~\cite{Johnson2019MIMICCXR} and CheXpert~\cite{Irvin2019CheXpert} release the output of the CheXpert labeler for all 14 findings. The labels of NIH ChestX-ray14~\cite{Wang2017ChestXray14}, PadChest~\cite{Bustos2020PadChest}, VinDr-CXR~\cite{Nguyen2022VinDrCXR}, and VinDr-PCXR~\cite{Pham2022VinDrPCXR} were mapped onto the 14 findings by name. Where the names differ, pleural thickening was mapped to pleural other, nodule and mass to lung lesion, effusion to pleural effusion, pulmonary edema to edema, rib and clavicle fractures to fracture, and pacemaker to support devices. Pleural thickening, rib and clavicle fractures, and pacemakers are narrower than pleural other, fracture, and support devices. In the sources that label them, the prevalence of these three findings is the prevalence of the narrower label. A label is missing where a source does not label the finding and where the CheXpert labeler marks a mention as uncertain. Per finding, 0\% to 8\% of the MIMIC-CXR radiographs and 0\% to 13\% of the CheXpert radiographs have an uncertain mention and therefore no label. The other sources have no missing label among their labeled findings. The positive count and the prevalence of each finding in each source after the mapping are in Supplementary Table~\ref{stab:label_mapping}. Every radiograph of the chest pool has an identifier composed of the source name and the source's own image key. Every analysis that subsamples the pool resolves its radiographs through these identifiers.

Only the binary presence columns of TAIX-Ray~\cite{Truhn2026TAIXRay} were mapped onto its five presence findings. Its graded columns were read separately, since four of them share their names with presence findings. The congestion label is used in place of edema, since pulmonary congestion is the early stage of cardiogenic pulmonary edema and the synonyms of edema in the CheXpert labeler include vascular congestion~\cite{Irvin2019CheXpert}. TAIX-Ray marks cardiomegaly as present from grade 2. It marks congestion, pleural effusion, lung opacity, and atelectasis as present only from grade 3. The laterality labels of pleural effusion, lung opacity, and atelectasis were read from the released binary columns of the left and right sides, which mark a side as positive from grade 3. The laterality contrast takes the radiographs positive on one side only, at matched grade. In the ReXGradient-160K pool~\cite{Zhang2025Rexgradient}, a study with more than one recorded manufacturer is its own level of the manufacturer variable.

\refstepcounter{snote}
\section*{Supplementary Note \thesnote: The synthetic model and the positive control}
\label{snote:synthetic}

We built a synthetic generative model with labels of controlled informativeness. With it, we tested the prediction that two models of an objective that retains the variation of its input agree with each other at every informativeness of the labels. Under the same prediction, two models trained to predict a target agree only as far as the target is a function of the signal. The signal latent $s \in \mathbb{R}^{32}$ and the distractor latent $z \in \mathbb{R}^{128}$ are standard normal. A sample $x \in \mathbb{R}^{160}$, its class label $y$, and its continuous target $t$ are
\begin{equation}\label{eq:synthetic}
x = \gamma\, s A_{\mathrm{sig}} + z A_{\mathrm{noi}} + \varepsilon, \qquad y = \arg\max_{c}\big(\alpha\, s W_{c} + (1 - \alpha)\, \eta_{c}\big), \qquad t = \alpha\, s w + (1 - \alpha)\, \xi,
\end{equation}
where $A_{\mathrm{sig}}$ and $A_{\mathrm{noi}}$ have unit-norm columns, $\varepsilon$ is Gaussian noise with a standard deviation of 0.1, and $\gamma$ is the signal-to-noise ratio in $\{0.5, 1, 2\}$. The matrix $W$ has one unit-norm prototype per class for 10 classes, $w$ is a unit-norm vector, and $\eta$ and $\xi$ are standard normal. Both targets are pure noise at an informativeness $\alpha$ of 0 and deterministic functions of the signal at $\alpha = 1$. The informativeness takes the values 0, 0.1, 0.25, 0.5, 0.75, and 1. For each seed, $A_{\mathrm{sig}}$, $A_{\mathrm{noi}}$, $W$, and $w$ are drawn once. Every sample of every condition with this seed is generated through the same draw. Every random draw and every initialization has a fixed seed. The 90 conditions are the combinations of the 6 values of $\alpha$, the 3 signal-to-noise ratios, and 5 seeds. In every condition, two training samples of 2,000 were drawn. On each sample, one two-layer perceptron with a hidden width of 64 and a linear head was trained per objective, with Adam~\cite{KingmaBa2015Adam} at a learning rate of $3 \times 10^{-3}$ for 200 full-batch steps. The four objectives are reconstruction of $x$, denoising of $x$ corrupted by Gaussian noise with a standard deviation of 0.5, classification of $y$, and regression of $t$. Reconstruction and denoising retain the variation of their input, and classification and regression predict a target. In every pair, one model was trained on the first training sample and the other model was trained on the second, under the same objective or under two different objectives. The embedding of a test sample is the output of the perceptron before the head, scaled to unit length. Two models were compared on the embeddings of the same 500 test samples by the mKNN at $k = 10$~\cite{Huh2024Platonic} and by the orthogonal Procrustes distance~\cite{Schonemann1966Procrustes}.

\begin{figure*}[p]
\centering
\includegraphics[width=\textwidth]{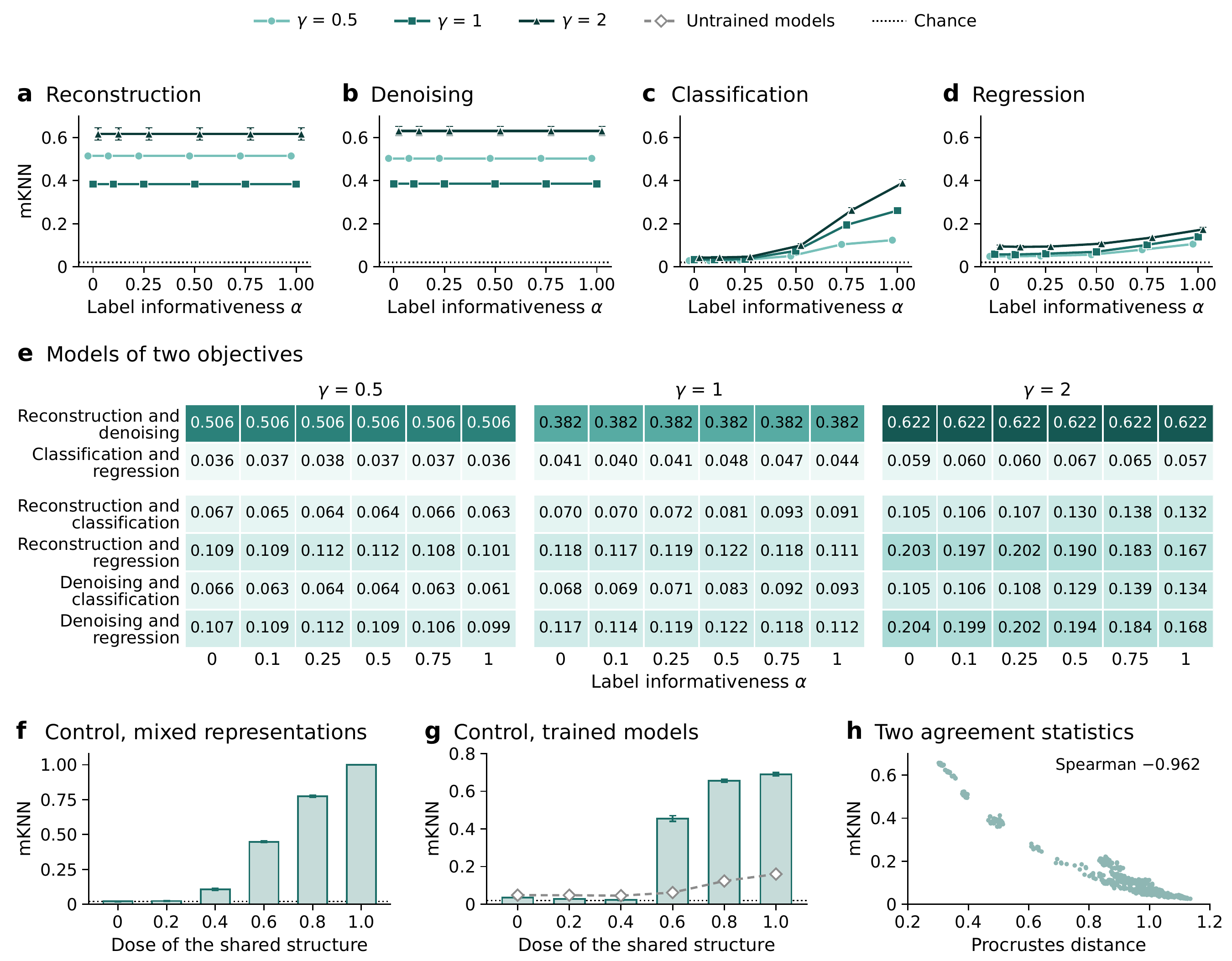}
\caption{Agreement between small models trained on the synthetic generative model, and the positive control. Two-layer perceptrons were trained on samples drawn as in Eq.~\ref{eq:synthetic}, with label informativeness $\alpha$ and signal-to-noise ratio $\gamma$. Agreement is the mKNN on 500 test samples at $k = 10$. Reconstruction and denoising retain the variation of their input, and classification and regression predict a target. In \textbf{a} to \textbf{d}, color shade and marker mark $\gamma$. The markers of one $\alpha$ are offset horizontally. The dotted line marks the chance level of 0.020. In \textbf{a} to \textbf{d}, \textbf{f}, and \textbf{g}, a marker or a bar is the mean over five seeds. Its error bar spans one standard deviation on each side of the mean. Panels \textbf{e} and \textbf{h} draw no error bar. Counts per panel: 5 model pairs per value of $\alpha$ and $\gamma$ (\textbf{a} to \textbf{e}), 5 pairs of representations or of models per dose (\textbf{f}, \textbf{g}), and the 900 model pairs of \textbf{a} to \textbf{e} (\textbf{h}). \textbf{a}--\textbf{d}, Agreement between two models of one objective against $\alpha$: reconstruction (\textbf{a}), denoising (\textbf{b}), classification (\textbf{c}), and regression (\textbf{d}). \textbf{e}, Agreement between models of two objectives, with one block of columns per $\gamma$. Each cell is the mean over five seeds. The first two rows pair two objectives of one group, and the other four rows pair objectives of different groups. \textbf{f}, Agreement of two random representations that weight one shared structure by the dose and a separate random part by one minus the dose. \textbf{g}, The same for two models trained on samples built this way, and for two untrained models on the same test inputs (diamonds on a dashed line). \textbf{h}, mKNN against the Procrustes distance for every model pair of \textbf{a} to \textbf{e}. The Spearman correlation over the 900 pairs is printed in the panel. mKNN, mutual $k$-nearest-neighbor overlap.}
\label{sfig:synthetic}
\end{figure*}

On the 500 test samples, the chance level of the mKNN is 0.020. Over the 900 model pairs, the Spearman correlation between the Procrustes distance and the mKNN is $-0.962$ (Supplementary Fig.~\ref{sfig:synthetic}h). Reconstruction and denoising do not use the labels. Across the signal-to-noise ratios, two models of either objective agree at 0.384 to 0.631 (Supplementary Fig.~\ref{sfig:synthetic}a,b). A reconstruction model agrees with a denoising model at 0.382 to 0.622 (Supplementary Fig.~\ref{sfig:synthetic}e). Under classification, the agreement is 0.028 to 0.042 at $\alpha$ = 0 and 0.123 to 0.389 at $\alpha$ = 1 across the signal-to-noise ratios (Supplementary Fig.~\ref{sfig:synthetic}c). Under regression, it is 0.048 to 0.094 at $\alpha$ = 0 and 0.105 to 0.173 at $\alpha$ = 1 (Supplementary Fig.~\ref{sfig:synthetic}d). A classification model agrees with a regression model at 0.036 to 0.067 at every $\alpha$. A reconstruction or denoising model agrees with a classification or regression model at 0.061 to 0.204 (Supplementary Fig.~\ref{sfig:synthetic}e). The prediction therefore holds inside the synthetic model. In the controlled study, masked image modeling retains the variation of its input. Two masked-image-modeling models agree between seeds at 0.411 on chest radiographs at ViT-S, and two label-supervised models agree at 0.062 (Fig.~\ref{fig:matrix}a). No claim of the paper depends on the synthetic model.

In the positive control, we added a shared neighborhood structure at a known dose (Supplementary Fig.~\ref{sfig:synthetic}f,g). Every negative result about agreement is interpreted only after this control has passed. At the level of the statistic, each of two representations of 500 samples is the sum of one shared random representation multiplied by the dose and a private random representation multiplied by one minus the dose. The agreement of the two representations is at the chance level of 0.020 at dose 0 and is 0.023, 0.106, 0.448, and 0.774 at doses of 0.2, 0.4, 0.6, and 0.8. At dose 1, the agreement is 1.000, since the two representations are identical. At the level of trained models, two models are trained by reconstruction. Their generative map consists of one shared block and one private block per model, which are mutually orthogonal parts of one orthonormal basis per seed. The training samples of each model are generated through the shared block with a weight equal to the dose and through its private block with a weight of one minus the dose. Both models embed the same 500 test samples, which are generated through the shared block and both private blocks. At doses of 0 to 0.4, two trained models agree at 0.023 to 0.035. Two untrained models agree at 0.045 to 0.048 on the same test inputs. At doses of 0.6 to 1, the trained models agree at 0.455 to 0.691, and the untrained models agree at 0.062 to 0.160. The control passes at both levels under the rule in Supplementary Table~\ref{stab:rules}.

\refstepcounter{snote}
\section*{Supplementary Note \thesnote: Findings, subgroups, and the TAIX-Ray radiologists}
\label{snote:readers}

\begin{figure*}[p]
\centering
\includegraphics[width=\textwidth]{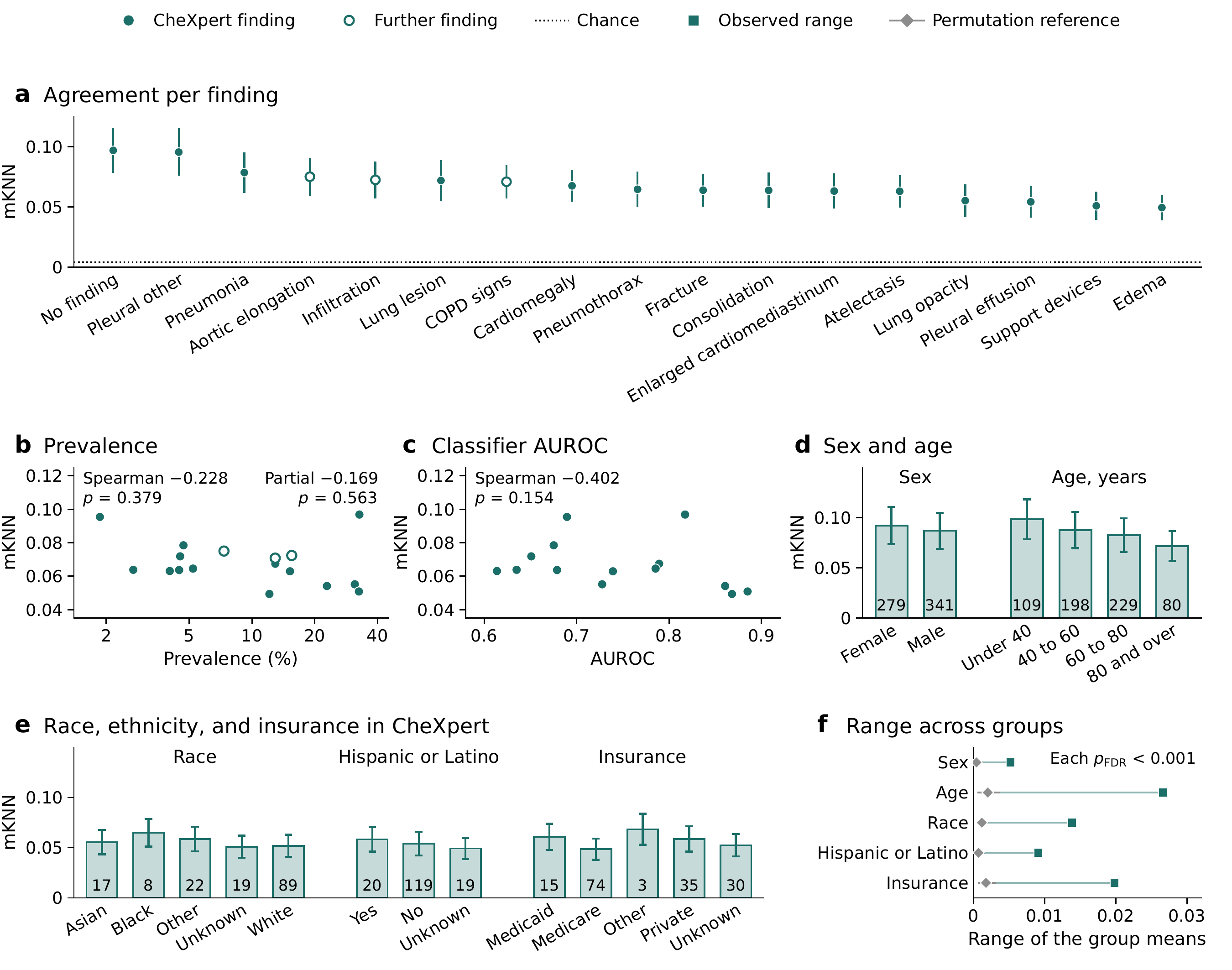}
\caption{Agreement between the 18 public models inside subsets of the chest pool, at matched numbers of radiographs. Agreement is the mKNN, averaged over the 153 model pairs. Filled circles mark the 14 findings of the CheXpert vocabulary and open circles mark 3 further findings. In \textbf{a}, the dotted line marks the chance level, $k$ divided by the 500 radiographs. In \textbf{a}, \textbf{d}, and \textbf{e}, the error bar spans the 95\% confidence interval, the mean $\pm$ 1.96 standard errors of the delete-one-model jackknife. Counts per panel: 17 findings, each measured on 500 of its positive radiographs in a 50{,}000-radiograph subsample at $k = 2$ (\textbf{a}, \textbf{b}), the 14 findings with a classifier (\textbf{c}), 2{,}500 radiographs per group at $k = 10$ (\textbf{d}, \textbf{e}), and 5 attributes (\textbf{f}). \textbf{a}, mKNN among the positive radiographs of each finding. \textbf{b}, mKNN against the prevalence of the finding in the chest pool. Partial marks the Spearman correlation with the AUROC of \textbf{c} held fixed. The $p$ of a Spearman correlation in \textbf{b} and \textbf{c} is asymptotic. \textbf{c}, mKNN against the AUROC of a linear classifier for the finding on the embeddings of the public DINOv3 model. \textbf{d},\textbf{e}, mKNN inside each group of an attribute recorded in every dataset (\textbf{d}) or in CheXpert alone (\textbf{e}). The number in each bar is the count of radiographs of the group in the pool, in thousands. \textbf{f}, Range of the group means of each attribute (square) against its range when the group label is permuted within each model pair. The diamond marks the mean of the permuted ranges, and the gray bar spans their 2.5th to 97.5th percentile. The largest $p_{\mathrm{FDR}}$ over the 5 attributes is printed in the panel. Six groups with fewer than 2{,}500 radiographs were not measured: two of sex, three of race, and one of ethnicity. AUROC, area under the receiver operating characteristic curve; COPD, chronic obstructive pulmonary disease; mKNN, mutual $k$-nearest-neighbor overlap; $p_{\mathrm{FDR}}$, $p$-value corrected for the false discovery rate.}
\label{sfig:subsets}
\end{figure*}

Agreement per finding was measured for the 17 findings with at least 500 positive radiographs in a 50,000-radiograph subsample of the chest pool. For each of them, the mKNN was computed on 500 of its positive radiographs at $k = 2$ (Supplementary Fig.~\ref{sfig:subsets}a). These are the 14 findings of the CheXpert vocabulary and 3 further findings, namely aortic elongation and signs of chronic obstructive pulmonary disease from PadChest and infiltration from NIH ChestX-ray14 and VinDr-CXR. The other 33 further findings have fewer positive radiographs and are listed in Supplementary Table~\ref{stab:label_mapping}. The 500 positive radiographs of a finding come from 1 to 6 source datasets. The mKNN of each finding was correlated with the logarithm of its prevalence in the pool by Spearman's rank correlation~\cite{Spearman1904Correlation}. The same correlation was then computed with the AUROC of a linear classifier for the finding held fixed, as a partial correlation. The partial correlation is the Spearman correlation between the residuals of the two quantities after a linear regression on that AUROC. The classifier was fitted on the DINOv3 ViT-L embeddings of the training radiographs of the pool and scored on its validation and test radiographs. The mKNN is not significantly correlated with the prevalence ($\rho = -0.228$, $p = 0.379$; Supplementary Fig.~\ref{sfig:subsets}b). The partial correlation is $-0.169$ ($p = 0.563$). In a post hoc analysis over the 14 findings with a classifier, the mKNN is not significantly correlated with the AUROC either ($\rho = -0.402$, $p = 0.154$; Supplementary Fig.~\ref{sfig:subsets}c).

Agreement inside demographic groups was measured for five attributes: sex, age band, race, Hispanic or Latino ethnicity, and insurance (Supplementary Fig.~\ref{sfig:subsets}d,e). The mKNN of each group was computed at $k = 10$ on 2,500 of its radiographs, drawn from those with an embedding from every model. Among the six sources of the chest pool, race, ethnicity, and insurance are recorded in CheXpert alone. Two groups of sex, three of race, and one of ethnicity have fewer such radiographs and were not measured. The range of the group means of each attribute was compared with its range under 1,000 permutations of the group label within each model pair, with FDR correction over the five attributes. Each group is one draw of radiographs. The variation between draws is therefore not part of this reference. The range of the group means is 0.005 in mKNN for sex, 0.027 for age, 0.014 for race, 0.009 for ethnicity, and 0.020 for insurance. Each observed range is larger than all 1,000 of its permuted ranges ($p_{\mathrm{FDR}} < 0.001$ for each attribute; Supplementary Fig.~\ref{sfig:subsets}f).

\begin{figure*}[p]
\centering
\includegraphics[width=\textwidth]{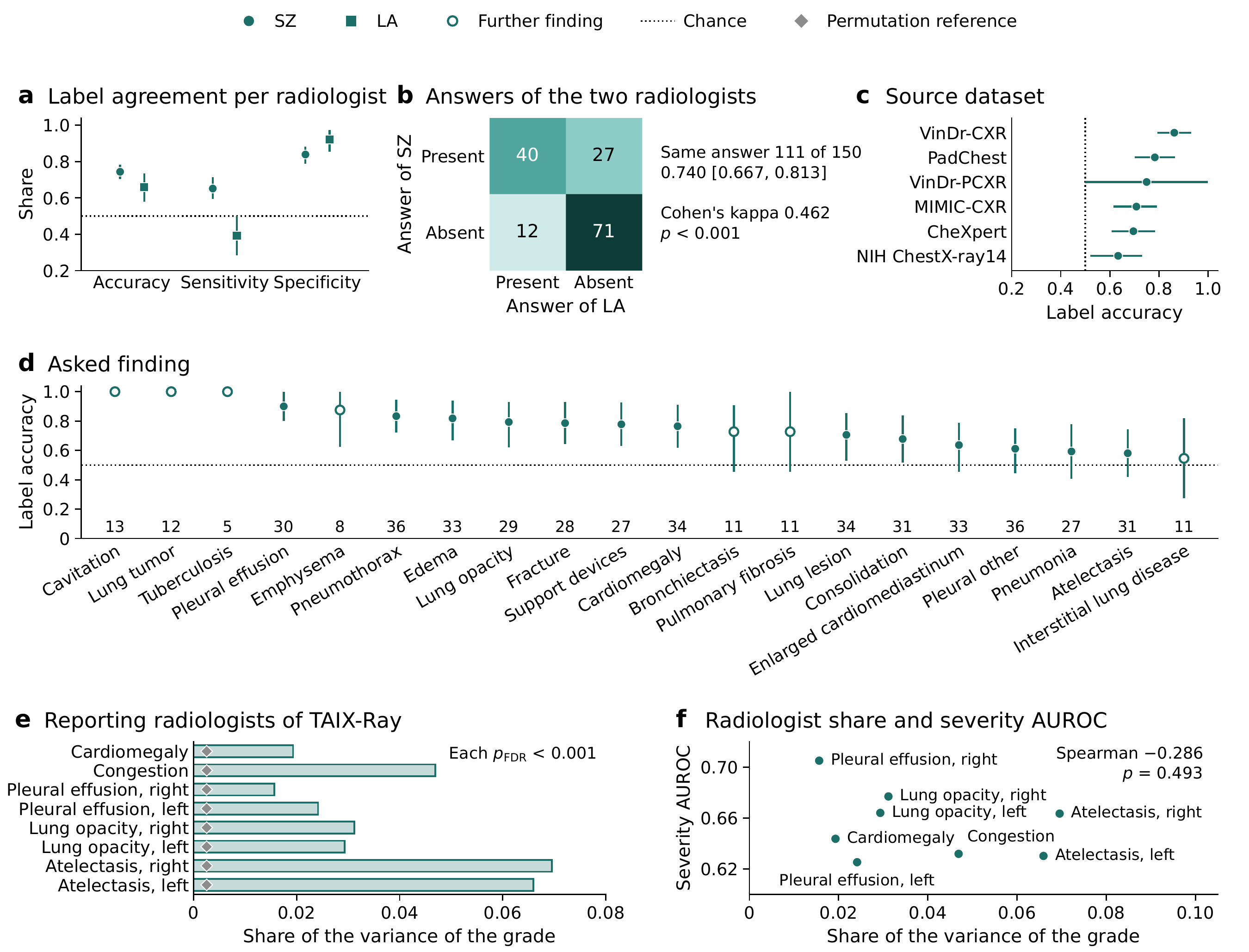}
\caption{The dataset labels against two radiologists and the variation between the reporting radiologists of TAIX-Ray. SZ and LA, two board-certified radiologists, answered for chest radiographs whether one named finding was present. Circles mark SZ and squares mark LA. An open circle marks a finding outside the CheXpert vocabulary. The dotted line marks 0.5, the chance level of a random yes-or-no answer. Each marker in \textbf{a}, \textbf{c}, and \textbf{d} is a share with its 95\% CI from 1{,}000 bootstrap resamples of the radiographs. Counts per panel: 480 radiographs of SZ and 150 further radiographs of LA (\textbf{a}), 150 radiographs read by both radiologists (\textbf{b}), 12 to 102 radiographs per source dataset (\textbf{c}), 5 to 36 radiographs per finding, printed at the foot of each column (\textbf{d}), and 50{,}000 TAIX-Ray radiographs reported by 128 radiologists for 8 graded items (\textbf{e},\textbf{f}). \textbf{a}, Accuracy, sensitivity, and specificity of the dataset label against the answer of each radiologist. Sensitivity is the share of the radiographs labeled present that were called present. Specificity is the share of those labeled absent that were called absent. \textbf{b}, Answers of the two radiologists on the radiographs read by both radiologists, as counts, beside the share of same answers with its CI and Cohen's kappa with its two-sided $p$ from 1{,}000 permutations. \textbf{c}, Accuracy of the dataset label against SZ per source dataset. \textbf{d}, The same per asked finding, highest first. \textbf{e}, Share of the variance of each graded TAIX-Ray item explained by the identity of the reporting radiologist (bars) and its mean under 1{,}000 permutations of the radiologists (diamonds). The largest FDR-corrected $p$ of the 8 items is printed in the panel. \textbf{f}, The same share against the mean severity AUROC of the 18 public models for the item, with the Spearman correlation over the 8 items and its asymptotic $p$. No model output is used in \textbf{a} to \textbf{e}. AUROC, area under the receiver operating characteristic curve; CI, confidence interval; FDR, false discovery rate.}
\label{sfig:readers}
\end{figure*}

On the 50,000 TAIX-Ray radiographs, the share of the variance of each of the eight graded items explained by the identity of the 128 reporting radiologists was computed as a one-way decomposition against 1,000 permutations of reader identity. It was correlated by Spearman's rank correlation with the models' mean severity AUROC for the item over the eight items (Supplementary Fig.~\ref{sfig:readers}e,f). Reader identity is not randomized against case mix. Each of these analyses is exploratory.

\clearpage
\begin{table}[t]
\centering
\caption{Summary of the findings, in the order of the main text. Agreement is the mKNN, and every correlation is a Spearman coefficient. Each value is a mean over the pairs, models, or classifiers named in the main text. AUROC, area under the receiver operating characteristic curve; FDR, false discovery rate; mKNN, mutual $k$-nearest-neighbor overlap.}
\label{stab:summary}
{\scriptsize\hyphenpenalty=10000\exhyphenpenalty=10000
\setlength{\tabcolsep}{3pt}
\renewcommand{\arraystretch}{0.94}
\begin{tabular*}{\textwidth}{@{\extracolsep{\fill}}>{\raggedright\arraybackslash}p{0.12\textwidth}>{\raggedright\arraybackslash}p{0.19\textwidth}>{\raggedright\arraybackslash}p{0.4\textwidth}>{\raggedright\arraybackslash}p{0.24\textwidth}@{}}
\toprule
Analysis & Comparison & Result & Interpretation \\
\midrule
\multicolumn{4}{@{}l}{\textbf{Agreement between public medical foundation models across five imaging modalities}} \\
\midrule
Public models & 18 models against the random network, 2{,}500 images per modality & mKNN 0.091 on chest radiographs against 0.036, and at most 0.300 in the other modalities. Above the random network in four of five modalities & Public models agree more than the random network and share few nearest neighbors \\
\addlinespace[0.5pt]
Lineage, size, and width & The public models & UNI and Phikon-v2 among the five most aligned pairs in four of five modalities. Spearman 0.575 with the parameter count on chest radiographs, and 0.274 at a fixed width & The most aligned pairs share a recipe or an initialization, and size has no significant association at a fixed width \\
\addlinespace[0.5pt]
Images and reports & Image models on radiographs, text models on their reports & mKNN 0.013 against 0.008 for the random network on MIMIC-CXR and CheXpert, and 0.012 against 0.007 on ReXGradient-160K & Image and text models share about one of 100 nearest neighbors \\
\midrule
\multicolumn{4}{@{}l}{\textbf{Label supervision yields the least convergent models}} \\
\midrule
Pretraining objective & Seed pairs of five objectives, all else fixed & 0.062 under label supervision and 0.411 to 0.653 under the other objectives on chest radiographs at ViT-S. The rule is met in four of five settings & Label supervision is the least convergent objective in every setting \\
\addlinespace[0.5pt]
Size and training patients & Size pairs, and image-image contrastive models on different patients & Size pairs 0.354 to 0.497 under the other objectives and 0.057 under label supervision on chest radiographs. 0.780 for a change of patients against 0.594 for a change of seed & Label supervision agrees least between sizes. A change of patients lowers agreement less than a change of seed \\
\addlinespace[0.5pt]
Initialization & A chest model at ViT-S against its DINOv3 initialization & 0.033 to 0.146 with the initialization and 0.411 to 0.653 between two models of one objective other than label supervision & A model agrees less with its initialization than two models of its objective agree with each other \\
\addlinespace[0.5pt]
Agreement and accuracy & Mean agreement of a controlled model against its accuracy & Spearman $-0.174$ over the 35 chest models. Label supervision second most accurate (AUROC 0.797) and least convergent & Convergence and accuracy are different properties \\
\midrule
\multicolumn{4}{@{}l}{\textbf{Classifier transfer between public models through a linear mapping fitted on unlabeled radiographs}} \\
\midrule
Linear mapping & 14 findings over 306 ordered pairs of public models & Retention 1.017, and AUROC 0.011 higher than a refit with up to 25{,}000 labels. Retention 0.987 with 4{,}096 unlabeled radiographs & A classifier can be transferred to a new model without new labels \\
\addlinespace[0.5pt]
Relative representation and agreement & The anchor route, and retention against agreement over 153 pairs & Retention 0.866. Spearman 0.625 for the linear mapping, added after the rules were fixed, and 0.240 for the relative representation. The pre-specified rule is not met & Models that agree more transfer classifiers with less loss through the linear mapping \\
\midrule
\multicolumn{4}{@{}l}{\textbf{Severity and laterality classification after classifier transfer on TAIX-Ray}} \\
\midrule
Severity and laterality & The two routes on presence, severity, and laterality & Linear mapping 0.787, 0.674, and 0.705 after transfer against 0.767, 0.645, and 0.730 before. Relative representation 0.572 for laterality & The linear mapping keeps severity and most of laterality, and the relative representation loses laterality \\
\midrule
\multicolumn{4}{@{}l}{\textbf{Agreement depends on the source dataset and radiographic view more than on the findings}} \\
\midrule
Source dataset and view & Agreement per radiograph regressed on four blocks of variables & Coefficient of determination 0.483 for the source dataset, 0.359 for the view, and 0.104 for the findings. Manufacturer accuracy 0.631 against 0.319 for the majority class & Agreement depends more on the source dataset and the view than on the findings \\
\addlinespace[0.5pt]
Exposure and preprocessing & Unseen datasets, and a shared preprocessing & Agreement 0.009 higher on TAIX-Ray and 0.038 higher on ReXGradient-160K than on anteroposterior MIMIC-CXR radiographs. Change of $-0.011$ under a shared preprocessing & Agreement on unseen data is not significantly lower than on MIMIC-CXR. A shared preprocessing does not change it significantly \\
\addlinespace[0.5pt]
Agreement and accuracy & Mean agreement of a public model against its AUROC & Spearman $-0.185$ ($p_{\mathrm{FDR}} = 0.463$). RAD-DINO, BiomedCLIP, and TorchXRayVision among the four least aligned and the four most accurate & Agreement does not indicate accuracy \\
\midrule
\multicolumn{4}{@{}l}{\textbf{Comparison with radiologists}} \\
\midrule
Clinical similarity & Triplets judged by two radiologists and by each model & Same choice by the two radiologists in 73 of 100 triplets (0.730). Accuracy 0.427 to 0.513 for the public and 0.442 to 0.522 for the label-supervised models & No model predicts a radiologist's choice significantly above chance \\
\bottomrule
\end{tabular*}}
\end{table}

\begin{table}[t]
\centering
\caption{The decision rules fixed before the results were computed, with the values that decide each rule and the outcome. Two rules were amended after their first computation, as stated under each. The last two rows were added after the rules were fixed. Agreement is the mean $\pm$ SD [95\% CI] from the delete-one-model jackknife, and $\rho$ is the Spearman correlation. CI, confidence interval; CKNNA, centered kernel nearest-neighbor alignment; FDR, false discovery rate; mKNN, mutual $k$-nearest-neighbor overlap; SD, standard deviation.}
\label{stab:rules}
{\scriptsize\hyphenpenalty=10000\exhyphenpenalty=10000
\setlength{\tabcolsep}{3pt}
\renewcommand{\arraystretch}{1.0}
\begin{tabular*}{\textwidth}{@{\extracolsep{\fill}}>{\raggedright\arraybackslash}p{0.2\textwidth}>{\raggedright\arraybackslash}p{0.66\textwidth}>{\raggedright\arraybackslash}p{0.1\textwidth}@{}}
\toprule
Unit & Deciding values & Outcome \\
\midrule
\multicolumn{3}{@{}l}{\textbf{Precondition, convergence of the consensus fit}} \\
\midrule
\multicolumn{3}{@{}>{\raggedright\arraybackslash}p{\textwidth}@{}}{Rule. The mean distance of the models to the consensus changes by less than $10^{-6}$ per iteration.} \\
Chest pool & A change of $3.2 \times 10^{-7}$ at iteration 4, 18 image models & Met \\
\midrule
\multicolumn{3}{@{}l}{\textbf{Precondition, a positive control with a shared structure of known dose}} \\
\midrule
\multicolumn{3}{@{}>{\raggedright\arraybackslash}p{\textwidth}@{}}{Rule. For two random representations mixed at a dose, the agreement is within three chance levels of chance at dose 0, does not fall with the dose, and at dose 1 exceeds dose 0 by more than three chance levels.} \\
\multicolumn{3}{@{}>{\raggedright\arraybackslash}p{\textwidth}@{}}{Rule. For two trained models, the agreement at dose 0 exceeds that of two untrained models by at most three of their seed SDs plus the chance level. Between consecutive doses, it falls by no more than three seed SDs. At dose 1, it exceeds dose 0 and the untrained models by more than three times the larger of the seed SD and the chance level.} \\
Mixed representations & 0.020 at dose 0 and 1.000 at dose 1, chance level 0.020 & Met \\
Trained models & 0.035 at dose 0 against 0.048 untrained, 0.691 at dose 1 against 0.160 untrained & Met \\
\midrule
\multicolumn{3}{@{}l}{\textbf{First hypothesis, public models share more neighborhood structure than the random network (Eq.~\ref{eq:h1})}} \\
\midrule
\multicolumn{3}{@{}>{\raggedright\arraybackslash}p{\textwidth}@{}}{Rule. The lower bound of the 95\% CI of the 18 models exceeds the upper bound of the 95\% CI of the random network, at the modality's own image count and $k$.} \\
Chest radiography & 0.092 $\pm$ 0.010 [0.074, 0.111] against 0.036 $\pm$ 0.004 [0.027, 0.045] & Met \\
Histopathology & 0.301 $\pm$ 0.035 [0.233, 0.369] against 0.131 $\pm$ 0.006 [0.118, 0.143] & Met \\
Dermoscopy & 0.163 $\pm$ 0.020 [0.123, 0.203] against 0.069 $\pm$ 0.008 [0.053, 0.085] & Met \\
Fundus photography & 0.146 $\pm$ 0.016 [0.116, 0.177] against 0.099 $\pm$ 0.012 [0.075, 0.122] & Not met \\
Mammography & 0.091 $\pm$ 0.009 [0.073, 0.109] against 0.057 $\pm$ 0.007 [0.042, 0.072] & Met \\
\midrule
\multicolumn{3}{@{}l}{\textbf{Second hypothesis, the pretraining objective sets how much two models agree (Eq.~\ref{eq:h2})}} \\
\midrule
\multicolumn{3}{@{}>{\raggedright\arraybackslash}p{\textwidth}@{}}{Rule. On mKNN, the spread of the objective means of seed-pair agreement exceeds the largest spread of seed-pair agreement inside one objective. The permutation $p$ and CKNNA are given beside the rule.} \\
Chest, ViT-S & mKNN 0.590 against 0.054 ($p$ = 0.035), CKNNA 0.614 against 0.050 ($p$ = 0.019) & Met \\
Chest, ViT-B & mKNN 0.519 against 0.031 ($p$ = 0.043), CKNNA 0.550 against 0.030 ($p$ = 0.035) & Met \\
Histopathology, ViT-S & mKNN 0.330 against 0.029 ($p$ = 0.057), CKNNA 0.348 against 0.036 ($p$ = 0.053) & Met \\
Histopathology, ViT-B & mKNN 0.108 against 0.299 ($p$ = 0.349), CKNNA 0.112 against 0.309 ($p$ = 0.424) & Not met \\
Second chest setting, ViT-S & mKNN 0.702 against 0.012 ($p$ = 0.088), CKNNA 0.715 against 0.012 ($p$ = 0.030) & Met \\
\midrule
\multicolumn{3}{@{}l}{\textbf{Third hypothesis within each model, presence, severity, and laterality are decodable}} \\
\midrule
\multicolumn{3}{@{}>{\raggedright\arraybackslash}p{\textwidth}@{}}{Rule. Pooled over the 18 models, more than half of the linear classifiers score above the 97.5th percentile of their label-permutation reference.} \\
Presence & 90 of 90 classifiers & Met \\
Severity & 144 of 144 classifiers & Met \\
Laterality & 54 of 54 classifiers & Met \\
\midrule
\multicolumn{3}{@{}l}{\textbf{Third hypothesis between models, transfer through the relative representation scores above chance}} \\
\midrule
\multicolumn{3}{@{}>{\raggedright\arraybackslash}p{\textwidth}@{}}{Rule. In more than half of the combinations of an ordered model pair and a target, the transferred classifier scores above the 97.5th percentile of 20 label permutations.} \\
\multicolumn{3}{@{}>{\raggedright\arraybackslash}p{\textwidth}@{}}{Amendment. The criterion was first significance after FDR correction, under which no level was met, since with 20 permutations the smallest $p$ is 1/21. It was changed to the 97.5th percentile after this first computation.} \\
Presence & 1{,}346 of 1{,}530 combinations & Met \\
Severity & 1{,}724 of 2{,}448 combinations & Met \\
Laterality & 354 of 918 combinations & Not met \\
\midrule
\multicolumn{3}{@{}l}{\textbf{Fourth hypothesis, models that agree more exchange classifiers with less loss}} \\
\midrule
\multicolumn{3}{@{}>{\raggedright\arraybackslash}p{\textwidth}@{}}{Rule. Over the 153 model pairs, the retention through 1{,}024 anchors, averaged over the 14 findings, has a positive Spearman correlation with agreement on the chest pool at $p_{\mathrm{FDR}} < 0.05$, the jackknife $p$ corrected over the three anchor counts, and its linear fit has a coefficient of determination above 0.} \\
\multicolumn{3}{@{}>{\raggedright\arraybackslash}p{\textwidth}@{}}{Amendment. The agreement was first averaged over the five modalities and tested with the naive $p$. The chest agreement and the jackknife $p$ replaced them after the first computation. With the naive $p$ of the chest pool, the rule would be met.} \\
Relative representation & $\rho$ = 0.240 over 153 pairs, jackknife $p$ = 0.271, $p_{\mathrm{FDR}}$ = 0.406, naive $p$ = 0.003 & Not met \\
Linear mapping & $\rho$ = 0.625 over 153 pairs, jackknife $p$ $<$ 0.001 & Associated \\
Controlled study & $\rho$ = 0.111 over the 595 pairs of the 35 chest models, jackknife $p$ = 0.467 & Not associated \\
\bottomrule
\end{tabular*}}
\end{table}

\begin{table}[t]
\centering
\caption{The five most aligned pairs of public models in each imaging modality. Each group heading states the number of images of the modality and the neighbor count $k$, which is 0.4\% of the images. The pairs are ranked by their mKNN on all of those images, and the two models of a pair are listed in alphabetical order. The same pairs are drawn as arcs in Fig.~\ref{fig:panel}c. mKNN, mutual $k$-nearest-neighbor overlap.}
\label{stab:pair_types}
{\footnotesize\hyphenpenalty=10000\exhyphenpenalty=10000
\setlength{\tabcolsep}{14pt}
\renewcommand{\arraystretch}{1.08}
\begin{tabular}{@{}rlr@{}}
\toprule
Rank & Model pair & mKNN \\
\midrule
\multicolumn{3}{@{}l}{\textbf{Chest radiography, 20{,}000 images, $k$ = 80}} \\
\midrule
1 & Phikon-v2 and UNI & 0.245 \\
2 & LLaVA-Med and LLaVA-OneVision & 0.200 \\
3 & Prov-GigaPath and UNI & 0.192 \\
4 & Phikon-v2 and Prov-GigaPath & 0.188 \\
5 & LLaVA-OneVision and MedGemma & 0.163 \\
\midrule
\multicolumn{3}{@{}l}{\textbf{Histopathology, 20{,}000 images, $k$ = 80}} \\
\midrule
1 & UNI2-h and Virchow2 & 0.609 \\
2 & Prov-GigaPath and UNI & 0.608 \\
3 & Phikon-v2 and UNI & 0.603 \\
4 & UNI and Virchow2 & 0.602 \\
5 & UNI and UNI2-h & 0.600 \\
\midrule
\multicolumn{3}{@{}l}{\textbf{Dermoscopy, 5{,}987 images, $k$ = 24}} \\
\midrule
1 & Phikon-v2 and UNI & 0.381 \\
2 & LLaVA-OneVision and MedGemma & 0.357 \\
3 & Prov-GigaPath and UNI & 0.345 \\
4 & UNI and UNI2-h & 0.330 \\
5 & Prov-GigaPath and Virchow2 & 0.318 \\
\midrule
\multicolumn{3}{@{}l}{\textbf{Fundus photography, 3{,}738 images, $k$ = 15}} \\
\midrule
1 & CLIP ViT-L/14 and LLaVA-Med & 0.306 \\
2 & Phikon-v2 and UNI & 0.295 \\
3 & LLaVA-Med and LLaVA-OneVision & 0.273 \\
4 & DINOv3 ViT-L and LLaVA-OneVision & 0.254 \\
5 & DINOv3 ViT-L and LLaVA-Med & 0.244 \\
\midrule
\multicolumn{3}{@{}l}{\textbf{Mammography, 2{,}500 images, $k$ = 10}} \\
\midrule
1 & LLaVA-OneVision and MedGemma & 0.239 \\
2 & CLIP ViT-L/14 and LLaVA-Med & 0.234 \\
3 & LLaVA-Med and LLaVA-OneVision & 0.198 \\
4 & LLaVA-Med and MedGemma & 0.185 \\
5 & Prov-GigaPath and UNI & 0.183 \\
\bottomrule
\end{tabular}}
\end{table}

\begin{table}[t]
\centering
\caption{Agreement of the randomly initialized ViT-S model of each objective with the DINOv3-initialized ViT-S models, per modality. The own-objective column is its mean mKNN with the three DINOv3-initialized models of its objective. The other-objectives column is its mean mKNN with the DINOv3-initialized models of the other objectives. Every value is measured on 2{,}500 held-out images at $k$ = 10. The difference is the mean $\pm$ SD, with the jackknife standard error over the partner models as the SD. Image-report contrastive learning was not trained on histopathology patches. mKNN, mutual $k$-nearest-neighbor overlap; SD, standard deviation; ViT-S, small vision transformer.}
\label{stab:cross_init}
{\footnotesize\hyphenpenalty=10000\exhyphenpenalty=10000
\setlength{\tabcolsep}{10pt}
\renewcommand{\arraystretch}{1.08}
\begin{tabular}{@{}lrrr@{}}
\toprule
Objective & Own objective & Other objectives & Difference \\
\midrule
\multicolumn{4}{@{}l}{\textbf{Chest radiographs}} \\
\midrule
Self-distillation & 0.143 & 0.123 & 0.020 $\pm$ 0.025 \\
Image-image contrastive & 0.298 & 0.063 & 0.236 $\pm$ 0.011 \\
Image-report contrastive & 0.245 & 0.135 & 0.110 $\pm$ 0.016 \\
Masked image modeling & 0.053 & 0.022 & 0.032 $\pm$ 0.003 \\
Label supervision & 0.065 & 0.124 & $-$0.060 $\pm$ 0.006 \\
\midrule
\multicolumn{4}{@{}l}{\textbf{Histopathology patches}} \\
\midrule
Self-distillation & 0.215 & 0.316 & $-$0.101 $\pm$ 0.030 \\
Image-image contrastive & 0.389 & 0.275 & 0.115 $\pm$ 0.031 \\
Masked image modeling & 0.243 & 0.158 & 0.085 $\pm$ 0.014 \\
Label supervision & 0.221 & 0.289 & $-$0.068 $\pm$ 0.019 \\
\bottomrule
\end{tabular}}
\end{table}

\begin{table}[t]
\centering
\caption{Accuracy of a linear classifier on the frozen embeddings of every controlled model. Each classifier was fitted on 70\% of the held-out patients and scored on the other 30\%. The 1{,}216 held-out NCT-CRC patches, which have no patient identifiers, were split at random into 869 training and 347 test patches. A cell is the mean $\pm$ SD over seeds or the value of a single model. The three label-supervised ViT-S models in histopathology have the same accuracy. Any other SD of 0.000 is below 0.0005. Models with random weights were trained on chest radiographs and histopathology patches. The models on the two sets of patients are ViT-S models. AUROC, area under the receiver operating characteristic curve; N/A, not applicable; SD, standard deviation; ViT-S and ViT-B, small and base vision transformers.}
\label{stab:utility}
{\footnotesize\hyphenpenalty=10000\exhyphenpenalty=10000
\setlength{\tabcolsep}{8pt}
\renewcommand{\arraystretch}{1.08}
\begin{tabular*}{\textwidth}{@{\extracolsep{\fill}}>{\raggedright\arraybackslash}p{0.42\textwidth}rrr@{}}
\toprule
Objective & \begin{tabular}[b]{@{}r@{}}ViT-S,\\DINOv3\end{tabular} & \begin{tabular}[b]{@{}r@{}}ViT-B,\\DINOv3\end{tabular} & \begin{tabular}[b]{@{}r@{}}ViT-S,\\random weights\end{tabular} \\
\midrule
\multicolumn{4}{@{}l}{\textbf{Chest radiographs, mean AUROC over the 14 findings}} \\
\midrule
Self-distillation & 0.782 $\pm$ 0.001 & 0.779 $\pm$ 0.001 & 0.747 \\
Image-image contrastive & 0.734 $\pm$ 0.002 & 0.733 $\pm$ 0.002 & 0.738 \\
Image-report contrastive & 0.825 $\pm$ 0.001 & 0.829 $\pm$ 0.001 & 0.791 \\
Masked image modeling & 0.776 $\pm$ 0.002 & 0.788 $\pm$ 0.001 & 0.728 \\
Label supervision & 0.797 $\pm$ 0.002 & 0.804 $\pm$ 0.001 & 0.784 \\
Image-image contrastive, first set of patients & 0.731 $\pm$ 0.004 & N/A & N/A \\
Image-image contrastive, second set of patients & 0.730 $\pm$ 0.003 & N/A & N/A \\
\midrule
\multicolumn{4}{@{}l}{\textbf{Histopathology patches, nine-class accuracy}} \\
\midrule
Self-distillation & 0.969 $\pm$ 0.010 & 0.964 $\pm$ 0.007 & 0.928 \\
Image-image contrastive & 0.960 $\pm$ 0.006 & 0.944 $\pm$ 0.011 & 0.957 \\
Masked image modeling & 0.977 $\pm$ 0.003 & 0.982 $\pm$ 0.004 & 0.839 \\
Label supervision & 0.980 $\pm$ 0.000 & 0.987 $\pm$ 0.002 & 0.965 \\
\midrule
\multicolumn{4}{@{}l}{\textbf{TAIX-Ray, mean AUROC over the 5 presence findings}} \\
\midrule
Label supervision, presence labels & 0.829 $\pm$ 0.001 & 0.818 $\pm$ 0.004 & N/A \\
Label supervision, severity labels & 0.830 $\pm$ 0.002 & 0.821 $\pm$ 0.002 & N/A \\
Label supervision, laterality labels & 0.813 $\pm$ 0.005 & 0.803 $\pm$ 0.007 & N/A \\
\midrule
\multicolumn{4}{@{}l}{\textbf{Second chest setting, mean AUROC over the 14 findings}} \\
\midrule
Image-image contrastive & 0.791 $\pm$ 0.001 & 0.792 & N/A \\
Image-report contrastive & 0.827 $\pm$ 0.000 & 0.829 & N/A \\
Label supervision & 0.802 $\pm$ 0.001 & 0.803 & N/A \\
Image-report contrastive, demographic sentence added & 0.825 & 0.828 & N/A \\
Image-report contrastive, demographic words removed & 0.827 & 0.830 & N/A \\
\bottomrule
\end{tabular*}}
\end{table}

\begin{table}[t]
\centering
\caption{AUROC of a classifier transferred through the relative representation minus the AUROC of a classifier refitted on the new model, per number of anchors and of labeled radiographs. The refit is fitted on the embedding of the new model or on its relative representation over 1{,}024 anchors. Its AUROC is the mean over 20 draws of the labeled radiographs. A positive value favors the transferred classifier. Cells are the mean $\pm$ SD [95\% CI] over the 4{,}284 combinations of an ordered model pair and a finding, with the delete-one-model jackknife. AUROC, area under the receiver operating characteristic curve; CI, confidence interval; SD, standard deviation.}
\label{stab:anchor_route}
{\footnotesize\hyphenpenalty=10000\exhyphenpenalty=10000
\setlength{\tabcolsep}{8pt}
\renewcommand{\arraystretch}{1.08}
\begin{tabular*}{\textwidth}{@{\extracolsep{\fill}}>{\raggedright\arraybackslash}p{0.34\textwidth}rrr@{}}
\toprule
Refit & 256 anchors & 1{,}024 anchors & 4{,}096 anchors \\
\midrule
\multicolumn{4}{@{}l}{\textbf{Refit on the embedding of the new model}} \\
\midrule
50 labeled radiographs & \begin{tabular}[t]{@{}r@{}}0.015 $\pm$ 0.004\\{[}0.007, 0.023{]}\end{tabular} & \begin{tabular}[t]{@{}r@{}}0.023 $\pm$ 0.004\\{[}0.016, 0.030{]}\end{tabular} & \begin{tabular}[t]{@{}r@{}}0.033 $\pm$ 0.005\\{[}0.024, 0.043{]}\end{tabular} \\
200 labeled radiographs & \begin{tabular}[t]{@{}r@{}}$-$0.017 $\pm$ 0.005\\{[}$-$0.026, $-$0.008{]}\end{tabular} & \begin{tabular}[t]{@{}r@{}}$-$0.009 $\pm$ 0.004\\{[}$-$0.017, $-$0.001{]}\end{tabular} & \begin{tabular}[t]{@{}r@{}}0.002 $\pm$ 0.004\\{[}$-$0.006, 0.009{]}\end{tabular} \\
1{,}000 labeled radiographs & \begin{tabular}[t]{@{}r@{}}$-$0.055 $\pm$ 0.005\\{[}$-$0.066, $-$0.045{]}\end{tabular} & \begin{tabular}[t]{@{}r@{}}$-$0.048 $\pm$ 0.005\\{[}$-$0.057, $-$0.038{]}\end{tabular} & \begin{tabular}[t]{@{}r@{}}$-$0.037 $\pm$ 0.003\\{[}$-$0.043, $-$0.030{]}\end{tabular} \\
\midrule
\multicolumn{4}{@{}l}{\textbf{Refit on the relative representation of the new model}} \\
\midrule
50 labeled radiographs & \begin{tabular}[t]{@{}r@{}}0.036 $\pm$ 0.004\\{[}0.028, 0.045{]}\end{tabular} & \begin{tabular}[t]{@{}r@{}}0.044 $\pm$ 0.004\\{[}0.037, 0.051{]}\end{tabular} & \begin{tabular}[t]{@{}r@{}}0.055 $\pm$ 0.005\\{[}0.046, 0.064{]}\end{tabular} \\
200 labeled radiographs & \begin{tabular}[t]{@{}r@{}}0.000 $\pm$ 0.005\\{[}$-$0.008, 0.009{]}\end{tabular} & \begin{tabular}[t]{@{}r@{}}0.008 $\pm$ 0.004\\{[}0.001, 0.016{]}\end{tabular} & \begin{tabular}[t]{@{}r@{}}0.019 $\pm$ 0.004\\{[}0.012, 0.026{]}\end{tabular} \\
1{,}000 labeled radiographs & \begin{tabular}[t]{@{}r@{}}$-$0.041 $\pm$ 0.006\\{[}$-$0.052, $-$0.030{]}\end{tabular} & \begin{tabular}[t]{@{}r@{}}$-$0.033 $\pm$ 0.005\\{[}$-$0.043, $-$0.023{]}\end{tabular} & \begin{tabular}[t]{@{}r@{}}$-$0.022 $\pm$ 0.004\\{[}$-$0.030, $-$0.015{]}\end{tabular} \\
\bottomrule
\end{tabular*}}
\end{table}

\begin{table}[t]
\centering
\caption{AUROC of a classifier fitted on the MIMIC-CXR training radiographs in the relative representation over 1{,}024 anchors, per finding. In the MIMIC-CXR column, the classifier is scored on the MIMIC-CXR test radiographs. In the other columns, it is scored on every labeled radiograph of five other chest datasets. Each cell is the mean over the 18 public models. N/A marks a finding without labels in that dataset or with fewer than 20 positive radiographs in it. Edema on VinDr-CXR has one positive radiograph. AUROC, area under the receiver operating characteristic curve; N/A, not applicable.}
\label{stab:cross_site}
{\footnotesize\hyphenpenalty=10000\exhyphenpenalty=10000
\setlength{\tabcolsep}{4pt}
\renewcommand{\arraystretch}{1.08}
\begin{tabular*}{\textwidth}{@{\extracolsep{\fill}}>{\raggedright\arraybackslash}p{0.2\textwidth}rrrrrr@{}}
\toprule
Finding & \begin{tabular}[b]{@{}r@{}}\\MIMIC-CXR\end{tabular} & \begin{tabular}[b]{@{}r@{}}\\CheXpert\end{tabular} & \begin{tabular}[b]{@{}r@{}}NIH\\ChestX-ray14\end{tabular} & \begin{tabular}[b]{@{}r@{}}\\PadChest\end{tabular} & \begin{tabular}[b]{@{}r@{}}\\VinDr-CXR\end{tabular} & \begin{tabular}[b]{@{}r@{}}\\VinDr-PCXR\end{tabular} \\
\midrule
Atelectasis & 0.759 & 0.618 & 0.710 & 0.728 & 0.739 & N/A \\
Cardiomegaly & 0.749 & 0.681 & 0.691 & 0.787 & 0.831 & N/A \\
Consolidation & 0.751 & 0.673 & 0.749 & 0.793 & 0.823 & N/A \\
Edema & 0.853 & 0.753 & 0.838 & 0.923 & N/A & N/A \\
Enlarged cardiomediastinum & 0.706 & 0.532 & N/A & N/A & N/A & N/A \\
Fracture & 0.642 & 0.612 & N/A & N/A & 0.624 & N/A \\
Lung lesion & 0.666 & 0.676 & 0.651 & 0.646 & 0.696 & N/A \\
Lung opacity & 0.717 & 0.658 & N/A & N/A & 0.758 & N/A \\
No finding & 0.823 & 0.809 & 0.711 & 0.780 & 0.762 & 0.597 \\
Pleural effusion & 0.840 & 0.777 & 0.786 & 0.892 & 0.835 & N/A \\
Pleural other & 0.770 & 0.635 & 0.693 & 0.702 & 0.713 & N/A \\
Pneumonia & 0.667 & 0.654 & 0.688 & 0.729 & 0.741 & 0.592 \\
Pneumothorax & 0.777 & 0.679 & 0.741 & 0.681 & 0.741 & N/A \\
Support devices & 0.855 & 0.747 & N/A & 0.833 & N/A & N/A \\
\bottomrule
\end{tabular*}}
\end{table}

\begin{table}[t]
\centering
\caption{Spearman correlation between the agreement of two models and the retention of a classifier transferred between them, over the pairs of each row. Retention is averaged over the findings and both directions of a pair. Each row reports the naive $p$ over pairs and the $p$ of the delete-one-model jackknife. In the first group, $p_{\mathrm{FDR}}$ corrects the jackknife $p$ over the three anchor counts, separately for the retention and the chance-corrected retention. In the last group, it corrects over the four groups of pairs and the three anchor counts. The other rows are single tests. FDR, false discovery rate; N/A, not applicable.}
\label{stab:link}
{\footnotesize\hyphenpenalty=10000\exhyphenpenalty=10000
\setlength{\tabcolsep}{6pt}
\renewcommand{\arraystretch}{1.08}
\begin{tabular*}{\textwidth}{@{\extracolsep{\fill}}>{\raggedright\arraybackslash}p{0.42\textwidth}rrrrr@{}}
\toprule
Analysis & $\rho$ & Naive $p$ & Jackknife $p$ & $p_{\mathrm{FDR}}$ & Pairs \\
\midrule
\multicolumn{6}{@{}l}{\textbf{Public models, relative representation}} \\
\midrule
256 anchors & 0.298 & $<$0.001 & 0.151 & 0.406 & 153 \\
1{,}024 anchors, the pre-specified rule & 0.240 & 0.003 & 0.271 & 0.406 & 153 \\
4{,}096 anchors & 0.067 & 0.412 & 0.778 & 0.778 & 153 \\
1{,}024 anchors, chance-corrected retention & 0.296 & $<$0.001 & 0.165 & 0.248 & 153 \\
\midrule
\multicolumn{6}{@{}l}{\textbf{Public models, linear mapping}} \\
\midrule
Retention & 0.625 & $<$0.001 & $<$0.001 & N/A & 153 \\
Chance-corrected retention & 0.535 & $<$0.001 & $<$0.001 & N/A & 153 \\
\midrule
\multicolumn{6}{@{}l}{\textbf{Controlled chest models, relative representation over 1{,}024 anchors}} \\
\midrule
All 35 models & 0.111 & 0.007 & 0.467 & N/A & 595 \\
All 35 models, chance-corrected retention & 0.107 & 0.009 & 0.556 & N/A & 595 \\
The 30 DINOv3-initialized models & 0.046 & 0.336 & 0.772 & N/A & 435 \\
The 16 models of the second chest setting & $-$0.122 & 0.185 & 0.613 & N/A & 120 \\
\midrule
\multicolumn{6}{@{}l}{\textbf{Public models by objective, relative representation over 1{,}024 anchors}} \\
\midrule
Two self-distillation or masked image modeling models & 0.832 & $<$0.001 & $<$0.001 & $<$0.001 & 45 \\
Two image-text contrastive models & $-$0.331 & 0.143 & 0.114 & 0.341 & 21 \\
One model of each of the two groups above & 0.109 & 0.369 & 0.625 & 0.924 & 70 \\
TorchXRayVision with a model of the first group & 0.103 & 0.777 & 0.764 & 0.924 & 10 \\
\bottomrule
\end{tabular*}}
\end{table}

\begin{table}[t]
\centering
\caption{Prevalence of the chest findings in each dataset of the chest pool after the labels were mapped onto the CheXpert vocabulary. In the first section, each cell is the share of positive radiographs among the labeled radiographs of the dataset, with the number of positive radiographs in parentheses. N/A marks a finding without labels in that dataset. The second section lists the findings outside that vocabulary with fewer than 500 positive radiographs in the 50{,}000-radiograph subsample of Supplementary Fig.~\ref{sfig:subsets}, with that number. Their agreement was not measured. Supplementary Note~\ref{snote:data} describes the mapping and the labels of TAIX-Ray. N/A, not applicable.}
\label{stab:label_mapping}
{\scriptsize\hyphenpenalty=10000\exhyphenpenalty=10000
\setlength{\tabcolsep}{4pt}
\renewcommand{\arraystretch}{1.08}
\begin{tabular*}{\textwidth}{@{\extracolsep{\fill}}>{\raggedright\arraybackslash}p{0.18\textwidth}rrrrrr@{}}
\toprule
\multicolumn{7}{@{}l}{\textbf{Findings of the CheXpert vocabulary}} \\
\midrule
Finding & \begin{tabular}[b]{@{}r@{}}\\MIMIC-CXR\end{tabular} & \begin{tabular}[b]{@{}r@{}}\\CheXpert\end{tabular} & \begin{tabular}[b]{@{}r@{}}NIH\\ChestX-ray14\end{tabular} & \begin{tabular}[b]{@{}r@{}}\\PadChest\end{tabular} & \begin{tabular}[b]{@{}r@{}}\\VinDr-CXR\end{tabular} & \begin{tabular}[b]{@{}r@{}}\\VinDr-PCXR\end{tabular} \\
\cmidrule{1-7}
Atelectasis & 21\% (48{,}790) & 19\% (26{,}313) & 10\% (11{,}559) & 6\% (6{,}166) & 1\% (148) & N/A \\
Cardiomegaly & 20\% (47{,}673) & 13\% (19{,}890) & 2\% (2{,}776) & 9\% (9{,}845) & 12\% (2{,}126) & N/A \\
Consolidation & 5\% (11{,}525) & 7\% (9{,}542) & 4\% (4{,}667) & 2\% (1{,}666) & 1\% (217) & N/A \\
Edema & 13\% (29{,}331) & 27\% (43{,}213) & 2\% (2{,}303) & 1\% (1{,}298) & 0\% (1) & N/A \\
Enlarged cardiomediastinum & 3\% (7{,}657) & 5\% (7{,}787) & N/A & N/A & N/A & N/A \\
Fracture & 2\% (4{,}781) & 4\% (6{,}445) & N/A & N/A & 0\% (55) & N/A \\
Lung lesion & 3\% (6{,}632) & 4\% (5{,}829) & 10\% (11{,}207) & 4\% (4{,}747) & 3\% (585) & N/A \\
Lung opacity & 23\% (54{,}769) & 47\% (73{,}961) & N/A & N/A & 4\% (631) & N/A \\
No finding & 33\% (81{,}117) & 11\% (17{,}000) & 54\% (60{,}361) & 33\% (36{,}148) & 70\% (12{,}652) & 66\% (6{,}050) \\
Pleural effusion & 24\% (57{,}721) & 43\% (65{,}142) & 12\% (13{,}317) & 6\% (6{,}984) & 4\% (745) & N/A \\
Pleural other & 1\% (2{,}083) & 1\% (2{,}035) & 3\% (3{,}385) & 3\% (3{,}372) & 6\% (1{,}051) & N/A \\
Pneumonia & 8\% (17{,}222) & 3\% (3{,}964) & 1\% (1{,}431) & 5\% (5{,}222) & 4\% (717) & 12\% (1{,}090) \\
Pneumothorax & 5\% (11{,}235) & 10\% (16{,}277) & 5\% (5{,}302) & 0\% (416) & 0\% (76) & N/A \\
Support devices & 30\% (73{,}294) & 58\% (90{,}967) & N/A & 2\% (2{,}294) & N/A & N/A \\
\midrule
\multicolumn{7}{@{}l}{\textbf{Further findings with fewer than 500 positive radiographs in the subsample}} \\
\midrule
Finding & Positives & \multicolumn{4}{l}{Finding} & Positives \\
\cmidrule{1-7}
Scoliosis & 418 & \multicolumn{4}{l}{Other lesion} & 35 \\
Infiltrates & 360 & \multicolumn{4}{l}{Other disease, VinDr-PCXR} & 32 \\
Other diseases, VinDr-CXR & 358 & \multicolumn{4}{l}{Calcification} & 28 \\
Chronic changes & 347 & \multicolumn{4}{l}{Interstitial lung disease} & 28 \\
Pulmonary fibrosis & 291 & \multicolumn{4}{l}{Lung tumor} & 15 \\
Air trapping & 270 & \multicolumn{4}{l}{Mediastinal shift} & 9 \\
Emphysema & 270 & \multicolumn{4}{l}{Hyaline membrane disease} & 6 \\
Kyphosis & 213 & \multicolumn{4}{l}{Lung cavity} & 3 \\
Aortic enlargement & 191 & \multicolumn{4}{l}{Congenital emphysema} & 1 \\
Sternotomy & 145 & \multicolumn{4}{l}{Congenital pulmonary airway malformation} & 1 \\
Hernia & 143 & \multicolumn{4}{l}{Enlarged pulmonary artery} & 1 \\
Bronchiectasis & 117 & \multicolumn{4}{l}{Mediastinal tumor} & 1 \\
Volume loss & 116 & \multicolumn{4}{l}{Situs inversus} & 1 \\
Bronchitis or bronchiolitis & 110 & \multicolumn{4}{l}{Chronic obstructive pulmonary disease} & 0 \\
Pulmonary congestion & 65 & \multicolumn{4}{l}{Diaphragmatic hernia} & 0 \\
Tuberculosis & 57 & \multicolumn{4}{l}{Lung cyst} & 0 \\
Cavitation & 38 & \multicolumn{4}{l}{} &  \\
\bottomrule
\end{tabular*}}
\end{table}

\end{document}